\PassOptionsToPackage{inline}{enumitem}
\documentclass[journal]{new-aiaa}
\usepackage[utf8]{inputenc}

\usepackage{float}

\usepackage{mathtools}

\usepackage{todonotes}

\usepackage[acronym]{glossaries}

\usepackage{subcaption}
\newsavebox{\largestimage}

\usepackage{booktabs}

\usepackage{doi}

\usepackage{graphicx}

\usepackage{amsmath}

\usepackage[binary-units = true,per-mode=symbol]{siunitx}

\usepackage{longtable,tabularx}
\hypersetup{
  colorlinks   = true, 
  urlcolor     = blue, 
  linkcolor    = blue, 
  citecolor   = red 
}

\DeclareSIUnit\degree{deg}
\DeclareSIUnit\pixel{px}
\DeclareSIUnit\percentagepoint{pp}
\DeclareSIUnit[number-unit-product = ]\percent{\char`\%}

\newglossarystyle{custom_acronyms}
{
    \setglossarystyle{long3col}%
}

\newignoredglossary{hidden} 

\allowdisplaybreaks

\DeclareMathAlphabet{\mathcal}{OMS}{cmsy}{m}{n}	
\DeclareMathAlphabet{\mathbb}{U}{msb}{m}{n}

\DeclareMathOperator*{\argmax}{arg\,max}
\DeclareMathOperator*{\argmin}{arg\,min}
\DeclareMathOperator{\adse}{Ad}
\DeclareMathOperator{\adset}{ad}
\DeclareMathOperator*{\diag}{diag}

\newcommand{\vect}[1]{\boldsymbol{#1}}		            
\newcommand{\vectbar}[1]{\boldsymbol{\bar{#1}}}		    
\newcommand{\vecthat}[1]{\boldsymbol{\hat{#1}}}		    
\newcommand{\vectdot}[1]{\boldsymbol{\dot{#1}}}		    

\newcommand{\rframe}[1]{\mathcal{F}_{\mathcal{#1}}}     
\newcommand{\sethree}[0]{SE\text{(3)}}
\newcommand{\sothree}[0]{SO\text{(3)}}
\newcommand{\sutwo}[0]{SU\text{(2)}}
\newcommand{\setthree}[0]{\mathfrak{se}\text{(3)}}
\newcommand{\sotthree}[0]{\mathfrak{so}\text{(3)}}
\newcommand{\urd}[0]{\mathrm{d}}

\newcommand{\icol}[1]{
  \left(\begin{matrix}#1\end{matrix}\right)%
}
\newcommand{\icolsmall}[1]{
  (#1)%
}

\makeglossaries

\newacronym{ekf}{EKF}{extended Kalman filter}
\newacronym{ml}{ML}{maximum likelihood}
\newacronym{lm}{LM}{Levenberg-Marquardt}
\newacronym{cad}{CAD}{computer aided design}
\newacronym{mli}{MLI}{multi-layer insulation}
\newacronym{sar}{SAR}{synthetic aperture radar}
\newacronym{zm}{ZM}{Zernike moment}
\newacronym{gmm}{GMM}{Gaussian mixture model}
\newacronym{ip}{IP}{image processing}
\newacronym{adr}{ADR}{active debris removal}
\newacronym{ls}{LS}{least squares}
\newacronym{irls}{IRLS}{iteratively reweighed least squares}
\newacronym{mad}{MAD}{median absolute deviation}
\newacronym{roi}{ROI}{region of interest}
\newacronym{dof}{DOF}{degrees of freedom}
\newacronym{em}{EM}{expectation-maximization}
\newacronym{pnp}{P$n$P}{Perspective-$n$-Problem}
\newacronym{lvlh}{LVLH}{local-vertical local-horizontal}
\newacronym{fov}{FOV}{field-of-view}
\newacronym{rmse}{RMSE}{root mean square error}
\newacronym{orb}{ORB}{Oriented FAST and Rotated BRIEF}
\newacronym{freak}{FREAK}{Fast Retina Keypoint}
\newacronym{nndr}{NNDR}{nearest-neighbour distance ratio}

\glsmoveentry{ekf}{hidden}
\glsmoveentry{ml}{hidden}
\glsmoveentry{lm}{hidden}
\glsmoveentry{cad}{hidden}
\glsmoveentry{mli}{hidden}
\glsmoveentry{sar}{hidden}
\glsmoveentry{zm}{hidden}
\glsmoveentry{gmm}{hidden}
\glsmoveentry{ip}{hidden}
\glsmoveentry{adr}{hidden}
\glsmoveentry{ls}{hidden}
\glsmoveentry{irls}{hidden}
\glsmoveentry{mad}{hidden}
\glsmoveentry{roi}{hidden}
\glsmoveentry{dof}{hidden}
\glsmoveentry{em}{hidden}
\glsmoveentry{pnp}{hidden}
\glsmoveentry{lvlh}{hidden}
\glsmoveentry{fov}{hidden}
\glsmoveentry{rmse}{hidden}
\glsmoveentry{orb}{hidden}
\glsmoveentry{freak}{hidden}
\glsmoveentry{nndr}{hidden}

\newacronym{ds1}{DS1}{Deep Space 1}
\newacronym{esa}{ESA}{European Space Agency}
\newacronym{gnc}{GNC}{guidance, navigation and control}
\newacronym{vvs}{VVS}{virtual visual servoing}
\newacronym{gpu}{GPU}{graphics processing unit}
\newacronym{pmf}{PMF}{probability mass function}
\newacronym{cdf}{CDF}{cumulative distribution function}
\newacronym{fast}{FAST}{Features from Accelerated Segment Test}
\newacronym{brief}{BRIEF}{Binary Robust Invariant Scalable Keypoints}

\glsmoveentry{ds1}{hidden}
\glsmoveentry{esa}{hidden}
\glsmoveentry{gnc}{hidden}
\glsmoveentry{vvs}{hidden}
\glsmoveentry{gpu}{hidden}
\glsmoveentry{pmf}{hidden}
\glsmoveentry{cdf}{hidden}
\glsmoveentry{fast}{hidden}
\glsmoveentry{brief}{hidden}

\title{Robust On-Manifold Optimization for Uncooperative Space Relative Navigation with a Single Camera}

\author{Duarte Rondao\footnote{PhD Candidate, Centre for Electronic Warfare, Information and Cyber.}}
\affil{Cranfield University, Defence Academy of the United Kingdom, SN6 8LA Shrivenham, United Kingdom}

\author{Nabil Aouf\footnote{Professor of Robotics and Autonomous Systems, Department of Electrical and Electronic Engineering.}}
\affil{City University of London, ECV1 0HB London, United Kingdom}

\author{Mark A. Richardson\footnote{Professor of Electronic Warfare, Centre for Electronic Warfare, Information and Cyber.}}
\affil{Cranfield University, Defence Academy of the United Kingdom, SN6 8LA Shrivenham, United Kingdom}

\author{Vincent Dubanchet\footnote{R\&D Engineer in AOCS and Robotics, CCPIF/AP-R\&T.}}
\affil{Thales Alenia Space, 06150 Cannes, France}

\begin{document}

\maketitle

\begin{abstract}
Optical cameras are gaining popularity as the suitable sensor for relative navigation in space due to their attractive sizing, power and cost properties when compared to conventional flight hardware or costly laser-based systems. However, a camera cannot infer depth information on its own, which is often solved by introducing complementary sensors or a second camera. In this paper, an innovative model-based approach is instead demonstrated to estimate the six-dimensional pose of a target object relative to the chaser spacecraft using solely a monocular setup. The observed facet of the target is tackled as a classification problem, where the three-dimensional shape is learned offline using Gaussian mixture modeling. The estimate is refined by minimizing two different robust loss functions based on local feature correspondences. The resulting pseudo-measurements are then processed and fused with an extended Kalman filter. The entire optimization framework is designed to operate directly on the \textit{SE}(3) manifold, uncoupling the process and measurement models from the global attitude state representation. It is validated on realistic synthetic and laboratory datasets of a rendezvous trajectory with the complex spacecraft Envisat. It is demonstrated how it achieves an estimate of the relative pose with high accuracy over its full tumbling motion.
\end{abstract}

\section{Introduction}

The concept of using optical sensors for spacecraft navigation has originated concurrently to the need for developing autonomous operations. The main rationale behind is twofold: the acquisition and processing of images is relatively simple enough to be self-contained on-board the spacecraft, avoiding the requirement of ground processing the data; and the same sensor feeds used for imaging detection and recognition can be used for navigation and mapping, alleviating additional sensors size and cost constraints. The maiden voyage of autonomous navigation was the \gls{ds1} mission \cite{bhaskaran1998orbit}, which culminated in a fly-by with comet 19P/Borrelly in 2001. More recently, in 2014 the Rosetta mission \cite{castellini2015far} used optical navigation to rendezvous with the comet 67P/Churyumov–Gerasimenko. Historically, passive camera-based navigation has been reserved for orbit determination, cruise, and fly-by sequences. For proximity operations to small bodies, such as rendezvous, docking, or landing, the full relative pose is typically required. The difficulty of these scenarios is amplified when the target is uncooperative, meaning it does not relay direct information about its state. This is often overcome by resorting to more precise active sensors, such as Lidar, which have the advantage of supplying range information and being invariant to illumination changes \cite{bercovici2017point,bercovici2019robust,kechagias2019high}, but remain difficult to integrate in on-board systems due to size and power constraints. The motivation is therefore set to increase the technology readiness level of passive optical navigation through the development of supporting \gls{ip} techniques, in order to make it a viable alternative to active systems for the close range relative navigation problem. \gls{esa}'s e.Deorbit mission is a proponent of this strategy and is set to test it in an \gls{adr} scenario by performing an uncooperative rendezvous with the non-functioning Envisat spacecraft to ultimately capture and de-orbit it \cite{biesbroek2017edeorbit}.

The most used manner to retrieve depth information in camera-based navigation systems involves the addition of a second camera forming a stereo setup: knowing the baseline between both, common landmarks detected in each image can be triangulated to obtain their relative distances to the camera frame. This has been investigated by Tweddle et al. \cite{tweddle2015factor} for the reconstruction and relative pose estimation of a tumbling body within the microgravity environment of the International Space
Station. The center of mass and ratios of inertia of the target were also estimated. Still, adding a second camera increases not only the physical size of the system, but also the \gls{ip} requirements. An alternative method is to measure the depth of the landmarks with a different sensor, or to initialize them based on a conjecture. Olson et al. \cite{olson2016small} and Razgus et al. \cite{razgus2017relative} estimate a chaser's pose relative to an asteroid with catalogued landmarks. For the latter, the distance is obtained using a laser range finder, but it is not mentioned how the automatic correspondences between the observed landmarks and the catalog are established. Y{\i}lmaz et al. \cite{yilmaz2017using} and Augenstein and Rock \cite{augenstein2009simultaneous} estimate the shape and relative pose of man-made crafts through landmark depth initialization with information from the previous rendezvous stage (e.g. another sensor or inertial data); they are suitable when the distance to the target is large when compared to its dimensions, and the convergence of each landmark's depth to their true values becomes the responsibility of the algorithm.

A different technique, and the one adopted in this paper, consists in a model-based approach, which assumes that information about the three-dimensional structure of the target is known a priori. This structure can be decomposed into elementary landmarks (``features'', as they are commonly called in the computer vision literature) that are annotated with 3D information. \gls{ip} algorithms are then developed to solve the model-to-image registration problem, i.e. the coupled pose and correspondence problems, the latter which consists in establishing matches between the target's 3D structural information and the 2D features obtained by the camera and often overlooked in pure \gls{gnc} literature. In the circumstances where the target is artificial, such as in \gls{adr}, on-orbit servicing, or docking, it is justifiable to assume that its structure, or at least part of it, is known. Indeed, to the authors' best knowledge, the first model-based yet marker-free method to estimate the six relative pose parameters via passive spacecraft imagery goes back to Cropp's doctoral thesis from 2001 \cite{cropp2001pose}. In his work, pre-generated 3D straight line features of a model of the target matched to detected 2D image lines were used to retrieve the pose. In 2006, Kelsey et al. \cite{kelsey2006vision} expanded further on this idea, applying concepts from industrial \gls{vvs} to track the pose of a spacecraft in laboratory using its known wireframe model. Later on, Petit et al. \cite{petit2013robust} upgraded the \gls{vvs} pipeline to include information from point and color features for tracking. A significant drawback of his design is that it relies on a \gls{gpu} for real-time rendering of the target model's depth map, making an implementation on current flight-ready hardware unlikely. For the past decade, advances on model-based methods have focused more on feature tracking \cite{oumer2014monocular,cai2015tsr,zou2016combining,gansmann20173d}, and not as much on feature initialization.

In this paper, a complete and innovative relative navigation framework using a monocular setup on the visible wavelength is proposed. The main contribution is the accurate pose estimation, with a single passive sensor, for the full ``360 degrees'' circular trajectory of a complex spacecraft undergoing tumbling motion. Additional contributions are made in the subjects of \begin{enumerate*} [label=\roman*\upshape)] \item pose initialization; \item cataloguing 3D model information for online use;  \item and the fusion in a robust estimation process of different \gls{ip} feature types \end{enumerate*}. The spacecraft considered for simulations and also for experimental validations is Envisat, one of the few \gls{esa}-owned debris in low Earth orbits and a possible target of the e.Deorbit mission. The presented work follows a coarse-to-fine approach where a collection of training keyframes representing different facets of the target object are rendered offline using a 3D model of it. The method first determines the keyframe in the database closest to what the camera is imaging, producing a coarse estimate of the relative pose, which is then refined using local feature matching. This reduces the problem into a 2D-2D matching process, and shifts most of the computational burden to an offline training stage. Different hypothesis generated by the matching of features are fused with an \gls{ekf}, where the error state is defined to lie on the tangent space of the special Euclidean group \textit{SE}(3), providing a concise and elegant way to update the attitude using the exponential map. The prediction stage of the \gls{ekf} is taken advantage of to help predict the locations of the features in the next frame, greatly improving the matching performance under adverse imaging conditions. Numerical simulations show that the coarse pose estimator achieves an accuracy of \SI{90}{\percent} for \SI{20}{\deg} bounds in azimuth and \SI{92}{\percent} for \SI{20}{\deg} in elevation, whereas the fine pose estimation algorithm errors do not exceed \SI{5}{\percent} of range for the translation and \SI{2.5}{\degree} for the attitude.

Section \ref{sec:mp} provides a review of the background theory used as the basis of this paper. Section \ref{sec:so} presents a top-view outline of the developed framework.  Section \ref{sec:cp} illustrates the classifier designed to retrieve a coarse estimate of the relative pose. Section \ref{sec:motionestimation} explains the motion estimation pipeline that runs nominally to generate fine estimates of the pose based on local feature matching. The \gls{ekf} used for measurement fusion is presented in Section \ref{sec:fil}. Lastly, Section \ref{sec:res} showcases the results of the designed synthetic simulations and laboratory experiments, and Section \ref{sec:co} presents the gathered conclusions. 

\section{Mathematical Preliminaries}
\label{sec:mp}

\subsection{Camera Geometry}
\label{sec:mp-camgeo}

\begin{figure}[t]
\centering
\includegraphics[width=0.75\textwidth]{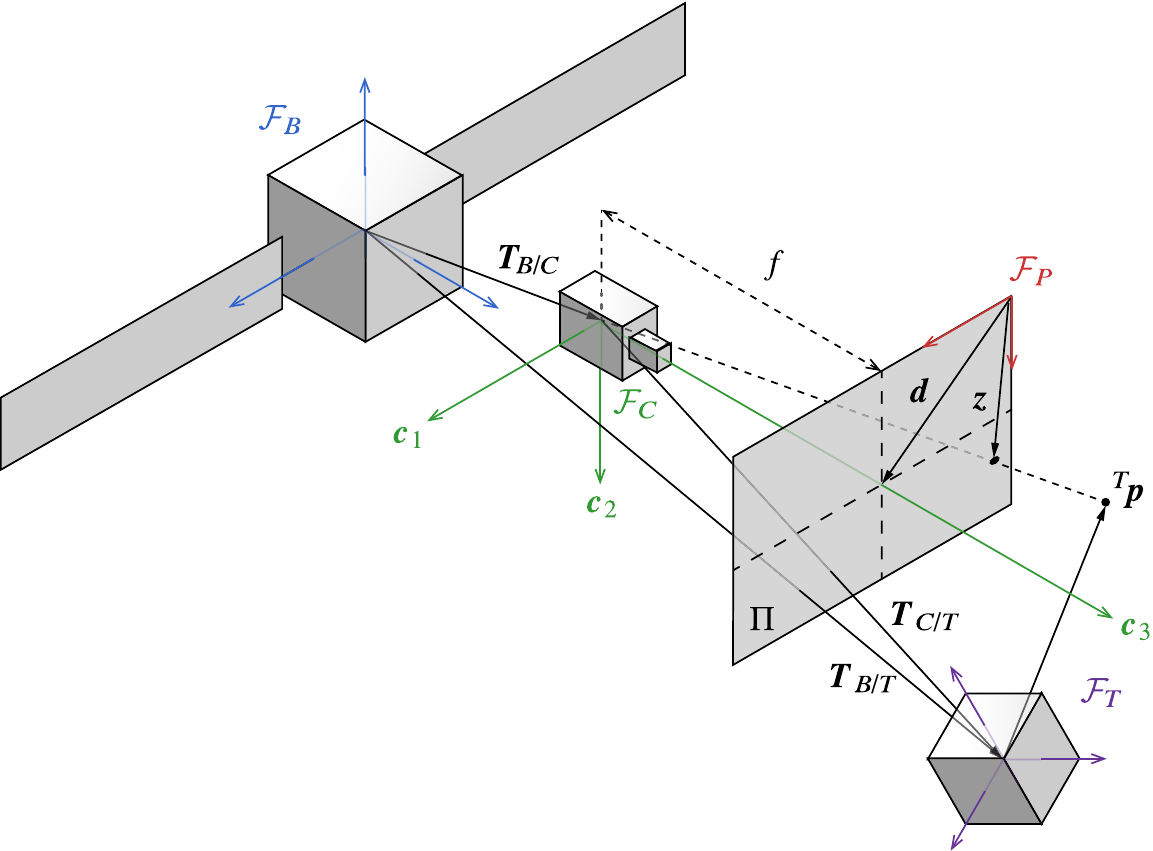}
\caption{Geometric relationships between frames of reference and landmark imaging.}
\label{fig:bg-frames}
\end{figure}

The relative pose estimation problem can be defined in terms of determining the rigid body transformation ${\vect{T} = \vect{T}_{B/T}}$ that links the frame of reference centered on the target object, $\rframe{T}$, to the chaser spacecraft's body frame, $\rframe{B}$. It is assumed that the chaser carries an on-board digital camera, which defines an additional frame of reference ${\rframe{C} = \{\vect{c}_1,\vect{c}_2,\vect{c}_3\}}$, and that $\vect{T}_{B/C}$ is known. Figure \ref{fig:bg-frames} illustrates the different frames of reference used.

Let the origin of $\rframe{C}$ define the camera center of projection, or optical center, and let $\vect{c}_3$ be aligned with the sensor's boresight. To model the relationship between the three-dimensional scene and the two-dimensional image, a pinhole camera model is adopted, which assumes the projection of all rays through the common optical center \cite{szeliski2010computer}. Then, the scene is said to be projected on a plane $\Pi$ perpendicular to $\vect{c}_3$ at a distance $f$ to the optical center, i.e. the image plane, represented by the coordinate system $\rframe{P}$. Thus, a point ${\vect{z}_P = \vect{z} = (z_1, z_2)^\top \in \Pi}$ is obtained from a point in space ${\vect{p}_T = \icolsmall{p_1, p_2, p_3}^\top_T \in \mathbb{R}^3}$ in coordinates of frame $\rframe{T}$ (cf. Fig. \ref{fig:bg-frames}) via perspective projection:

\begin{equation} 
\vect{z} = \vect{\pi}\left(\vect{K}\vect{T}\otimes\vect{p}_T\right),\quad \text{with } \vect{K} \coloneqq \begin{bmatrix}   f_x & 0 & d_1\\
                                                    0 & f_y & d_2\\
                                                    0 & 0 & 1 \end{bmatrix}, \label{eq:mp-reprojection}
\end{equation}

\noindent where $f_x, f_y$ are the scalings of $f$ by the sensor's dimensions and resolution in pixels, and ${\vect{d} = (d_1, d_2)^\top}$ are the coordinates of the principal point. The operator $\otimes$ is used to denote pose-point composition and general pose-pose composition. The matrix $\vect{K}$ represents the intrinsic parameters of the camera (in contrast to the extrinsic parameters, which are contained in $\vect{T}$), and can be obtained a priori through appropriate camera calibration. ${\vect{\pi}(\vect{\tilde{z}}) \coloneqq \tilde{z}_3^{-1}(\tilde{z}_1, \tilde{z}_2)^\top}$ is a projective function that applies the mapping from the 2D projective space $\mathbb{P}^2$ to $\mathbb{R}^2$ on a point expressed in homogeneous coordinates. Note that the equivalence ${\vect{\tilde{z}} = \lambda \vect{\tilde{z}}}$ exists for any $\lambda \in \mathbb{R} \setminus \{0\}$. For simplicity, the tilde $(\tilde{\cdot})$ notation for homogeneous points is dropped whenever the involved dimensions are unambiguous. Equation (\ref{eq:mp-reprojection}) shows that the depth of a 3D point is lost after projection.

\subsection{Lie Groups}  \nopagebreak[4]
\label{sec:mp-liegroups}

The rigid body transformation matrix $\vect{T}$ is the homogeneous representation of an element of the 3-dimensional special Euclidean group \cite{murray1994mathematical}:

\begin{equation}
\sethree \coloneqq \left\{\vect{T} = \begin{bmatrix}\vect{R} & \vect{t}\\ \vect{0}_{1\times 3} & 1 \end{bmatrix} \mathrel{\Big|} \vect{R} \in \sothree, \, \vect{t} \in \mathbb{R}^3 \right\} \subset \mathbb{R}^{4\times 4}.
\end{equation}

\noindent $\sethree$ is a 6-dimensional smooth manifold. In particular, it is a non-abelian matrix Lie group $\mathcal{G}$ with matrix multiplication as the group operation. Note that $\sethree$ (or, analogously, $\sothree$) is not a vector space. This means that the sum of two transformation (resp. rotation) matrices is not a valid transformation (resp. rotation) matrix. Since optimization frameworks are usually designed for corrective steps that consist in the addition of Euclidean spaces, incorporating a pose (resp. a rotation) is not a direct task.

However, one can exploit the local Euclidean of a manifold $\mathcal{M}$, i.e. the tangent space at each point ${x \in \mathcal{M}}$, $T_x\mathcal{M}$. The tangent space of a Lie group $\mathcal{G}$ at the identity, $T_I\mathcal{G}$, is the Lie algebra, which is a vector space \cite{stillwell2008naive}. The Lie algebra therefore linearizes the Lie group near the identity element while conserving its structure \cite{stillwell2008naive, gallier2013geometric}.

The retraction mapping ${T_I\mathcal{G} \rightarrow \mathcal{G}}$ is the exponential map, and for matrix Lie groups it corresponds to matrix exponentiation:

\begin{equation}
\exp\left(\vect{X}\right) = \sum \limits_{k=0}^\infty \frac{1}{k!}\vect{X}^k, \quad \vect{X} \in \mathbb{R}^{n \times n}. \label{eq:mp-matexp}
\end{equation}

\noindent The $(\cdot)^\wedge$ operator\footnote{Not to be confused with $\hat{\cdot} \neq (\cdot)^\wedge$.}  is used to map a vector ${\vect{\phi} \in \mathbb{R}^3}$ to the Lie algebra of $\sothree$:

\begin{equation}
  (\cdot)^\wedge_{\sotthree} \colon \mathbb{R}^3 \to \sotthree, \quad
  \vect{\phi}^\wedge \coloneqq 
  \begin{pmatrix}
  \phi_1 \\ \phi_2 \\ \phi_3
  \end{pmatrix}^\wedge
  \mapsto 
  \begin{bmatrix}
  0 & -\phi_3 & \phi_2\\
  \phi_3 & 0 & -\phi_1\\
  -\phi_2 & \phi_1 & 0
  \end{bmatrix}.
\end{equation}

\noindent This is frequently found in the literature with the analogous representation $(\cdot)^\times$ since the mapping yields a $3 \times 3$ skew-symmetric matrix such that ${\vect{a} \times \vect{b} = \vect {a}^\times \vect{b}}$. The inverse mapping ${\sotthree \to \mathbb{R}^3}$ is performed with the $(\cdot)^\vee$ operator. These two operators are overloaded to achieve a mapping between $\mathbb{R}^6$ and the Lie algebra of $\sethree$:
\begin{equation}
  (\cdot)^\wedge_{\setthree} \colon \mathbb{R}^6 \to \setthree, \quad
  \vect{\xi}^\wedge \coloneqq 
  \begin{pmatrix}
  \vect{\rho} \\ \vect{\phi}
  \end{pmatrix}^\wedge
  \mapsto 
  \begin{bmatrix}
  \vect{\phi}^\wedge & \vect{\rho}\\
  \vect{0}_{1\times 3} & 0
  \end{bmatrix} \quad \text{with} \quad \vect{\rho},\vect{\phi} \in \mathbb{R}^3.
\end{equation}

\noindent For $\sothree$ and $\sethree$, Eq. (\ref{eq:mp-matexp}) has a known closed form expression \cite{murray1994mathematical}:

\begin{equation}
\exp_{\sethree} \colon \setthree \to \sethree, \quad
\vect{\xi}^\wedge \mapsto 
\begin{bmatrix}
\exp_{\sothree}\left(\vect{\phi}^\wedge\right) & \vect{N}(\vect{\phi}) \vect{\rho}\\ \vect{0}_{1\times 3} & 1\end{bmatrix},
\label{eq:mp-exp_map_so3}
\end{equation}
with $\exp_{\sothree}(\cdot)$ given by the Rodrigues rotation formula and $\vect{N}(\vect{\phi}) \coloneqq \vect{I}_3 + (1 - \cos \| \vect{\phi}\|)\vect{\phi}^\wedge /\|\vect{\phi}\|^2 + (\| \vect{\phi} \| - \sin \|\vect{\phi}\|)\vect{\phi}^{\wedge 2}/\|\vect{\phi}\|^3$. 

It is occasionally convenient to use the adjoint action of a Lie group on its Lie algebra \cite{selig2004lie}. For $\sethree$:

\begin{equation}
\adse_{\sethree} \colon \sethree \to \mathbb{R}^{6 \times 6}, \quad
\vect{T} \mapsto 
\begin{bmatrix}
\vect{R} & \vect{t}^\wedge \vect{R}\\
\vect{0}_{3 \times 3} & \vect{R}
\end{bmatrix}. 
\label{eq:mp-liealgebra}
\end{equation}

\noindent Let $g \in \mathcal{G}$. If $\vect{T} = \vect{T}(g)$ is the homogeneous representation of the group element $g$, then ${\vect{\xi'}^\wedge = \vect{T}\vect{\xi}^\wedge\vect{T}^{-1}}$ also yields an element of $\setthree$ and the relation can be written linearly in $\mathbb{R}^6$ as ${\vect{\xi'} = \adse(\vect{T})\vect{\xi}}$. Furthermore, the adjoint action of the Lie algebra on itself is
\begin{equation}
\adset_{\setthree} \colon \setthree \to \mathbb{R}^{6 \times 6}, \quad
\vect{\xi}^\wedge \mapsto 
\begin{bmatrix}
\vect{\phi}^\wedge & \vect{\rho}^\wedge\\
\vect{0}_{3 \times 3} & \vect{\phi}^\wedge
\end{bmatrix},
\end{equation}
such that the expression for the Lie bracket of $\setthree$ can be written as $[\vect{\xi}_0, \vect{\xi}_1] \coloneqq \vect{\xi}_0\vect{\xi}_1 - \vect{\xi}_1\vect{\xi}_0 = (\adset(\vect{\xi}_0^\wedge)\vect{\xi}_1)^\wedge$.

\subsection{Optimization Framework for Manifolds}

The manifold optimization framework developed for this work revolves around the importance of the tangent space as a local vector space approximation for the pose manifold. Let $R_x$ be a retraction at ${x \in \mathcal{M}}$ that maps ${T_x\mathcal{M} \rightarrow \mathcal{M}}$. In addition, let $0_x$ be the 0-element of $T_x\mathcal{M}$ such that $R_x(0_x) = x$, and $h$ a real-valued function acting in $\mathcal{M}$. Then, if $\mathcal{M}$ is endowed with a Riemannian metric, one can write:

\begin{equation}
\nabla \left(h \circ R_x\right)\left(0_x\right) = \nabla h(x), \label{eq:mp-manifoldgrad}
\end{equation}

\noindent i.e. the retraction preserves gradients at $x$ \cite{absil2007optimization}. This means that optimization problems based on Euclidean spaces relying on the computation of gradients (or some approximation thereof) can be generalized to (nonlinear) manifolds via retraction mappings.

By extension, since a Lie group is a smooth manifold, any $\mathcal{G}$ such as $\sethree$ can be endowed with a Riemannian metric \cite{gallier2011notes} and Eq. (\ref{eq:mp-manifoldgrad}) applies. As such, the exponential map can be used as the bridge to locally convert an optimization problem stated in terms of $\vect{T}$ to the more tractable vector space of the corresponding Lie algebra element $\vect{\xi}^\wedge$ (or simply its compact representation ${\vect{\xi} \in \mathbb{R}^6}$), where methods of Euclidean analysis can be used. Formally, for an incremental correction $\vect{\xi}$, a solution in the manifold can be propagated as
\begin{equation}
\vect{T'} = \exp\left(\vect{\xi}^\wedge\right)\vect{T}, \quad \text{with} \quad \vect{T'}, \vect{T} \in \sethree \quad \text{and} \quad \vect{\xi}^\wedge \in \setthree, \label{eq:mp-poseexp}
\end{equation}
where the left-product convention has been adopted. It is also useful to see $\sethree$ as a semi-direct product of manifolds ${\sothree \rtimes \mathbb{R}^3}$, as one might be interested in working with isomorphic representations of $\sothree$, such as the special unitary group $\sutwo$ of unit quaternions, with the well-known isomorphism \cite{markley2014fundamentals}:

\begin{equation}
\vect{R}(\vect{q}) = \left(q^2 - \|\vect{e}\|^2\right)\vect{I}_3 - 2q\vect{e}^\wedge + 2\vect{e}\vect{e}^\top, \quad \text{with} \quad \vect{q} \in \sutwo \quad \text{and} \quad \vect{R} \in \sothree,
\end{equation}

\noindent where $\vect{e}$ and $q$ are the vector and scalar parts of the quaternion, respectively, which is written as
\begin{align}
    \vect{q} \coloneqq \begin{pmatrix} \vect{e}\\ q \end{pmatrix}
\end{align}
As it is familiar in the space domain, the composition of two attitude quaternions is taken in the form of Shuster's product \cite{shuster1993survey}, meaning that rotations are composed in the same order as for rotation matrices:
\begin{align}
\vect{q}_0 \otimes \vect{q}_1 &= 
\begin{bmatrix}
q_0\vect{I}_3 - \vect{e}_0^\wedge & \vect{e}_0\\
-\vect{e}_0^\top & q_0
\end{bmatrix} \vect{q}_1.
\end{align}
If a Lie group $\mathcal{G}$ is a manifold obtained through the semi-direct product of some isomorphism of $\sothree$ and $\mathbb{R}^3$, then $\mathcal{G}$ is isomorphic to $\sethree$ as a manifold, but not as a group \cite{gallier2013geometric}. The operator ${\oplus \colon \mathcal{G} \times \mathbb{R}^6 \to \mathcal{G}}$ is thus defined to generalize a composition of a group element ${g \in \mathcal{G}}$ representing a pose and an element $\vect{\xi}$ which is the compact representation in $\mathbb{R}^6$ of $\vect{\xi}^\wedge \in \setthree$:
\begin{equation}
g' = g \oplus \vect{\xi}, \quad \text{with} \quad g, g' \in \mathcal{G}, \quad \text{and} \quad \mathcal{G} \cong \sethree. \label{eq:mp-posecomposition}
\end{equation}
Likewise, one defines the inverse operation ${\ominus \colon \mathcal{G} \times \mathcal{G} \to \mathbb{R}^6}$ that yields the compact representation of an element of the Lie algebra.

\section{System Overview}
\label{sec:so}

\begin{figure}[t]
\centering
\includegraphics[width=\textwidth]{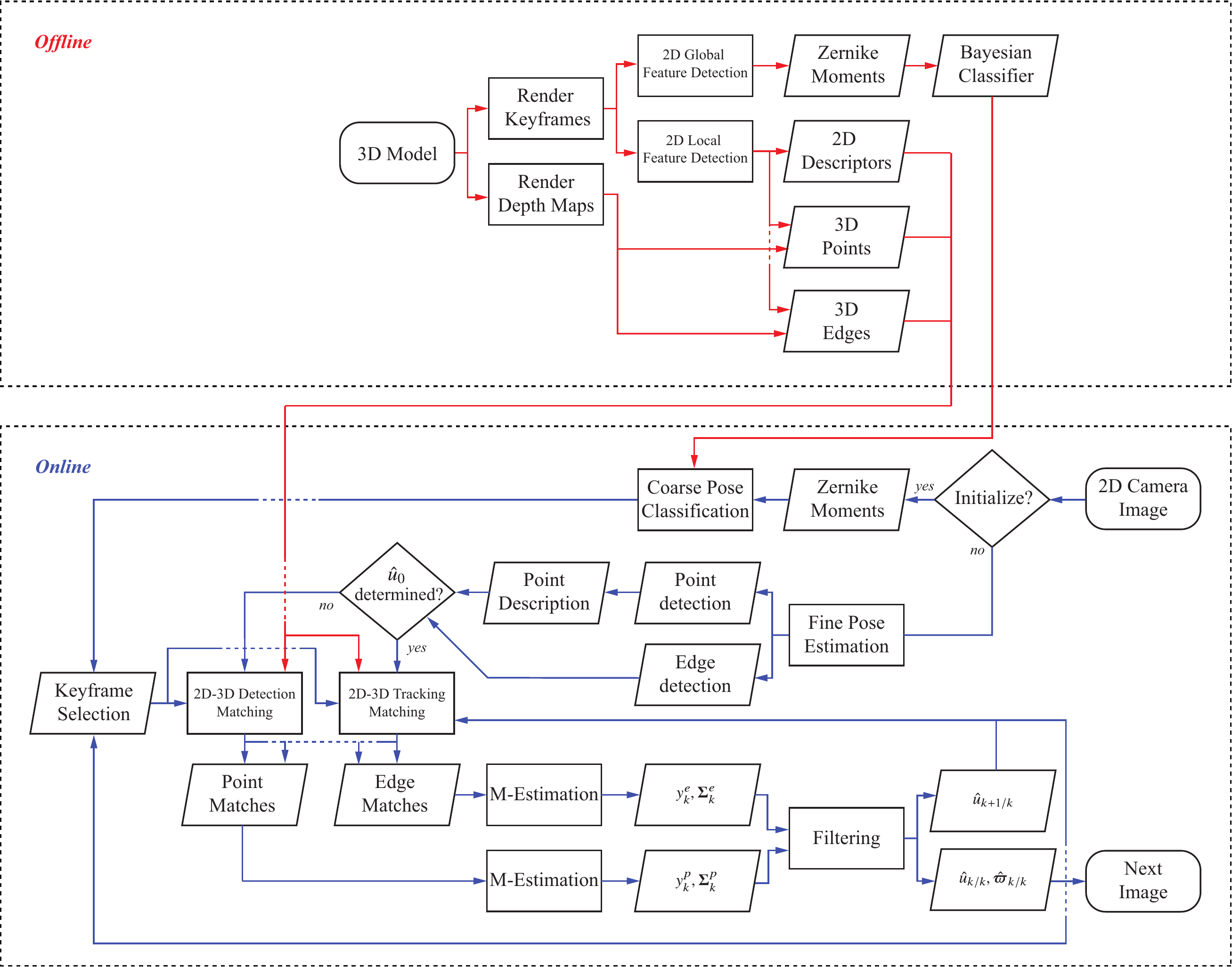}
\caption{Simplified flowchart of the relative navigation framework's structure.}
\label{fig:so-method}
\end{figure}

Figure \ref{fig:so-method} illustrates the architecture of the relative navigation framework presented in this paper. It is divided into two fundamental components: an offline training stage (in red), and an online pipeline (in blue).

The offline training stage has as its objective to discretize, categorize, and represent the three-dimensional structure of the target so that it can be utilized in the two-dimensional environment of the online stage. Two sets of images are sampled from different viewing angles of the target's \gls{cad} model. The first is a set of keyframes, each of which contains textural information from the target as imaged from that viewpoint. The second is a set of depth maps, each of which has the same scene structure as its corresponding keyframe, but the value of each pixel represents the distance of that point in the target to the image plane. For each keyframe, the shape of the target is mathematically represented using complex \gls{zm}. The distribution of the \gls{zm} feature vector elements per class is modeled using \glspl{gmm}, which will define the likelihood probabilities in the training of a Bayesian classifier later employed to match the target's facet as observed by the on-board camera to the closest keyframe in the database, defining the coarse pose classification module (Section \ref{sec:cp}). 

The keyframes are also processed with a feature point detector. The aim is to identify keypoints distinguishable enough to be matched to the same keypoint in the context of the online pipeline. Each keypoint is subjected to a feature descriptor, which generates a signature vector used to search and match in the descriptor space. Using the depth map corresponding to its keyframe, each keypoint is annotated with its position on the target's structure, generating a 3D-to-2D keypoint catalog to be used with \gls{ip} algorithms compatible with camera-based navigation. Additionally, the target's limb (or contour) in each keyframe is locally sampled into control points using edge detection. The edge points are converted to 3D using the depth map and grouped into 3D straight keylines. It was found that existing keyline descriptors were not mature enough for the present application, so alternative strategies were instead designed (Section \ref{sec:motionestimation}).

The online stage has the purpose of providing a fine pose estimate based on local feature matching after the closest keyframe has been found using coarse pose classification. If no estimate of the pose, $\hat{u} \in \mathcal{U} \cong \sethree$, has been determined, local features are matched by detection: keypoints from the database pertaining to the current keyframe are matched by brute-force to the ones detected in the camera image, whereas the edges are matched by aligning the keyframe contour to the camera image contour in the least squares sense. Otherwise, the features are matched by tracking. This is not meant in the typical sense that the features are propagated from one camera image to the next, but instead the search space is reduced by reprojecting them from 3D into 2D based on $\hat{u}$ (Section \ref{sec:me-structconst}).

The feature matches are processed separately and used to generate direct pseudo-measurements of the 6 \gls{dof} relative pose. This is achieved by minimizing the reprojection error using \gls{lm} (Section \ref{sec:me-problem}) in an M-estimation framework (Section \ref{sec:me-re}), which implements the rejection of outlying matches. The measurements are fused with an \gls{ekf} to produce an estimate of the relative pose and velocity (Section \ref{sec:fil}). Both the M-estimator and the filter are accordant in representing the pose error as an element of $\setthree$, meaning that the measurement covariance determined from the former is used directly as the measurement noise in the latter, avoiding the need for tuning. The pose predicted by the filter for the following time-step is used to select the next keyframe and in the matching by tracking, providing temporal consistency.

\section{Coarse Pose Classification}
\label{sec:cp}

\subsection{Viewsphere Sampling}
\label{subsec:viewsphere}

\begin{figure}[t]
	\centering
	\includegraphics[width=\textwidth]{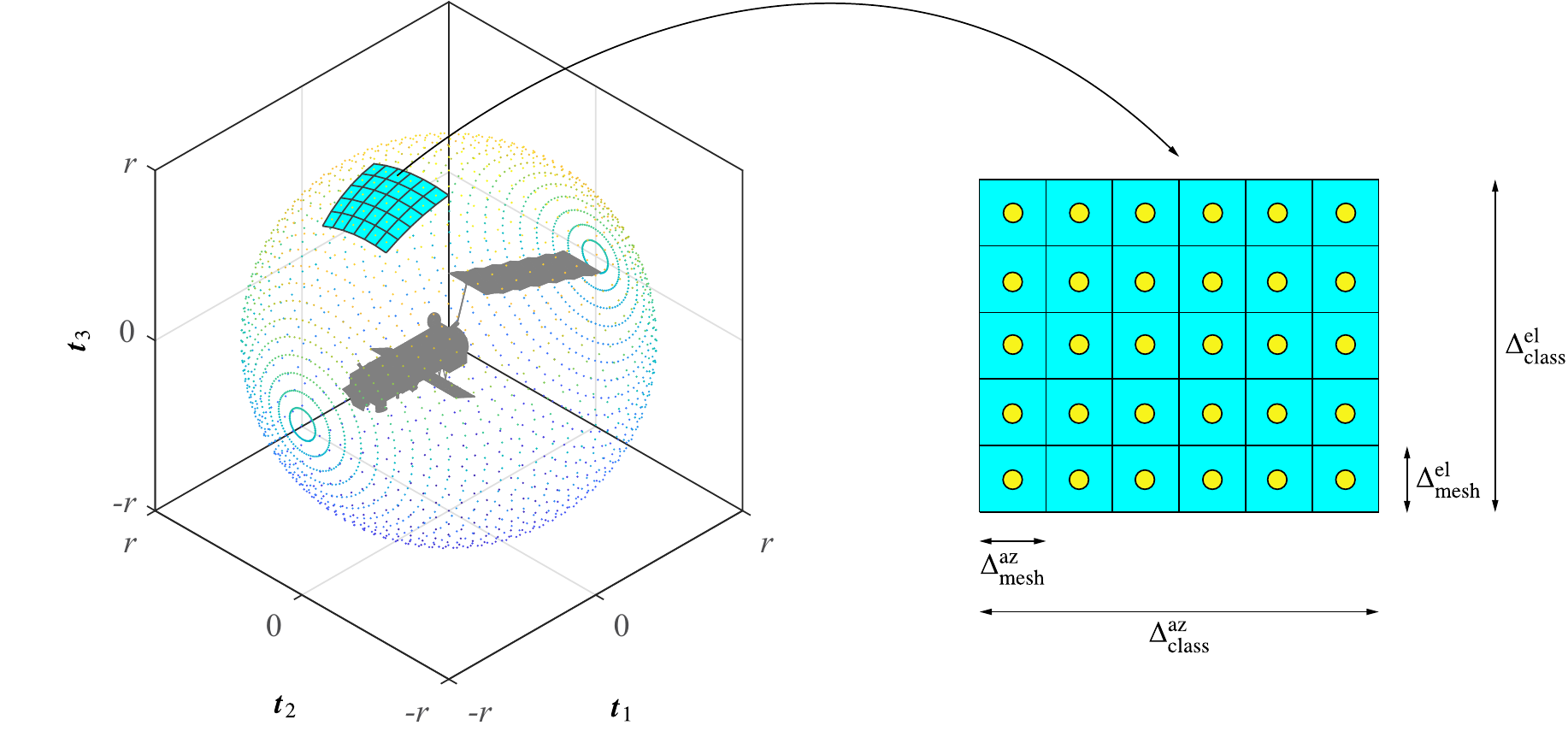}
  \caption{The viewsphere for aspect sampling of a spacecraft (not to scale).}
  \label{fig:cp-modelviewsphere}
\end{figure}

The concept of this module is to recover the viewpoint of the three-dimensional target object imaged in a two-dimensional scene using its pre-computed and known \gls{cad} model. The objective is to provide an initial, coarse, estimate of the appearance of the object based on its view classification so that then more precise pose estimation algorithms can be used to refine its pose.

In order to capture the full three-dimensional aspect of the target, sampled views from the \gls{cad} are generated by resorting to the concept of the viewsphere: the model is located at the centre of a sphere, on the surface of which several cameras are placed, pointed at its centre of mass. The necessary viewpoints can be obtained by varying the spherical coordinates of the camera's position, i.e. the azimuth, elevation, and distance. The viewsphere is illustrated in Fig. \ref{fig:cp-modelviewsphere}. Each dot represents a camera position on the target body frame $\rframe{T}$ that will sample a view. Regarding the training of the sampled data, two different approaches using this viewsphere can be outlined. The first approach involves treating each dot on the viewsphere as a class. This has the immediate disadvantage that if a very fine mesh is defined (low $\Delta_\text{mesh}$), the classes will not be distinctive enough, which could affect the performance of the view classification. On the other hand, selecting a high $\Delta_\text{mesh}$ does not solve the issue that each class will have only exactly one training image to use for the classification scheme. In order to solve both problems, a second approach is adopted in which dots are grouped into patches of width $\Delta_\text{class}$ to form a class, illustrated as the cyan patch in Fig. \ref{fig:cp-modelviewsphere}.

\subsection{Global Feature Description}

The following step is to select a measure to mathematically describe each training image obtained as explained above. Such a descriptor will be the basis to establish a correspondence between two viewpoints. The choice for a descriptor for viewpoint classification was motivated by by two main points: \begin {enumerate*} [label=\roman*\upshape)] \item it must be a global representation of the target and \item it must be robust to changes likely to be experienced during a space imaging scenario. \end{enumerate*} The first point is justified by the fact that the goal is a classification of the aspect of the target, i.e. what is the view from the database that most closely resembles what the camera is observing. While it is possible in theory to use local descriptors for this task the query space would be much larger given that several features would be required to describe a single view. When the target is a spacecraft, the same local features can be expected to be present in multiple views (e.g. those sampled from \gls{mli} or solar panels), which would make the view classification harder. The second point refers to robustness against the model and what is actually observed during the mission; since modeling all the expected cases would be intractable, the descriptors should be resilient towards these, namely: translation, rotation, and scale changes (i.e. the expected 6-\gls{dof} in space), off-center perspective distortions, and illumination changes.

\subsubsection{Image Moments}

One type of descriptor that satisfies the above requirements are image moments. Moments are projections of an image $\mathfrak{I}$ onto a d-variable polynomial basis $\chi_{\vect{n}}$, with ${\vect{n}=\icolsmall{n_1,\ldots,n_d}}$ of the space of image functions defined on $\Pi$ \cite{flusser20162d}. Formally:

\begin{equation}
	M_{\vect{n}} = \int_\Pi \chi_{\vect{n}}(\vect{z}) \mathfrak{I}(\vect{z}) \, \urd \vect{z}, \label{eq:cp-genmoment}
\end{equation}

\noindent $\mathfrak{I}(\vect{z})$ denotes the intensity value of pixel $\vect{z}$ in the image. Taking $\chi_{\vect{n}}(\vect{z}) = \vect{z}^{\vect{n}}$ leads to the well known geometric image moments that describe the ``mass distribution'' of the image: $M_{00}$ is the mass of the image, $M_{10}/M_{00}$ and $M_{01}/M_{00}$ define the centroid, and so on. Moment computation over a regular image is dependant on the intensity value $\mathfrak{I}(\vect{z})$ of each pixel. This implies that the result of Eq. (\ref{eq:cp-genmoment}) will not be robust to illumination changes. Normalizing the image would provide global illumination invariance, but not local, therefore another strategy is needed. To this end, the viewpoint image is first binarized before computing the moments. This involves processing the image such that the resulting pixel intensities are mapped to either $\mathfrak{I}(\vect{z}) = 0$ or $\mathfrak{I}(\vect{z}) = 1$. In this way, the target is analyzed in terms of its shape, independently of how each patch is illuminated.

\subsubsection{Complex Zernike Moments}

Consider the \acrfull{zm} of the $n$-th degree with repetition $\ell$, defined in 2D polar coordinates as \cite{flusser20162d}:
\begin{equation}
    A_{n\ell} = \frac{n + 1}{\pi}\int\limits^{2\pi}_0\int\limits^1_0 V^{\ast}_{n\ell}(r, \theta) f(r, \theta) r \, \urd r \urd \theta,
\end{equation}
where
\begin{align*}
    V^{\ast}_{n\ell}(r, \theta) &= R_{n\ell}(r)e^{i\ell \theta} \quad \text{with} \quad n = 0, 1, 2, \ldots \quad \text{and} \quad \ell = -n, -n+2,\ldots,n\\
    R_{n\ell}(r) &= \sum\limits_{s = 0}{(n - \lvert \ell \rvert)/2} (-1)^s \frac{(n - s)!}{s!\left(\frac{n + \lvert \ell \rvert}{2} - s \right)! \left(\frac{n - \lvert \ell \rvert}{2} - s \right)!} r^{n - 2s}.
\end{align*}
and $(\cdot)^\ast$ denotes complex conjugation. \glspl{zm} have two main attractive properties. Firstly, they are circular moments, meaning they change under rotation in a simple way which allows for a consistent rotation invariant design. Secondly, they are orthogonal moments, which means that they present significant computational advantages with respect to standard moments, such as low noise and uncorrelation. Additionally, orthogonal moments can be evaluated using recurrent relations. 

Since they carry these two traits, \glspl{zm} are said to be orthogonal on a disk. Scale invariance is obtained when the image is mapped to the unit disk before calculation of the moments. Translation invariance is obtained by changing the coordinate system to be centered on the centroid. Regarding rotation invariance, one option occasionally seen is to take the \gls{zm} as the magnitude $\lvert A_{n\ell} \rvert$. This is not a recommended approach, as essentially the descriptor is cut in half, leading to a likely loss in recognition power. Instead, this work will deal explicitly with both real and complex parts of each \gls{zm}, in which case rotation invariance can be achieved by normalizing with an appropriate, non-zero moment $A_{m'\ell'}$ (typically $A_{31}$).

A fast computation of the Zernike polynomials up to a desired order can be obtained recursively since any set of orthogonal polynomials obeys a recurrent relation for three terms; in the case of \glspl{zm} the following formula has been developed by Kintner \cite{kintner1976mathematical}:
%
\begin{equation}
 k_1 R_{n+2,\ell}(r) = (k_2 r^2 + k_3) R_{n\ell}(r) + k_4 R_{n-2,\ell}(r),
\end{equation}
where $k_i,\, i=1,\ldots,4$ are constants dependant on $n$ and $\ell$.

\subsection{Training the Data}

Given the process of generating the data and its descriptors, the final step is defining the classification method. The classifier algorithm shall recognize the aspect of the target given a database of \gls{zm} descriptor representation of viewpoints. Given the large volume of data involved, a Bayesian classifier is considered for this task, where the probability density function of each class is approximated using \glsreset{gmm}\glspl{gmm}.

\subsubsection{Bayesian Classification}

Given a specific class $\mathcal{C}_m$, ${m=1, \ldots, k}$, and a $d$-dimensional feature vector ${\vect{y} = \icolsmall{y_1,\ldots,y_d}^\top}$, a Bayesian classifier works by considering $\vect{y}$ as the realization of a random variable $\vect{Y}$ and maximizing the posterior probability ${P\left(\mathcal{C}_m \vert \vect{y} \right)}$, i.e. the probability that the feature vector $\vect{y}$ belongs to $\mathcal{C}_m$. This probability can be estimated using Bayes' formula \cite{duda2012pattern}:
\begin{equation}
P\left(\mathcal{C}_m \vert \vect{y} \right) = \frac{p\left(\vect{y} \vert \mathcal{C}_m \right) P \left(\mathcal{C}_m \right)}{\sum\limits_{i=1}^k p(\vect{y} \vert \mathcal{C}_i) P(\mathcal{C}_i)}. \label{eq:cp-bayesformula}
\end{equation}
The denominator is independent from $\mathcal{C}_m$ and hence can be simply interpreted as a scaling factor ensuring ${P\left(\mathcal{C}_m \vert \vect{y} \right) \in [0, 1]}$. Therefore, maximizing the posterior is equivalent to maximizing the numerator in Eq. (\ref{eq:cp-bayesformula}).

The prior probability, $P\left(\mathcal{C}_m \right)$, expresses the relative frequency with which $\mathcal{C}_m$ will appear during the mission scenario; for a general case where one has no prior knowledge of the relative motion, an equiprobable guess can be made and the term can be set to $1/N$ for any $m$. The challenge is therefore to estimate the likelihood $p(\vect{y} \vert \mathcal{C}_m )$ of class $\mathcal{C}_m$, which is given by the respective probability density.

\subsubsection{Gaussian Mixture Modeling}

The Gaussian distribution is frequently used to model the probability density of some dataset. In the scope of the present work, it may prove overly optimistic to assume that all elements of the \gls{zm} descriptor vectors for each class are clustered into a single group. On the other hand, it can be too restrictive to model their distribution using hard-clustering techniques in case boundaries are not well defined. A more controllable approach to approximate a probability density function, while keeping the tractability of a normal distribution, is to assume the data can be modelled by a mixture of Gaussians:
\begin{equation}
p(\vect{y} \vert \vect{\theta}) = \sum\limits^n_{i = 1} \alpha_i \mathcal{N}\left(\vect{y};\vect{\mu}_i,\vect{\Sigma}_i\right), 
\quad
\text{with}
\quad
\mathcal{N}\left(\vect{y};\vect{\mu},\vect{\Sigma}\right) = \frac{1}{\sqrt{(2\pi)^d \lvert \vect{\Sigma} \rvert}} \exp \left( -\frac{1}{2}\left(\vect{y} - \vect{\mu}\right)^T \vect{\Sigma}^{-1} \left(\vect{y} - \vect{\mu}\right) \right),
\end{equation}
\noindent where $\alpha_i$ are scalar weighing factors, $n$ is the number of mixture components, $\vect{\mu}$ denotes the mean vector, and $\vect{\Sigma}$ the covariance matrix, and ${\vect{\theta} = \{\vect{\mu}_1,\vect{\Sigma}_1,\alpha_1, \ldots, \vect{\mu}_n,\vect{\Sigma}_n,\alpha_n \}}$ is the full set of parameters required to define the \gls{gmm}.

When the number of mixture components $n$ is known, the ``optimal'' mixture for each class, in the \gls{ml} estimation sense, can be determined using the classical \gls{em} algorithm. \gls{em} works on the interpretation that the set of known points ${\mathcal{Y} = \{\vect{y}_1,\ldots,\vect{y}_m\}}$ is part of a broader, complete, data set ${\mathcal{X} = \mathcal{Y} \cup \check{\mathcal{Y}}}$ that includes unknown features \cite{duda2012pattern}. In the case of \glspl{gmm}, or finite mixtures in general, ${\check{\mathcal{Y}} = \{\vect{\check{y}}_1,\ldots,\vect{\check{y}}_m\}}$ can be defined as the set of $m$ labels denoting which component generated each sample in $\mathcal{X}$. Each ${\vect{\check{y}}_i = \icolsmall{\check{y}_{i,1},\ldots,\check{y}_{i,n}}^\top}$ is a binary vector such that ${\check{y}_{i,p} = 1, \check{y}_{i,q} = 0}$ for all ${p\neq q}$ if sample $\vect{\check{y}}_i$ has been produced by component $p$. 

However, the number of components is usually not known a priori. There are several methods to iteratively estimate the $n$; for this work the method of Figueiredo and Jain \cite{figueiredo2002unsupervised} is adopted. The algorithm provides an alternative to the generation of several candidate models, with different numbers of mixture components, and subsequent selection of the best fit, as this approach would still suffer from the drawbacks of \gls{em}; namely, the fact that it is highly dependant on initialization, and the possibility of one of the mixtures' weight $\alpha_i$ approaching zero (i.e. the boundary of the parameter space) and the corresponding covariance becoming close to singular. Instead, Figueiredo and Jain's method aims to find the best overall model directly. This is achieved by applying the minimum message length criterion to derive the following cost function for finite mixtures:
\begin{equation}
    L\left(\vect{\theta},\mathcal{Y}\right) = \frac{N}{2}\sum\limits^n_{i=1} \ln \left(\frac{m \alpha_i}{12}\right)+\frac{n}{2}\ln\frac{m}{12} + \frac{n\left(N + 1\right)}{2} - \ln p\left(\mathcal{Y};\vect{\theta}\right), \label{eq:cp-fjequation}
\end{equation}
where $N$ is the number of parameters specifying each component. A modified \gls{em} is utilized to minimize Eq. (\ref{eq:cp-fjequation}), with the M-step given by:
\begin{equation}
\hat{\alpha}_p(k+1) = \frac{ \max \left\{0, \left(\sum_{i=1}^m w_{i,p}\right) - \frac{N}{2} \right\} }{\sum_{j=1}^n  \max \left\{0, \left(\sum_{i=1}^m w_{i,j}\right) - \frac{N}{2} \right\} }, \quad
\hat{\vect{\theta}}_m(k+1) \leftarrow \argmax_{\vect{\theta}_m} Q(\vect{\theta},\hat{\vect{\theta}}(k)),
\end{equation}
for $m = 1,\ldots,n$, where $t=k,k+1$ are sequential time-steps, and ${w_{i,p} \in \mathcal{W} \coloneqq E[\check{\mathcal{Y}}\vert \mathcal{Y};\hat{\vect{\theta}}(k)]}$ is the a posteriori probability that $\check{y}_{ip} = 1$ after observing $\vect{y}_i$, computed as in the regular \gls{em}. The modified M-step performs explicit component annihilation, meaning that when one of the $m$ components becomes unsupported by the data (i.e. close to zero), it is removed, thus impeding the algorithm from approaching the boundary of the parameter space. On the other hand, robustness towards initialization is achieved by starting the procedure with a large $n$ and iteratively removing the unnecessary ones. If $n$ is too large, it may occur that no component is granted enough initial support, leading the $\hat{\alpha}_i$ to be underdetermined. This is avoided by performing a component-wise update, i.e. recomputing $\mathcal{W}$ every time each element $\alpha_i, \vect{\theta}_i$ is updated, rather than doing it until the last $i = n$; in this way, if one component dies off, its probability mass is automatically redistributed to the other components, increasing their chance of survival. The proposed modifications will allow the modeling of each training class as a probability density in an unsupervised way.

\subsubsection{Remarks}

\begin{figure}[t]
	\centering
	\begin{subfigure}[t]{0.5\textwidth}	\centering
	\includegraphics[height = 0.3\textheight]{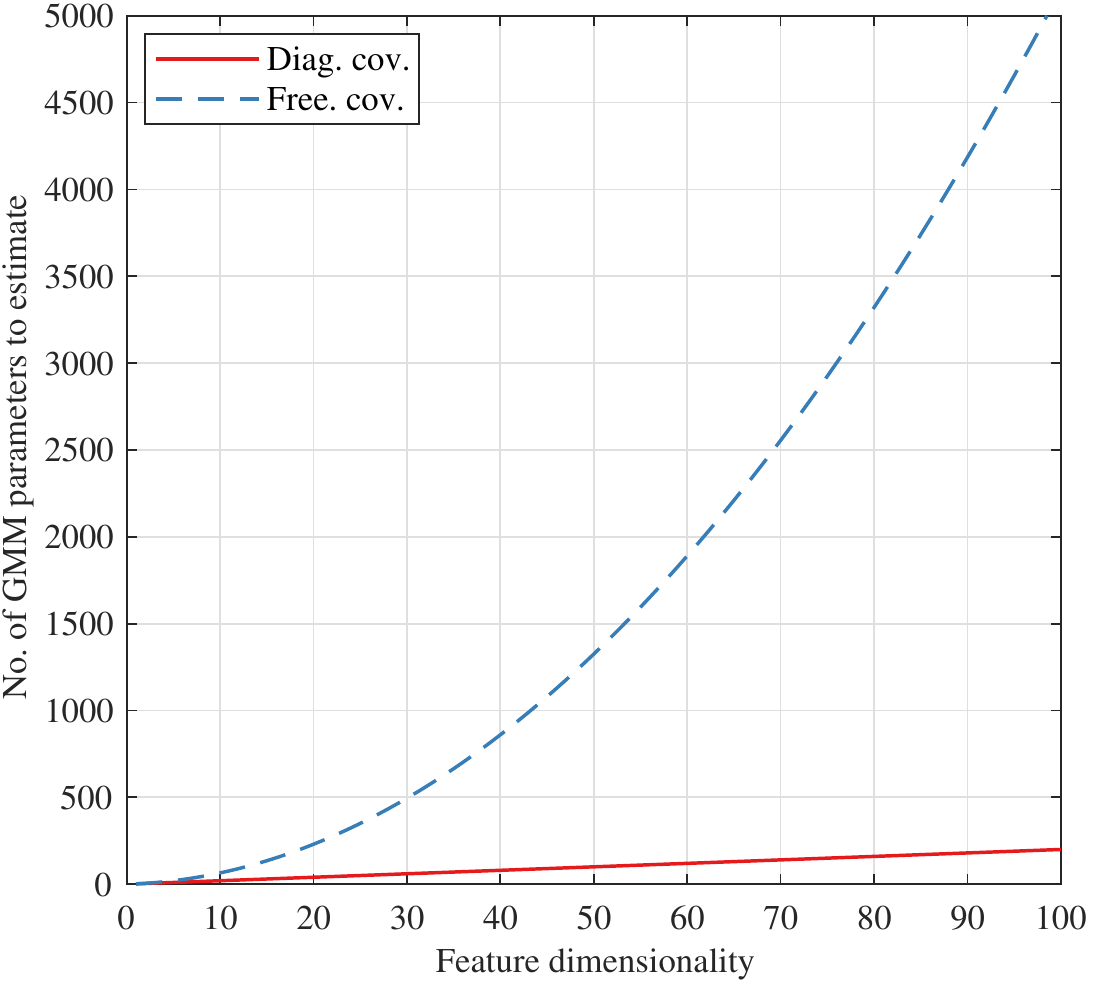}
	\caption{Necessary mixture parameters \textit{vs.} $d$ for $n=1$} \label{fig:cp-freeparams_data}
	\end{subfigure}\hfill%
	\begin{subfigure}[t]{0.5\textwidth}	\centering
	\includegraphics[height = 0.3\textheight]{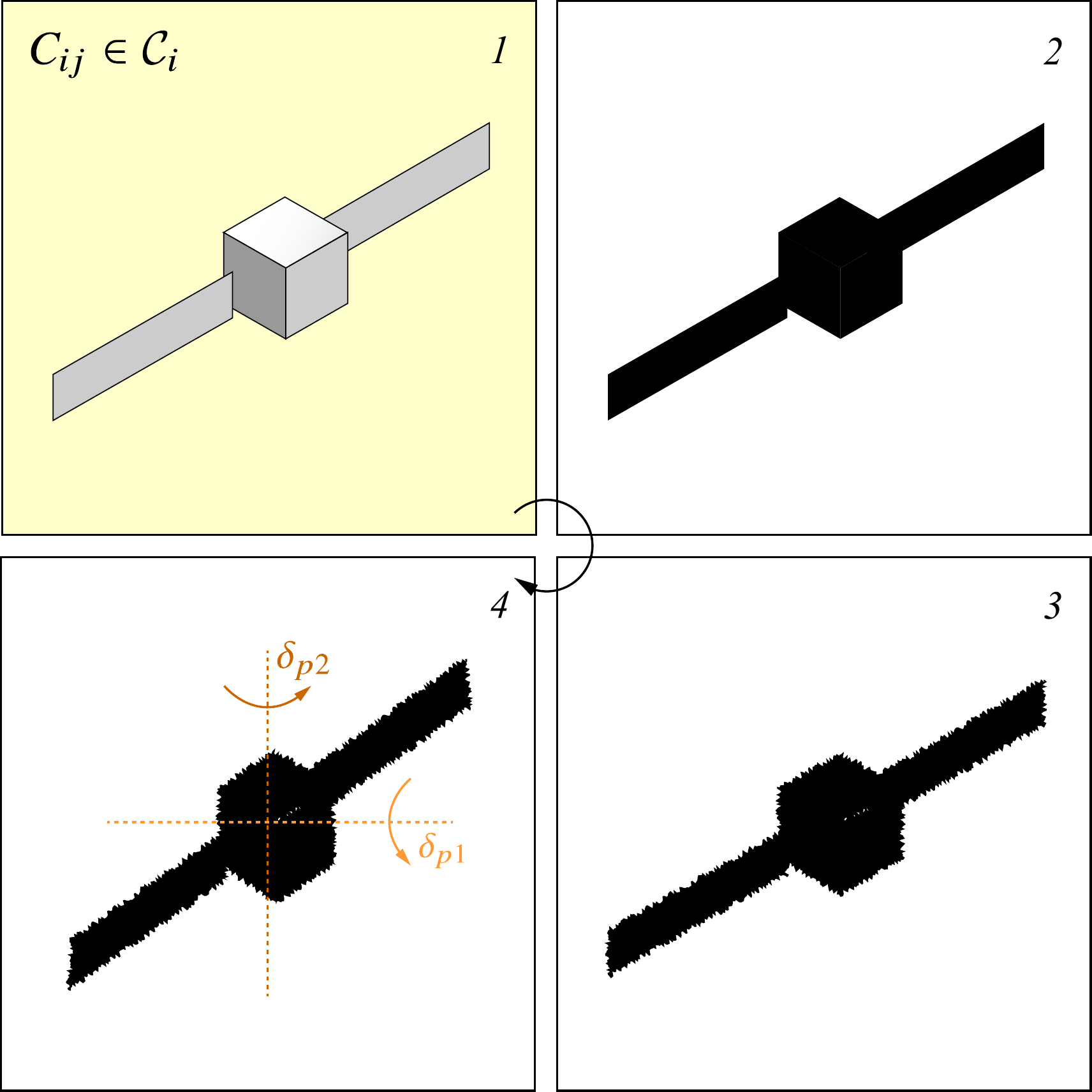}
	\caption{Artificial transformations on training data} \label{fig:cp-freeparams_trans}
	\end{subfigure}%
 \caption{Creating a training population.}
\label{fig:cp-freeparams}
\end{figure}

This section is concluded with some practical observations on the training procedure. The number of free parameters on a \gls{gmm} will depend on the dimensionality of the data $d$, on the number of mixture parameters $n$, and on the constraints placed on the covariance $\vect{\Sigma}$. A ``free'' covariance matrix will have ${1/2(d^2 + d)}$ independent elements, since it is symmetric, and hence the total number of mixture parameters will be ${(1/2 d^2 + 3/2 d + 1)n - 1}$. On the other hand, the covariance can be assumed as diagonal, in which case the total number of parameters to estimate becomes ${2nd - 1}$. Figure \ref{fig:cp-freeparams_data} plots the evolution of the number of parameters to estimate for a free covariance matrix and for a diagonal one in terms of the dimensionality of the features for $n=1$. It can be considered as a lower bound for the number of samples $m$ to be used in the training. The quadratic term in the free covariance case quickly diminishes the tractability of the problem when $d$ is increased, which can pose a problem when training data is limited. 

In \cite{li2008complex}, it was shown that the recognition power of complex \glspl{zm} for image retrieval begins to plateau beyond moments of the tenth order, which corresponds to approximately $d = 60$. This corresponds to 1890 parameters to be estimated for the free covariance case, while only 120 are necessary if a diagonal covariance is assumed. Since the \glspl{zm} are orthogonal, the correlation between moments is minimized and a diagonal covariance is an acceptable approximation. However, even if adjacent keyframes are grouped to form classes, the generated data might not be enough in terms of training. To this end, each keyframe $C_{ij}$ imaged for class $\mathcal{C}_i$ is subjected to a pipeline before the \glspl{zm} are computed and added to the training pool (Fig. \ref{fig:cp-freeparams_trans}). First, the target is segmented from the background and binarized. Then, a closed contour (the limb) can be defined. The contour is perturbed by changing the segmentation threshold to account for noise in the online images. Lastly, small perspective distortions are applied along the in-plane axes (i.e. the axes of $\rframe{P}$) to account for the fact that in the online pipeline the target is not always guaranteed to be imaged at the center. 

\section{Motion Estimation}
\label{sec:motionestimation}

\subsection{Problem Statement}
\label{sec:me-problem}

The problem of solving the 2D-3D point correspondences for the 6-\gls{dof} pose of a calibrated camera is termed \gls{pnp} and has a well-known closed form solution for $n = 3$ points (P3P). It relies on the fact that the angle between any image plane points $\vect{z}_i, \vect{z}_j$ must be the same as the angle defined between their corresponding world points $\vect{p}_i, \vect{p}_j$ \cite{szeliski2010computer}. Additional methods have been developed for $n \geq 4$, such as EP$n$P \cite{lepetit2009epnp}, which expresses the $n$ 3D points as a weighted sum of four virtual control points and then estimates the coordinates of these control points in the camera frame $\rframe{C}$. While relatively fast to compute, these methods are notwithstanding less robust to noise and fail in the presence of erroneous correspondences. On the other hand, iterative approaches that take these aspects into account, giving the best possible estimate of the pose under certain assumptions are often called the ``gold standard'' algorithm \cite{hartley2004multiple}. In this section, given an initial, coarse evaluation of the relative pose, an iterative refinement of its estimate based on nonlinear manifold parameterization is proposed. 

\begin{figure}[t]
\centering
\includegraphics[width=\textwidth]{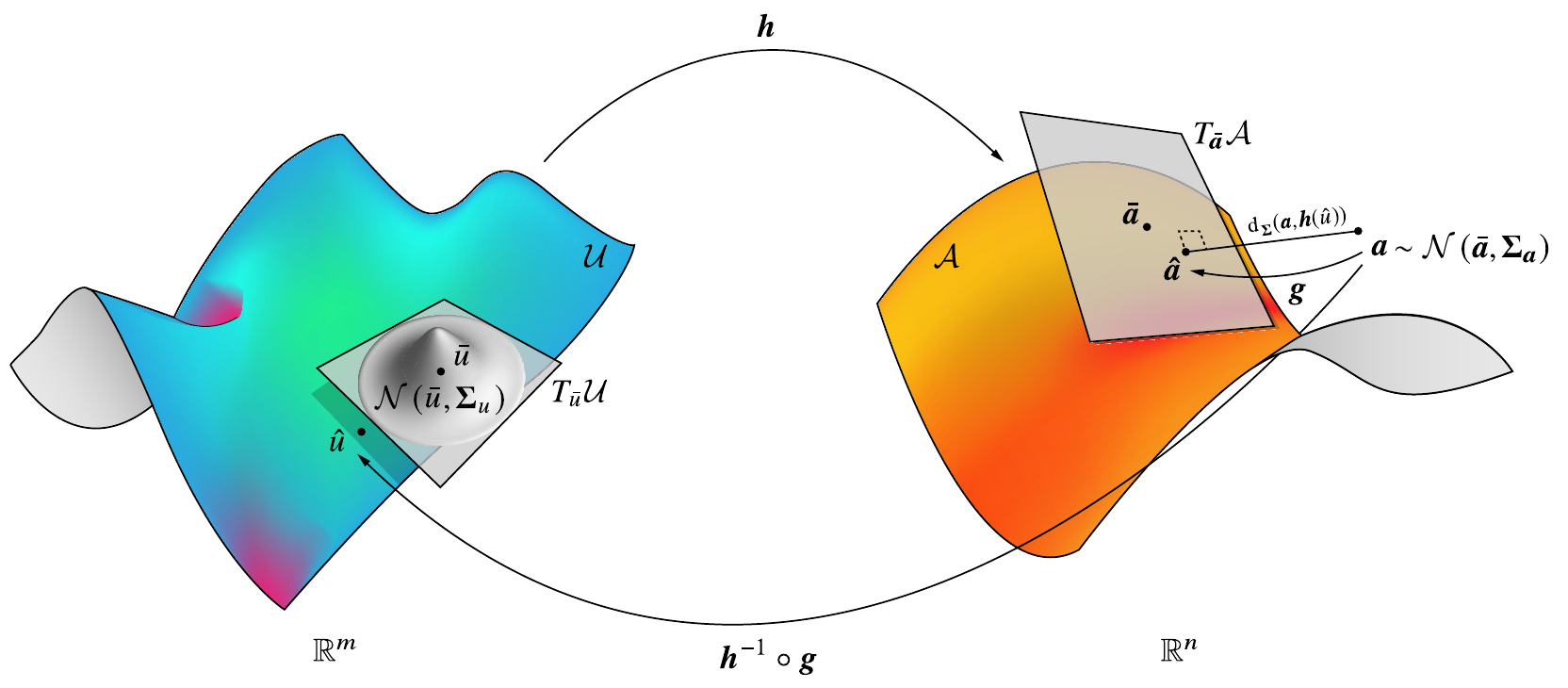}
\caption{Geometric correction and parametric fitting on manifolds.}
\label{fig:me-manifoldmin}
\end{figure}

Let $u$ represent the object to be determined. Its domain is a manifold ${\mathcal{U} \subset \mathbb{R}^m}$ which defines the parameter space. Let $\vect{a}$ be the vector of measurements in $\mathbb{R}^n$. Suppose $\vect{a}$ is observed in the presence of noise with a covariance matrix $\boldsymbol{\Sigma}_{\vect{a}}$, and let $\vectbar{a}$ be its true value, i.e. ${\vect{a} = \vectbar{a} + \Delta\vect{a}}$. Let ${\vect{h} \colon \mathbb{R}^m \to \mathbb{R}^n}$ be a mapping such that, in the absence of noise, ${\vect{h}(\bar{u}) = \vectbar{a}}$. Varying the value of $\bar{u}$ traces out a manifold ${\mathcal{A} \subset \mathbb{R}^n}$ defining the set of allowable measurements, i.e. the measurement space. The objective is, given a measurement $\vect{a}$, to find the vector ${\vecthat{a} \in \mathbb{R}^n}$ lying on $\mathcal{A}$ that is closest to $\vectbar{a}$ (Fig. \ref{fig:me-manifoldmin}).

Given the form of the multivariate normal probability density function, under the assumption of Gaussian noise, it is straightforward that the \acrfull{ml} solution is obtained by minimizing the Mahalanobis distance:

\begin{equation}
\urd_{\boldsymbol{\Sigma}_{\vect{a}}} \left(\vect{a}, \vect{h}(\hat{u})\right) \coloneqq \left(\left(\vect{a} - \vect{h}(\hat{u})\right)^\top \boldsymbol{\Sigma}_{\vect{a}}^{-1} \left(\vect{a} - \vect{h}(\hat{u})\right) \right)^{1/2}, \quad \text{with } \vect{h}(\hat{u}) = \vecthat{a}. \label{eq:me-mahalanobis}
\end{equation}
It is reasonable in most cases to assume that, in the neighborhood of $\vectbar{a}$, the surface of $\mathcal{A}$ is essentially planar and well approximated by the tangent space within the order of magnitude of the measurement noise variance \cite{hartley2004multiple}. Then, the \gls{ml} corrected measurement $\vecthat{a}$ is the foot of the perpendicular from $\vect{a}$ onto the tangent plane. The benefit of this approximation is that it allows the measurement residual error to be modelled as a Gaussian distribution in the normal space of $\mathcal{A}$, whereas the measurement estimation error is a Gaussian distribution in the tangent space $T_{\vectbar{a}}\mathcal{A}$. In computer vision it is common to have image or world points as measured variables, so typically one can safely write the estimation error as ${\vecthat{a} - \vectbar{a}}$. Analogously, the parameter estimation error ${\delta \vect{u}}$ is, to a first approximation, constrained to be in the tangent space $T_{\bar{u}}\mathcal{U}$. In general terms, the present work shall assume this local distribution approximation is valid for small errors whenever dealing with the probability distribution of a variable constrained to a manifold \cite{kanatani1996statistical}. Let ${\vect{g}\colon \mathbb{R}^n \to \mathcal{A}}$ map a point to the surface of the measurement space, as defined in Eq.  (\ref{eq:me-mahalanobis}). Assuming that $\vect{h}$ is invertible such that ${\vect{h}^{-1}\colon \mathcal{A} \to \mathbb{R}^m}$, then the mapping ${\vect{h}^{-1} \circ \vect{g}}$ can be used to propagate the measurement noise covariance $\boldsymbol{\Sigma}_{\vect{a}}$ to obtain the covariance matrix of the \gls{ml} estimate $\bar{u}$.

Let ${\varphi(u) = \urd_{\boldsymbol{\Sigma}_{\vect{a}}}^2(\vect{a}, \vect{h}(u))} $ for succinctness. Then, the problem can be posed as:
\begin{equation}
\hat{u} = \argmin_{u \in \mathcal{U}} \varphi(u) \label{eq:me-leastsquares}.
\end{equation}
When the function $\vect{h}$ is nonlinear, Eq. (\ref{eq:me-leastsquares}) may be solved iteratively by linearizing it around a reference parameter $\varphi(u_0)$. In the case where the parameter space $\mathcal{U}$ can be identified with Euclidean space, linearizing and differentiating $\varphi(\vect{u})$ at $\vect{u}_0$, ${\vect{u} \in \mathbb{R}^m}$, leads to the well-known normal equations yielding a correction $\Delta\vect{u}$ at iteration ${t = k}$ such that ${\vecthat{u}_{k+1} = \vecthat{u}_{k} + \Delta\vect{u}}$. If not, this update is not valid: as noted in Section \ref{sec:mp}, $\vect{u}_{k+1}$ is not guaranteed to be an element of $\mathcal{U}$.

One possible solution is to nevertheless apply the correction via standard addition and then project the result back to the parameter manifold $\mathcal{U}$, which could introduce additional noise in the system and drive the result away from the \gls{ml} estimate $\hat{u}$ . A more elegant alternative solution is to exploit the local Euclidean structure of $\mathcal{U}$ around $u_0$ to generate a new set of normal equations. Taking ${\mathcal{U} \cong \sethree}$ and using the composition operator from Eq. (\ref{eq:mp-posecomposition}), linearizing $\varphi(u)$ yields:
\begin{equation}
\varphi(u) \approx \varphi(u_0) + \left.(u \ominus u_0)^\top \nabla \varphi\right\rvert_{u \ominus u_0 = 0}. \label{eq:me-taylorse3}
\end{equation}
Equation (\ref{eq:me-taylorse3}) thus motivates working with the pose estimation error ${\delta\vect{u} = u\ominus u_0}$ explicitly, which is an element of $\setthree$. This can be shown to lead to the normal equations of the form:
\begin{equation}
\vect{J}^\top \vect{\Sigma}_{\vect{a}} \vect{J} \delta\vect{u} = -\vect{J}^\top \vect{\Sigma}_{\vect{a}} \vect{r}, \label{eq:me-normalse3_1}
\end{equation}
where ${\vect{r} = \vect{h}(u_0) - \vect{a}}$ is the residual vector. The other advantage of Eq. (\ref{eq:me-normalse3_1}) is that the Jacobian matrix ${\vect{J} \coloneq \frac{\partial h(u_0 \oplus \delta\vect{u})}{\partial (\delta\vect{u})}\rvert_{\delta\vect{u} = \vect{0}}}$ is computed with respect to the basis of $\setthree$. At the end of each iteration, the updated parameter is obtained via the exponential map by following Eq. (\ref{eq:mp-posecomposition}), thus ensuring it naturally remains an element of $\mathcal{U}$.

\subsection{Structural Model Constraints}
\label{sec:me-structconst}

\subsubsection{From Visual Point Feature Correspondences}

This subsection explains how to find a relationship between some pre-existing knowledge of the target's structure and measurements taken of it with a digital camera that allows for the relative pose to be estimated in accordance with the theory developed above. Note that Eq. (\ref{eq:mp-reprojection}) describes such a relationship, as $\vect{z}$ is the reprojection in the image plane of a 3D point $\vect{p}$ defined in $\rframe{T}$. Therefore, given a number of $m$ correspondences ${\vect{z}_i \leftrightarrow \vect{p}_i}$ between 2D image points and 3D structural points, the task is to find $\vect{T}$ such that ${\vect{z}_i = \vect{\pi}(\vect{K}\vect{T}\otimes\vect{p}_i)}$, for all ${i = 1,\ldots,m}$. Of course, the relative pose does not have to be represented in the homogeneous form $\vect{T}$; the problem is simply posited as such because of the significance of Eq. (\ref{eq:mp-reprojection}) in the computer literature and, as shall be seen, because it leads to a simple form of the Jacobian. 

The obvious difficulty in this formulation lies in solving the feature correspondence problem due to the topological difference between $\vect{z}_i$ and $\vect{p}_i$. The proposed work solves this difference by attributing to each structural point a representation on the image plane in an offline training stage. This is achieved as follows: in each model keyframe, point features (keypoints) are selected by a detector algorithm as centers of regions of interest (blobs) in the image. These regions are typically deemed ``interesting'' when they have a defining property (e.g. brightness) that differs from their surroundings. Once detected, a keypoint can then be extracted by applying a descriptor algorithm which will encode its characterizing traits into a vector. This descriptor vector normally incorporates information about the blob as well, allowing keypoints to be matched in different images of the same object in a robust manner with respect to changes in scale, orientation, and brightness, among others. Since the relative pose is known for each keyframe, Eq. (\ref{eq:mp-reprojection}) is inverted to generate a ray passing through each keypoint. The depth of the ray is the image of the keypoint in the depth map corresponding to that keyframe, thus determining the equivalent 3D structural point. In this way, each $\vect{p}_i$ is annotated offline with a 2D descriptor computed from the reprojection $\vect{z}^{\vect{\prime}}_i$ in the current keyframe. Then, computing a descriptor vector for the keypoints detected online $\vect{z}_i$ grants the equivalence ${ \{\vect{z}_i \leftrightarrow \vect{p}_i\} \Leftrightarrow \{\vect{z}_i \leftrightarrow \vect{z}_i^{\vect{\prime}} \}}$, reducing a 3D-2D correspondence problem to a 2D-2D one.

Since the structural points $\vect{p}_i$ are obtained via ground truth depth maps for keyframes with perfectly known $\vect{T}$, they are considered to be measured with maximum accuracy, and the error is thus concentrated in the measured image points $\vect{z}_i$. In other words, the measurement space $\mathcal{A}$ is a manifold embedded in $\mathbb{R}^{2m}$ (i.e. $\vect{a}$ is construed by stacking the $x$ and $y$ components of all $\vect{z}_i)$ and the parameter space $\mathcal{U}$ is 6-dimensional (i.e. the dimensions of $\setthree$).  Furthermore, as each measured image point is obtained algorithmically with the same feature detector, each $\vect{z}_i$ is modelled as a random variable sampled from an isotropic (Gaussian) distribution. The \gls{ml} estimate of the pose is in this manner obtained by minimizing the geometric error (cf. Eq. (\ref{eq:me-mahalanobis})) which is reduced to the standard squared Euclidean distance:

\begin{subequations}
\begin{equation}
    \vecthat{T} = \argmin_{\vect{T} \in \sethree} \sum\limits_{i=1}^m \urd \left(\vect{z}_i, \vect{\pi}\left(\vect{K}\vect{T}\otimes\vect{p}_i\right)\right)^2. \label{eq:me-objfunstruct}
\end{equation}

\noindent The solution to Eq. (\ref{eq:me-objfunstruct}) is found iteratively via \gls{lm} with the Jacobian matrix given by:

\begin{align}
    \vect{J}^s_{p} = \left.\frac{\partial \vect{\pi}(\vect{K}(\vect{T} \oplus \vect{\varepsilon})\otimes\vect{p} )}{\partial \vect{\varepsilon}}\right\rvert_{\vect{\varepsilon} = \vect{0}}
    &= \left.\frac{\partial \vect{\pi'}(\vect{p'})}{\partial \vect{p'}}\right\rvert_{\substack{\vect{\pi'}(\vect{p'}) \coloneqq \vect{\pi}(\vect{K}\vect{p'})\\ \vect{p'} = \vect{T}\otimes\vect{p}\hfill{}}} \left.\frac{\partial \vect{T'}\otimes\vect{p}}{\partial \vect{T'}}\right\vert_{\vect{T'}=\vect{T}\oplus\vect{\varepsilon}=\vect{T}} \left.\frac{\partial \exp(\vect{\varepsilon})\vect{T}}{\partial\vect{\varepsilon}}\right\rvert_{\vect{\varepsilon}=\vect{0}} \nonumber\\
    &= \begin{bmatrix}
    \frac{f_x}{p'_3}  & 0 & -f_x\frac{p'_1}{p^{\prime 2}_3} & -f_x\frac{p'_1 p'_2}{p^{\prime 2}_3} & f_x\left(1 + \frac{p^{\prime 2}_1}{p^{\prime 2}_3}\right) & - f_x\frac{p'_2}{p'_3}\\
    0 & \frac{f_y}{p'_3} & -f_y\frac{p'_2}{p^{\prime 2}_3} & -f_y\left(1 + \frac{p^{\prime 2}_2}{p^{\prime 2}_3} \right) & f_y\frac{p'_1 p'_2}{p^{\prime 2}_3} & f_y \frac{p'_1}{p'_3}
    \end{bmatrix}, \label{eq:me-jacstructline}
\end{align}
\end{subequations}
where $f_x,f_y$ is the focal length and $\vect{\varepsilon} \in \setthree$ is a small perturbation of the pose.

\subsubsection{From Visual Edge Feature Correspondences}

The structural model constraints may also be formulated in terms of different types of features, such as straight line segments. This is likewise an important element to consider in space relative navigation, as spacecraft often resemble cuboid shapes or are composed of elements shaped as such; therefore it is expected to have detectable line features (keylines) when imaging this kind of targets. It has been shown in this context that, while keypoints are more distinctive in the context of minimizing Eq. (\ref{eq:me-objfunstruct}), keylines are actually more robust in terms of preventing the solution from diverging \cite{rondao2018multiview}.

In two dimensions, a point $\vect{z}$ lies on a line ${\vect{l} = (l_1,l_2,l_3)^\top}$ if ${\vect{z}^\top \vect{l} = \vect{0}}$. Assume there exist $m$ correspondences $\vect{l}_i \leftrightarrow \vect{\ell}_i$ between the 2D lines and 3D lines. Therefore, one can formulate a geometric distance for 2D-3D line correspondences in terms of the reprojection of a point $\vect{p}_{ij} \in \vect{\ell}_i$ onto the image plane:

\begin{equation}
    \vecthat{T} = \argmin_{\vect{T} \in \sethree} \sum\limits_{i=1}^m \sum\limits_{j=0}^k \vect{l}_i^\top \vect{K} \vect{T} \otimes \vect{p}_{ij}, \label{eq:me-objfunstructline}
\end{equation}

\noindent The Jacobian matrix $\vect{J}^s_{l}$ corresponding to the minimization of Eq. (\ref{eq:me-objfunstructline}) can be derived in a similar manner to Eq. (\ref{eq:me-jacstructline}).

In practice, matching keylines is not as straightforward as matching keypoints, as the former are typically less distinctive than the latter. For the scope of this work, only the contour of the target is considered, which is discretized into a finite number of edge points that are assumed to belong to a (straight) keyline. Additionally, edge points can be registered in the same way as structural keypoints through the use of depth maps.

\subsection{Local Feature Processing}
\label{sec:me-lfp}

\subsubsection{Detection}

Distinct point features are identified in an image of the target using the \gls{orb} detector \cite{rublee2011orb}. The basis of \gls{orb} is the \gls{fast} algorithm \cite{rosten2006machine}, developed with the purpose of creating a high-speed keypoint finder for real-time applications\footnote{Although \gls{orb} has also been developed for feature description, using a modification of the \gls{brief} algorithm, it is applied herein for detection only.}. It first selects a pixel $\vect{z}_i$ in the image as candidate. A circle of 16 pixels around $\vect{z}_i$ and a threshold $\alpha$ are defined. If there exists a set of $n$ contiguous pixels in the circle which are all brighter than $I(\vect{z}_i) + \alpha$ or all darker than $I(\vect{z}_i) - \alpha$, then $\vect{z}_i$ is classified as a keypoint. The algorithm is made robust with an offline machine learning stage, training it to ignore regions in an image where it typically lacks interest points, thus improving detection speed. As the original method is not robust to changes in size or rotation, \gls{orb} applies a pyramidal representation of \gls{fast} for multi-scale feature detection and assigns an orientation to each one by defining a vector from its origin to the intensity baricenter of its support region. The choice for the keypoint algorithms (see also Section \ref{subsec:me-lfp-descirption}) is the product of the authors' previous survey on \gls{ip} techniques for relative navigation in space \cite{rondao2018multispectral}.

The Canny algorithm is the basis for edge detection \cite{canny1987computational}. First, the image is filtered with a Gaussian kernel in order to remove noise. Secondly, the intensity gradients at each pixel are computed. Then, non-maximal suppression and a double thresholding are applied to discard spurious responses and identify edge candidates. The method from Ref. \cite{lee2014outdoor} is used to efficiently extract keylines from the edge image by incrementally connecting edge pixels in straight lines and merging those with small enough differences in overlap and orientation.

\subsubsection{Description}
\label{subsec:me-lfp-descirption}

For each detected keypoint, a binary string is generated encoding information about its support region using the \gls{freak} descriptor \cite{alahi2012freak}, which takes inspiration in the design of the human retina. The method adopts the retinal sampling grid as the sampling pattern for pixel intensity comparisons, i.e. a circular design with decreasing density from the center outwards, achieved using different kernel sizes for the Gaussian smoothing of every sample point in each receptive field; these overlap for added redundancy leading to increased discriminating power. Each bit in the descriptor thus represents the result of each comparison test, which only has two possible outcomes: either the first pixel is brighter than the second, or vice-versa. A coarse-to-fine pair selection is employed to maximize variance and uncorrelation between pairs. In this way, the first 16 bytes of the descriptor represent coarse information, which is applied as a triage in the matching process, and a cascade of comparisons is performed to accelerate the procedure even further.

\subsubsection{Brute-Force Detection Matching} 

In an initial stage, the features are matched using brute force, since no estimate of the pose is yet available. In the case of the point features, this implies that all those detected in the initial frame are compared against those in the train keyframe. This is achieved by computing the Hamming distance $\urd_\mathrm{Ham}(\cdot, \cdot)$ between their corresponding descriptors, i.e. the minimum number of substitutions required to convert one into the other. It can be swiftly computed by applying the exclusive-OR (XOR) operator followed by a bit count, which provides an advantage in terms of computational performance with respect to the Euclidean distance test used with more traditional floating point descriptors. For each query, the two closest train descriptor matches are selected and subjected to a \gls{nndr} test: the matching of the descriptors $\vect{s}_i$ and $\vect{s}_j$ is accepted if
\begin{equation}
\frac{\urd_\mathrm{Ham}(\vect{s}_i, \vect{s}_j)}{\urd_\mathrm{Ham}(\vect{s}_i, \vect{s}_k)} < \mu_\mathrm{NNDR},
\end{equation}
where $\vect{s}_j, \vect{s}_k$ are the 1\textsuperscript{st} and 2\textsuperscript{nd} nearest neighbours to $\mathbf{s}_i$ and $\mu_\mathrm{NNDR}$ is a ratio from 0 to 1.

As descriptors for edge features are not employed, an alternative strategy was devised to match them. The full contours $\mathcal{D}_q, \mathcal{D}_t \subset \mathbb{R}^2$ of the target in the query image and the train keyframe, respectively, are considered. Each contour is a set of $n$ discretized edge points of the same size, i.e. ${\mathcal{D}_q = \{\vect{z}_{q,1}, \ldots, \vect{z}_{q,n}\}}$ and ${\mathcal{D}_t = \{\vect{z}_{t,1}, \ldots, \vect{z}_{t,n}\}}$. Even though the query image and train keyframe represent the same aspect of the target, there will be differences that are reflected on the contours. In particular, $\mathcal{D}_q$ and $\mathcal{D}_t$ will be different by a 2D affine transformation $\Lambda(\beta,\theta,\vect{t}):\mathcal{D}_t \rightarrow \mathcal{D}_q$, where $\beta>0$ is a scaling factor, $\theta \in [-\pi, \pi[$ \si{\radian} is an angle of rotation and $\vect{t} = \icolsmall{t_1,t_2}^\top$ is a translation vector. The contour alignment problem is posed in the least squares sense as 
\begin{equation}
\argmin_{\vect{t},\beta,\theta} \urd_\mathrm{Fro} \left( \mathcal{D}_q, \Lambda\left(\mathcal{D}_t\right) \right), \label{eq:me-nl_cont_align}
\end{equation}
where $\urd_\mathrm{Fro}(\cdot, \cdot)$ is the Frobenius distance. Because of the multiplicative trigonometric terms of $\Lambda$, Eq. (\ref{eq:me-nl_cont_align}) is nonlinear. However, the problem can be converted into an equivalent linear one by a change of variables \cite{markovsky2008least}:

\begin{equation}
    \argmin_{\icolsmall{t_1,t_2,b_1,b_2}\in\mathbb{R}^4} \urd \left( 
    \icol{
    z_{q,1,1}\\
    z_{q,1,2}\\
    \vdots\\
    z_{q,n,1}\\
    z_{q,n,2}
    },
    \begin{bmatrix}
    1 & 0 & z_{t,1,1} & -z_{t,1,2}\\
    0 & 1 & z_{t,1,2} & z_{t,1,1}\\
    \vdots & \vdots & \vdots & \vdots\\
    1 & 0 & z_{t,n,1} & -z_{t,n,2}\\
    0 & 1 & z_{t,n,2} & z_{t,n,1}\\
    \end{bmatrix} 
    \icol{
    t_1 \\ t_2 \\ b_1 \\ b_2
    }
    \right)
 \, \text{with} \, 
    \icol{
    b_1\\
    b_2
    } = 
    \beta \icol{
    \cos{\theta}\\
    \sin{\theta}
    } \Leftrightarrow 
    \icol{
    \theta\\
    \beta
    } = 
    \icol{
    \arcsin{\left(b_2/\sqrt{b_1^2 + b_2^2}\right)}\\
    \sqrt{b_1^2 + b_2^2}
    }, \label{eq:mp-me_l_cont_align}
\end{equation}

\noindent where $\vect{z}_{u,v} = \icolsmall{z_{u,v,1},z_{u,v,2}}^\top$. In this way, a global solution of the minimum can be calculated using standard linear algebra. However, Eq. (\ref{eq:mp-me_l_cont_align}) depends on the correspondences between the query and train edge points, which are not known a priori. To simultaneously solve for the edge point correspondence problem and contour alignment, the algorithm is modified by solving $n$ linear least squares problems, each time shifting the order of the edge points in $\mathcal{D}^t$ by one, and selecting the minimum of the $n$ residual norms. Thus, the only necessary inputs are two sets of sequential but not necessarily correspondent edge points. 

\subsubsection{Predictive Tracking Matching}

Once the algorithm is initialized, knowledge of the current solution can be used to improve the performance of the feature matching processes. In particular, the predicted estimate of the pose output by the filtering module is used to help anticipate where the features will be located in the next frame in time, in this way introducing a temporal tracking constraint that improves the pose estimation accuracy. 

In the case of point features, tracking matching is achieved by fitting a grid of $p\times q$ cells on the boundary of the target in the query camera image. The detected keypoins are binned into the resulting cells. Then, the 3D structural points of the currently selected database keyframe are reprojected onto the query image according to the predicted pose (cf. Eq. (\ref{eq:mp-reprojection})) and equally binned according to the grid. Lastly, descriptor-based matching is applied on a per-cell basis, vastly reducing the number of possible matching candidates. This step was found essential in order to maintain the accuracy of the algorithm during sequences where ambiguous modules are imaged (e.g. \gls{mli}) or when the query image is too distinct from the train one (e.g. due to reflections).

In the case of edge features, tracking matching is done by first detecting keylines on the query edge image. Each query keyline is then drawn on the image plane with a unique color. The 3D edge points and corresponding keylines from the train keyframe are reprojected onto the image plane. Then, the matching algorithm iterates over each reprojected edge point and a 1D search is performed perpendicularly to it according to the corresponding keyline, obtained in the offline training stage, until the closest colored pixel is found. Hence, 3D edge points are matched to 2D keylines satisfying the conditions to minimize Eq. (\ref{eq:me-objfunstructline}).

\subsection{Robust Estimation}
\label{sec:me-re}

\begin{figure}[t]
	\centering
	\begin{subfigure}[t]{0.33\textwidth}	\centering
	\includegraphics[width=0.95\textwidth]{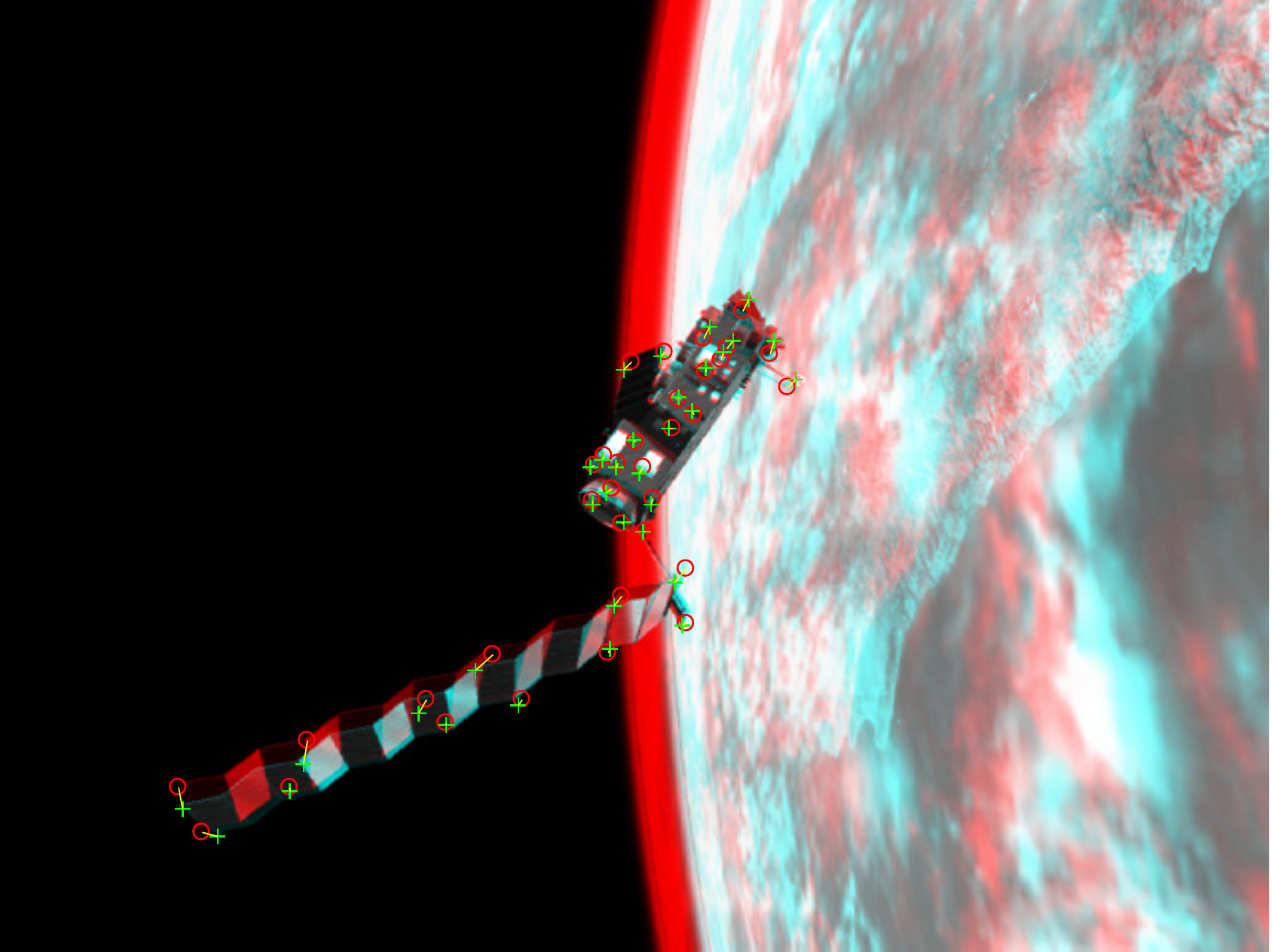}
	\caption{0\% outliers}
	\end{subfigure}%
		\begin{subfigure}[t]{0.33\textwidth}	\centering
	\includegraphics[width=0.95\textwidth]{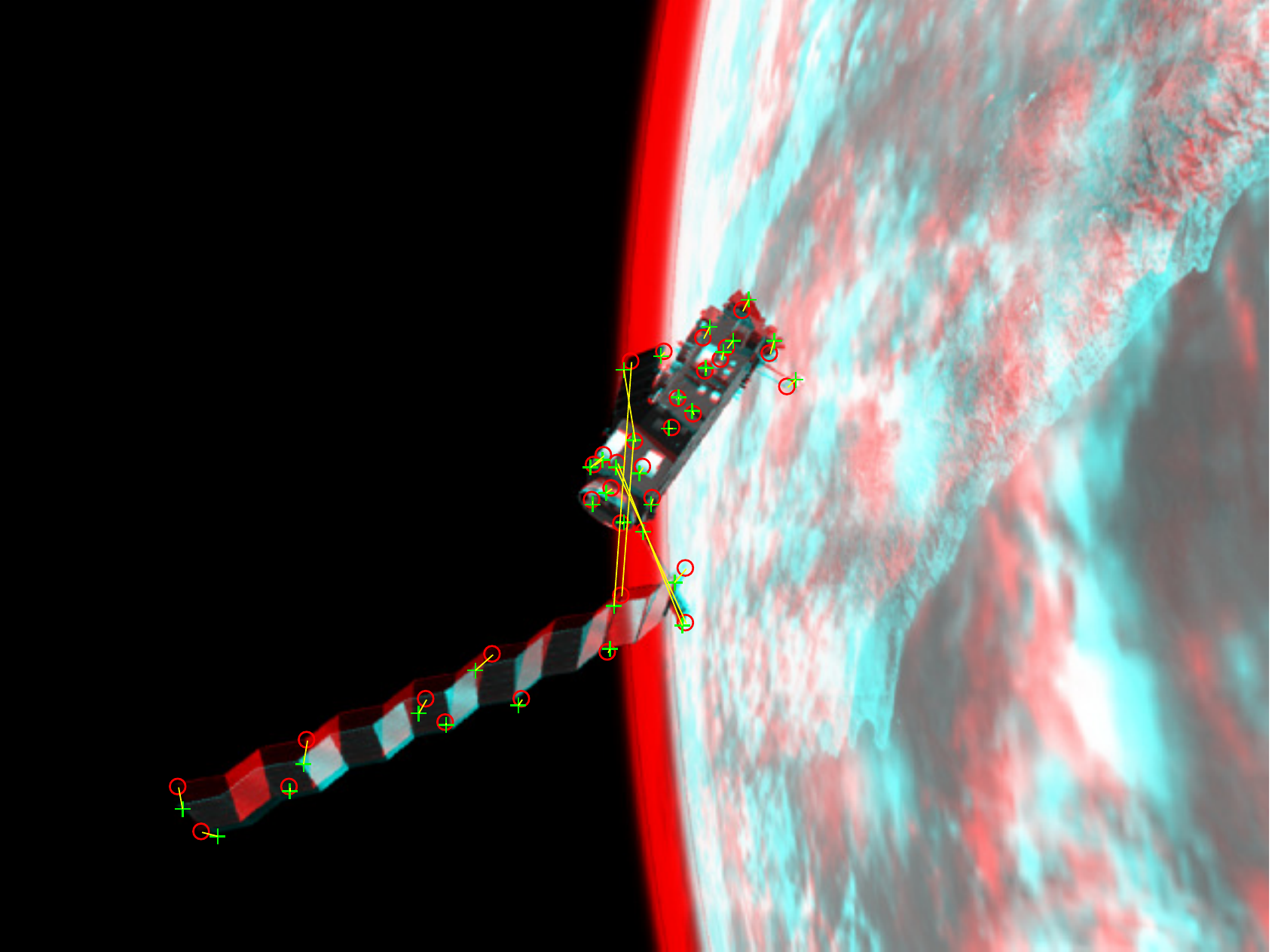}
	\caption{25\% outliers}
	\end{subfigure}%
		\begin{subfigure}[t]{0.33\textwidth}	\centering
	\includegraphics[width=0.95\textwidth]{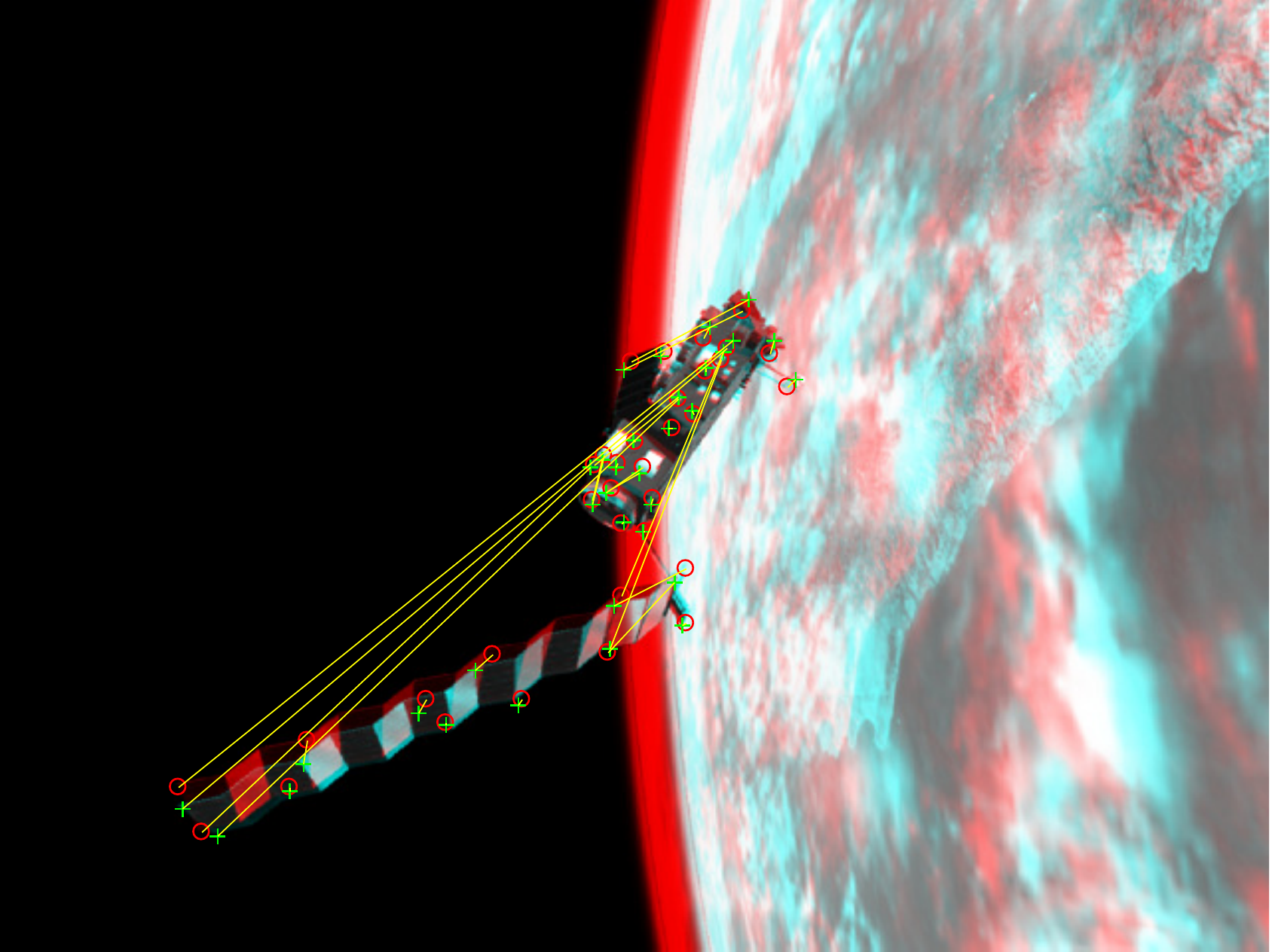}
	\caption{50\% outliers}
	\end{subfigure}
  \caption{Sequential feature matching in false-color composite overlay.}
  \label{fig:me-outliers}
\end{figure}

When the measurements are assumed to have equal variance, the \gls{ml} estimate is found by solving

\begin{equation}
\hat{u} = \argmin_{u \in \mathcal{U}} \vect{r}^\top\vect{r},   \label{eq:me-ls}
\end{equation}

\noindent where $\vect{r} \coloneqq \urd(\vect{a}, \vect{h}(u))$ is the residual vector. The problem becomes one of classical \gls{ls} estimation, where the covariance $\vect{\Sigma}_{\vect{a}}$ vanishes as Eq. (\ref{eq:me-ls}) is equivariant with respect to scale. However, ordinary \gls{ls} is not robust to outliers, i.e. spurious data that may contaminate the measurements. In the scope of this work, measurements are matches between features, which can be erroneous due to the typical space rendezvous scenario as imaged by a camera. For instance, a solar panel might resemble a repeating pattern that yields many features which look identical, or intense illumination from the Sun acting on the spacecraft can change its local aspect with respect to a model image. Consider Fig. \ref{fig:me-outliers}, where sequential frames of a simulated rendezvous sequence are represented with a false-color overlay. Point feature matches are also represented and connected by lines for several levels of outlier contamination. Whereas, from the reader's perspective, it may seem that for a \SI{25}{\percent} outlier level (middle image) the true trajectory can still be determined, in theory the presence of a single outlier is enough to make the \gls{ls} estimate diverge \cite{rousseeuw2005robust}.

 Robustness with respect to outliers can be achieved by generalizing Eq. (\ref{eq:me-ls}) into an M-estimator:
\begin{equation}
\hat{u} = \argmin_{u \in \mathcal{U}} \sum\limits_{i = 1}^N\rho\left(\frac{r_i}{\hat{\sigma}}\right),   \label{eq:me-mestimator}
\end{equation}
\noindent where $\rho$ is a symmetric, positive-definite function with subquadratic growth, and $\hat{\sigma}^2$ is an estimate of the variance, or scale, of $\vect{r}$. Solving Eq. (\ref{eq:me-mestimator}) implies
\begin{equation}
\sum\limits_{i = 1}^N\psi\left(\frac{r_i}{\hat{\sigma}}\right)\frac{\urd r_i}{\urd u}\frac{1}{\hat{\sigma}} = 0,
\end{equation}
\noindent where ${\psi(x) \coloneqq \urd \rho(x)/\urd x}$ is defined as the influence function of the M-estimator. This function measures the influence that a data point has on the estimation of the parameter $u$. A robust M-estimator $\rho(x)$ should meet two constraints: convexity in $x$, and a bounded influence function \cite{zhang1997parameter}. By acknowledging the latter point, it becomes clear why the general \gls{ls} is not robust, since ${\rho(x) = x^2/2}$ and therefore ${\psi(x)=x}$.

There are two possible approaches to define the normal equations for M-estimation that avoid the computation of the Hessian \cite{holland1977robust}:

\begin{subequations}
\begin{align}
\vect{J}^\top \vect{J} \delta\vect{u} &= -\vect{J}^\top \vect{\psi}\left(\frac{\vect{r}}{\hat{\sigma}}\right)\hat{\sigma},\label{eq:me-normeqshuber} \\
\vect{J}^\top \vect{W} \vect{J} \delta\vect{u} &= -\vect{J}^\top \vect{W}\vect{r}, \label{eq:me-normeqsirls}
\end{align}
\end{subequations}

\noindent where $\vect{W} = \diag(w(r_1/\hat{\sigma}),\ldots,w(r_n/\hat{\sigma}))$ and $w(x) \coloneqq \psi(x)/x$. The first method was developed by Huber \cite{huber1977robust} and generalizes the normal equations through the modification of the residuals via $\psi$ and $\hat{\sigma}$. Huber proposed a specific loss function, the Huber M-estimator $\rho_\text{Hub}(x)$. Huber's algorithm provides a way to jointly estimate the scale $\sigma$ alongside the parameter $u$ with proven convergence properties. The minimization algorithm (e.g. \gls{lm}) is simply appended with the procedure:
\begin{equation}
\sigma^2_{k+1} = \frac{1}{(n-p)\beta}\sum\limits_i^n \left(\frac{r_i}{\sigma_k}\right)^2 \sigma^2_k,
\end{equation}
\noindent where $\beta$ is a bias-correcting factor. The second method was developed by Beaton and Tukey \cite{beaton1974fitting} and is commonly known as \gls{irls}, due to the inclusion of the weights matrix $\vect{W}$ that assumes the role of $\vect{\Sigma}_{\vect{a}}$ (cf. Eq. (\ref{eq:me-normalse3_1})). Tukey proposed an alternative robust loss function, $\rho_\text{Tuk}(x)$.

Each robust loss function, $\rho_\text{Hub}(x)$ and $\rho_\text{Tuk}(x)$, can be compared regardless of the formulation. The Huber M-estimator is considered to be adequate for almost all situations, but does not eliminate completely the influence of large errors \cite{zhang1997parameter}. On the other hand, the Tukey M-estimator is non-convex, but is a ``hard redescender'', meaning that its influence function tends to zero quickly so as to aggressively reject outliers, explaining its frequent use in computer vision applications, where the outliers typically have small residual magnitudes \cite{stewart1999robust}. 

\begin{figure}[]
	\centering
	\begin{subfigure}[t]{0.33\textwidth} 	\centering
	\includegraphics[width=\linewidth]{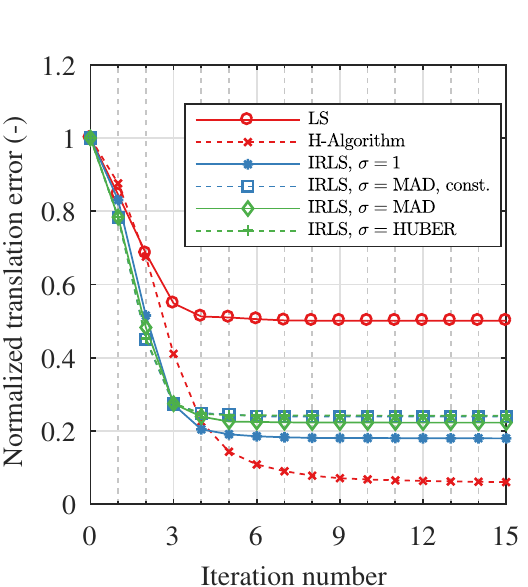}
	\end{subfigure}\hfill%
		\begin{subfigure}[t]{0.33\textwidth} 	\centering
	\includegraphics[width=\linewidth]{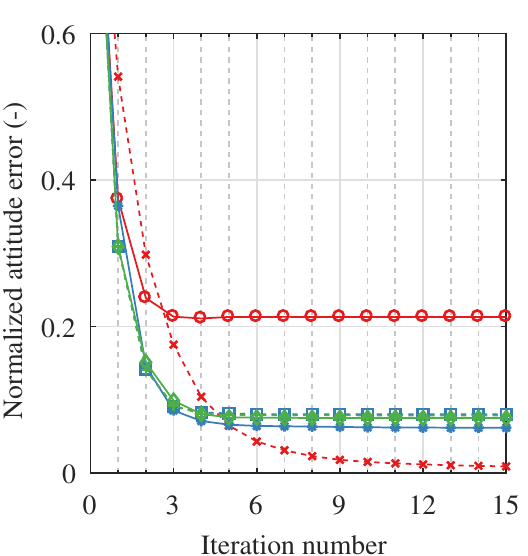}
	\caption{10\% outliers}
	\end{subfigure}\hfill%
		\begin{subfigure}[t]{0.33\textwidth} 	\centering
	\includegraphics[width=\linewidth]{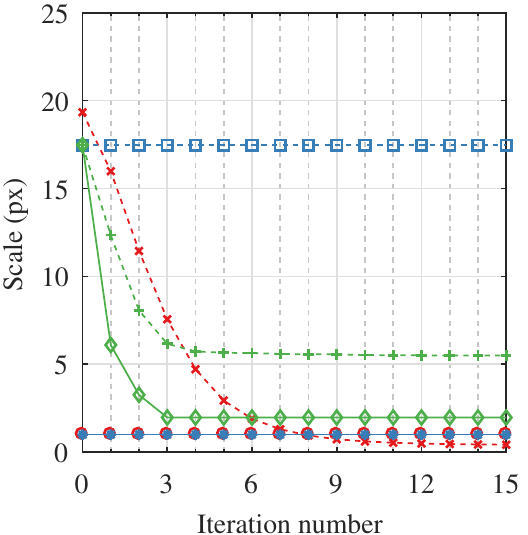}
	\end{subfigure}\\
		\begin{subfigure}[t]{0.33\textwidth} 	\centering
	\includegraphics[width=\linewidth]{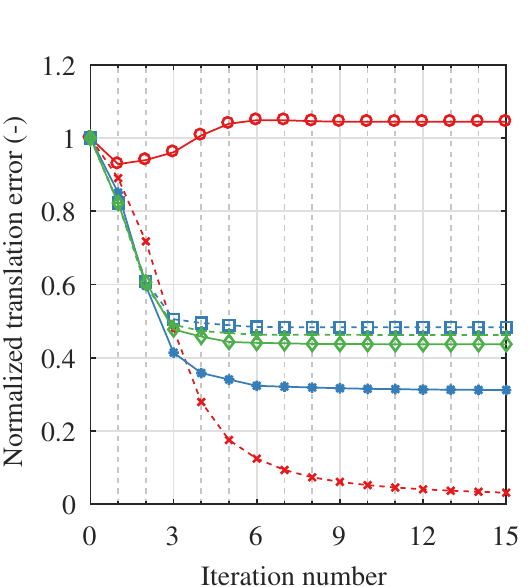}
	\end{subfigure}\hfill%
		\begin{subfigure}[t]{0.33\textwidth} 	\centering
	\includegraphics[width=\linewidth]{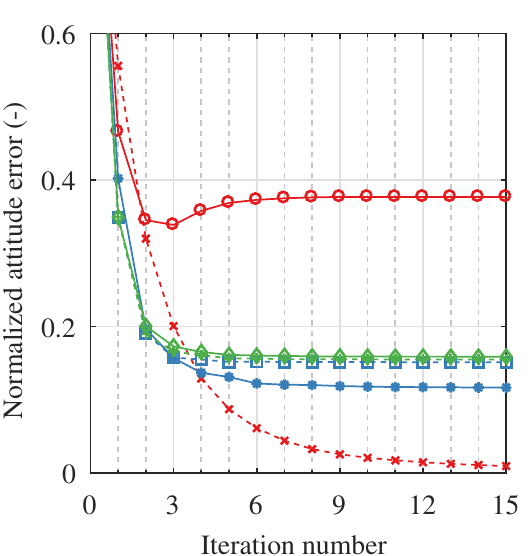}
	\caption{20\% outliers}
	\end{subfigure}\hfill%
		\begin{subfigure}[t]{0.33\textwidth} 	\centering
	\includegraphics[width=\linewidth]{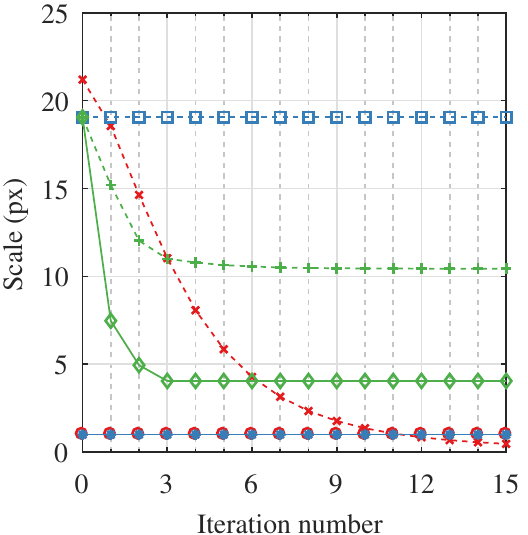}
	\end{subfigure}\\
		\begin{subfigure}[t]{0.33\textwidth} 	\centering
	\includegraphics[width=\linewidth]{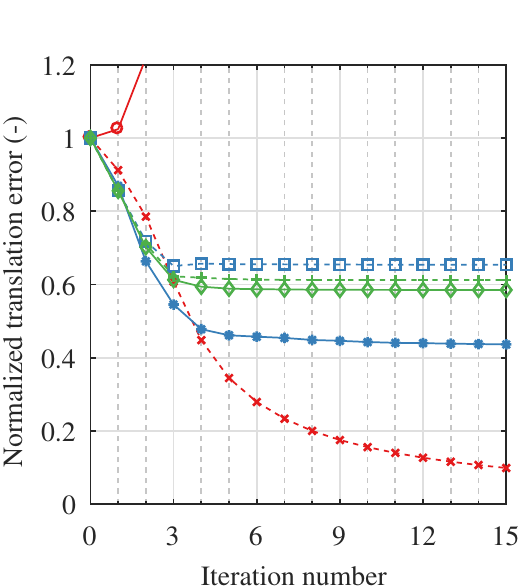}
	\end{subfigure}\hfill%
		\begin{subfigure}[t]{0.33\textwidth} 	\centering
	\includegraphics[width=\linewidth]{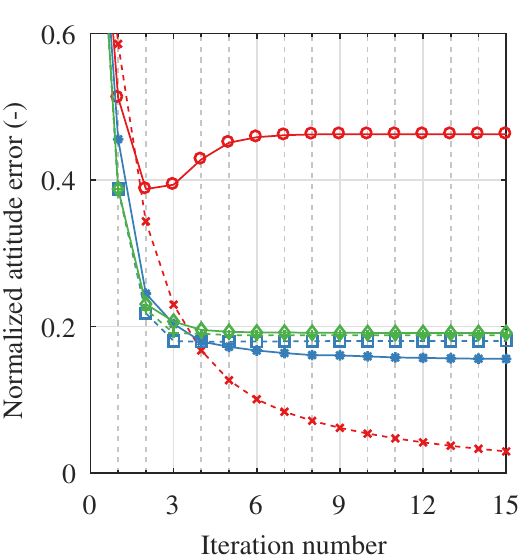}
	\caption{30\% outliers}
	\end{subfigure}\hfill%
		\begin{subfigure}[t]{0.33\textwidth} 	\centering
	\includegraphics[width=\linewidth]{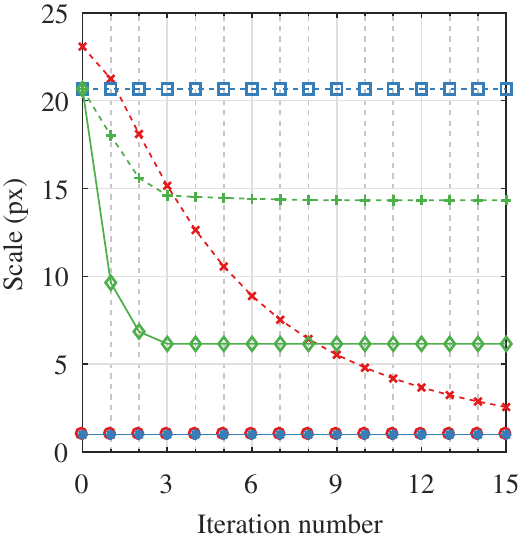}
	\end{subfigure}
  \caption{Minimization of the reprojection function from the images of a randomly generated point cloud, averaged over 100 runs.}
  \label{fig:me-mest_conv}
\end{figure}

The scale estimation step warrants special attention. In several applications, it can be found that $\sigma$ is often ignored and set to 1. This is erroneous since Eq.  (\ref{eq:me-mestimator}) is non-equivariant with respect to scale \cite{rousseeuw2005robust}. Whereas the Huber algorithm (Eq. (\ref{eq:me-normeqshuber})) grants a procedure to jointly estimate the parameter and scale, convergence is not guaranteed when applying the scale estimation step to \gls{irls} (Eq. (\ref{eq:me-normeqsirls})) \cite{huber2009robust}. Instead, a common method when resorting to \gls{irls} is to recursively estimate $\sigma$ using the \gls{mad} for the first few iterations, and then allowing the minimization to converge on $u$ with fixed $\sigma$ \cite{zhang1997parameter, stewart1999robust}.

In order to study the effect of scale estimation on the parameter estimation and to compare the different possible approaches, an experiment has been devised. First, a number of 3D world points is randomly sampled from the volume of a cube. They are subsequently projected onto the image plane according to a random pose. Points that fall outside the image plane are culled. Matches between 3D world points and 2D camera points are contaminated artificially with outliers. Then, the pose is M-estimated with ${\rho_\text{Hub}(x)}$ according to the cost function of Eq. (\ref{eq:me-objfunstruct}), where the initial guess is defined by contaminating the true pose with zero-mean, white, Gaussian noise. Five distinct methods are benchmarked: \begin{enumerate*} [label=\roman*\upshape)] \item \gls{ls}, \item Huber's algorithm, \item \gls{irls} with $\sigma = 1$, \item \gls{irls} with $\sigma$ estimated by one iteration of \gls{mad}, \item \gls{irls} with $\sigma$ estimated by three iterations of \gls{mad}, \item \gls{irls} with $\sigma$ estimated by Huber's algorithm\end{enumerate*}. The experiment is repeated for several trials.

The results are shown in Fig. \ref{fig:me-mest_conv}. The pose estimation error is decomposed into translation and rotation normalized according to the initial guess. The evolution of the scale estimation is also shown. The percentage of outliers present in the data ranges from \SI{10}{\percent} to \SI{30}{\percent}. It can be seen that Huber's algorithm yields the best estimate for every case. The regular \gls{ls} is able to somewhat reduce the attitude error in the presence of outliers, but diverges in the case of translation. Interestingly, all the \gls{irls} methods that estimate the scale perform worse than the case where the scale is ignored. These results show the impact on the solution of proper scale estimation and the preference of Huber's algorithm over others. This suggests that robust estimation should be initiated with Huber's algorithm until convergence; to ensure that the rejection of outliers is maximized, some additional iterations can be performed with \gls{irls} and a hard redescender, such as Tukey's function, using the (fixed) previously obtained estimate of $\sigma$, as suggested in Ref. \cite{zhang1997parameter}.

\section{Filtering}
\label{sec:fil}

\subsection{Rigid Body Kinematics}

The kinematics equation for $\sethree$ in matrix form is \cite{murray1994mathematical}:
\begin{equation}
\vectdot{T}_{B/T} = \vect{\varpi}_{B/T}^{B\wedge} \vect{T}_{B/T}, \quad \text{with } \vect{\varpi}_{B/T}^{B} \coloneqq \begin{pmatrix} \vect{\nu}_{B/T}^{B} \\ \vect{\omega}_{B/T}^{B} \end{pmatrix}, \label{eq:fil-se3kinematics}
\end{equation}
where $\vect{T}_{B/T}$ is the rigid body pose mapping $\rframe{T}$ to $\rframe{B}$, and $\vect{\varpi}_{B/T}^{B}$ is the rigid body velocity of $\rframe{T}$ with respect to $\rframe{B}$ expressed in $\rframe{B}$. It is the concatenation of two terms: $\vect{\omega}$, the instantaneous angular velocity of the target as seen from the chaser; and $\vect{\nu}$, the velocity of the point in $\rframe{T}$ that corresponds instantaneously to the origin of $\rframe{B}$. Dropping the subscripts and superscripts for succinctness, Eq. (\ref{eq:fil-se3kinematics}) is a first-order ordinary differential equation, and hence admits a closed-form solution of the form:
\begin{equation}
\vect{T}(t) = \exp((t-t_0)\vect{\varpi}^\wedge)\vect{T}(t_0). \label{eq:fil-se3kindiscrete}
\end{equation}
Equation (\ref{eq:fil-se3kindiscrete}) has the same form as Eq. (\ref{eq:mp-poseexp}), implying that $\vect{\varpi}$ is an element of $\setthree$. In agreement with the previous sections, this fact suggests that uncertainty can be introduced in the pose kinematics by modeling it as a local distribution in $\setthree$. As such, it is of interest to develop perturbation equations in terms of the kinematics in $\setthree$ so that these can be included as additive noise in a filtering scheme.

Following the approach of Ref. \cite{barfoot2017state}, the first two terms of Eq. (\ref{eq:mp-matexp}) are used to linearize Eq. (\ref{eq:mp-poseexp}) as ${\vect{T'} \approx (\vect{I} + \delta\vect{\xi}^\wedge)\vect{T}}$, where $\vect{T}$ is the nominal pose, $\delta\vect{\xi}$ is a small perturbation in $\setthree$, and hence $\vect{T'}$ is the resulting perturbed pose. Since ${\vect{\varpi} \in \setthree}$, this generalized velocity can be written directly as the sum of a nominal term with a small perturbation ${\vect{\varpi'} = \vect{\varpi} + \delta\vect{\varpi}}$. Substituting in Eq. (\ref{eq:fil-se3kinematics}), one has:
\begin{equation}
\frac{\urd}{\urd t}\left(\left(\vect{I} + \delta\vect{\xi}^\wedge\right)\vect{T}\right) \approx \left(\vect{\varpi} + \delta\vect{\varpi}\right)^\wedge \left(\vect{I} + \delta \vect{\xi}^\wedge\right)\vect{T}.
\end{equation}
Expanding, ignoring the product of small terms and applying the Lie bracket of $\setthree$ yields the perturbation kinematics equation for $\sethree$:

\begin{equation}
\delta\vectdot{\xi} = \adset(\vect{\varpi})\delta\vect{\xi} + \delta\vect{\varpi}, \label{eq:fil-se33kinematics}   
\end{equation}

\noindent which is linear in both $\delta\vect{\xi}$ and $\delta\vect{\varpi}$.

\subsection{Extended Kalman Filter Formulation}
\subsubsection{Motion Model}

Equation (\ref{eq:fil-se33kinematics}) describes effectively the linearization of the rigid body kinematics around a nominal pose. Since it is defined with respect to elements of $\setthree$, perturbations in the motion can be modelled stochastically in terms of a local distribution (Section \ref{sec:motionestimation}). The mean of this distribution may be injected into the nominal values via the exponential map. Under the assumption of Gaussian noise, this equation can therefore be regarded as the first step in defining an error-state to model how the motion evolves in time in the framework of an extended Kalman filter. 

The kinematics of the target's motion with respect to the chaser spacecraft are correctly modelled by Eqs. (\ref{eq:fil-se3kinematics}) and (\ref{eq:fil-se33kinematics}). Modelling the relative dynamics, however, is not a clear-cut task. In the case of an asteroid mission, for example, the chaser could be considered to be inside the sphere of influence of the target and then Newton's second law of motion and Euler's rotation equation could be applied. However, in the case where both chaser and target are under the influence of the same primary, the relative dynamics cannot be shaped as such. 

In order to design a filter exclusively with relative states, and inspired by the method of Ref. \cite{davison2007monoslam}, a broader constant generalized velocity motion model is adopted:
\begin{equation}
\vectdot{\varpi}(t) = \vect{\eta}_{\vect{\varpi}}(t), \quad\vect{\eta}_{\vect{\varpi}}(t)\sim \mathcal{N}(\vect{0}, \vect{Q}(t)\delta(t-\tau)), \label{eq:fil-constvel}
\end{equation}
where it is assumed that the spectral density matrix $\vect{Q}(t)$ is diagonal. Note that, as stated in Ref. \cite{davison2007monoslam}, this model does not assume that the chaser moves at a constant velocity over the entire sequence, but instead that undetermined accelerations with a Gaussian profile are expected to occur on average. In other words, one assumes that sizeable (relative) accelerations are unlikely to be experienced, which is a valid expectation for a space rendezvous. 

Integrating Eq. (\ref{eq:fil-constvel}) yields $\vect{\varpi}(t) = \vect{\varpi}(t_0) + \int_{t_0}^t \vect{\eta}_{\vect{\varpi}}(\tau) \, \urd \tau$. The relation ${\vect{\varpi^\prime} = \vect{\varpi} + \delta\vect{\varpi}}$ was assumed earlier, meaning that one can admit
\begin{equation}
\delta\vect{\varpi}(t) = \int\limits_{t_0}^t \vect{\eta}_{\vect{\varpi}}(\tau) \, \urd \tau.
\end{equation}
Defining the error state ${\delta\vect{x} \coloneqq \icolsmall{\delta\vect{\xi}, \delta\vect{\varpi}}^\top}$, the continuous-time error kinematics are written directly:
\begin{equation}
    \frac{\urd}{\urd t}\delta \vect{x}(t) = \vect{F}(t)\delta\vect{x}(t) + \vect{G}(t)\vect{w}(t) = \begin{bmatrix} \adset(\vect{\varpi}) & \vect{I}_6\\ \vect{0}_{6\times 6} & \vect{0}_{6\times 6}\end{bmatrix} \begin{pmatrix}\delta\vect{\xi}\\ \delta\vect{\varpi}\end{pmatrix} + \begin{bmatrix}\vect{0}_{6\times 6} \\ \vect{I}_6\end{bmatrix} \begin{pmatrix}\vect{0}_{3 \times 1} \\ \vect{\eta}_{\vect{\varpi}}\end{pmatrix}, \label{eq:fil-err_sta_dyn}
\end{equation}
which shows that process noise is introduced in the system through the error generalized velocity vector. Equation (\ref{eq:fil-err_sta_dyn}) has the familiar solution \cite{grewal2015kalman}:
\begin{equation}
    \delta\vect{x}(t) = \exp\left(\int\limits_{t_0}^t \vect{F}(\tau) \urd \tau \right) \delta\vect{x}(t_0) + \int\limits_{t_0}^t \exp\left(\int\limits_{s}^t \vect{F}(\tau) \urd \tau \right) \vect{G}(s)\vect{w}(s) \urd s
\end{equation}
The error-state transition matrix has a known closed form \cite{barfoot2017state}:
\begin{equation}
    \vect{\Phi}(t,s) \coloneqq \exp\left(\int\limits_{s}^t \vect{F}(\tau) \urd \tau \right) = \begin{bmatrix} \adse(\exp((t-s)\vect{\varpi}^\wedge)) & (t-s)\vect{B}((t-s)\vect{\varpi}))\\
    \vect{0}_{3\times 3} & \vect{I}_3 \end{bmatrix}, \quad \text{with } \vect{B}(\vect{\xi}) \coloneqq \begin{bmatrix}\vect{M}(\vect{\phi}) & \vect{N}(\vect{\xi})\\ \vect{0}_{3 \times 3} & \vect{M}(\vect{\phi})\end{bmatrix}, \label{fil-eq:stm}
\end{equation}

\noindent where, defining ${\phi \coloneqq \Vert \vect{\phi} \Vert}$,

\begin{equation}
\begin{split}
\vect{M}(\vect{\xi}) & \coloneqq \frac{1}{2}\vect{\rho}^\wedge + \left(\frac{\phi - \sin \phi}{\phi^3}\right) \left(\vect{\phi}^\wedge \vect{\rho}^\wedge +  \vect{\rho}^\wedge \vect{\phi}^\wedge + \vect{\phi}^\wedge \vect{\rho}^\wedge \vect{\phi}^\wedge \right) + \left(\frac{\phi^2 + 2\cos \phi - 2}{2 \phi^4}\right) \left(\vect{\phi}^\wedge \vect{\phi}^\wedge \vect{\rho}^\wedge + \vect{\rho}^\wedge \vect{\phi}^\wedge \vect{\phi}^\wedge - 3 \vect{\phi}^\wedge \vect{\rho}^\wedge \vect{\phi}^\wedge \right) \\
&+ \left(\frac{2\phi - 3\sin \phi +\phi \cos \phi}{2 \phi^5}\right) \left(\vect{\phi}^\wedge \vect{\rho}^\wedge \vect{\phi}^\wedge \vect{\phi}^\wedge + \vect{\phi}^\wedge \vect{\phi}^\wedge \vect{\rho}^\wedge \vect{\phi}^\wedge\right).
\end{split}
\end{equation}
A closed-form of the discrete-time error process noise covariance matrix is found by directly solving the integral:
\begin{equation}
   \vect{\Gamma}(t,s) \coloneqq  \int\limits_{t_0}^t  \vect{\Phi}(t,s)\vect{G}(s)\vect{Q}(s)\vect{G}^\top(s)\vect{\Phi}^\top(t,s) \, \urd s, \label{eq:fil-gamma_mat_init}
\end{equation}
The derivation is monotonous but a matter of integrating each matrix element. To simplify it, the small angle approximation is applied and terms of $\mathcal{O}((t-s)^4)$ are discarded; the final result is not presented here.

\subsubsection{Measurement Model}

The correction stage of the \gls{ekf} admits pseudo-measurements of the relative pose ${y_i \in \mathcal{Y} \cong \sethree}$ as obtained through the refinement scheme of visual features correspondence from Section \ref{sec:motionestimation}. These pseudo-measurements are acquired at each sampling time and modelled as being corrupted by a zero-mean white Gaussian noise term. One can thus write directly in discrete-time and matrix form:
\begin{equation}
\vect{Y} = \exp(\vect{\eta}_y^\wedge)\vect{T}, \quad \vect{\eta}_y\sim \mathcal{N}(\vect{0}, \vect{R}), \label{eq:fil-measmodel}
\end{equation}
where $\vect{Y} \in \sethree$ is the matrix form of $y$. To linearize Eq. (\ref{eq:fil-measmodel}), similarly to the motion model, the elements of $\sethree$ are rewritten as a small perturbation around a nominal term, i.e. ${\vect{Y^\prime} = \exp(\delta\vect{y}^\wedge)\vect{Y}}$, ${\vect{T^\prime} = \exp(\delta\vect{\xi}^\wedge)\vect{T}}$, and the exponential map is approximated by its first-order expression. Replacing in Eq. (\ref{eq:fil-measmodel}), expanding and neglecting the product of small terms, the following linearized relationship is obtained:

\begin{equation}
    \vect{Y^\prime} = \vect{T^\prime}, \quad \delta\vect{y} = \delta \vect{\xi} + \vect{\eta}_y.
\end{equation}

\noindent The full linearized measurement model is therefore:

\begin{equation}
    \delta\vect{y}_i = \vect{H}\delta\vect{x} + \vect{\eta}_{y_i}  = \begin{bmatrix} \vect{I}_6 & \vect{0}_{6 \times 6}\end{bmatrix}\begin{pmatrix}\delta\vect{\xi}\\ \delta\vect{\varpi}\end{pmatrix} + \vect{\eta}_{y_i}, \quad \vect{\eta}_{y_i}\sim \mathcal{N}(\vect{0}, \vect{R}_i) \label{eq:fil-meas}
\end{equation}

\noindent The covariance matrices of each pseudo-measurement are obtained as a product of the minimization scheme. In the case of the structural model constraints, the Jacobians $\vect{J}_s$ are of rank 6, so the covariance of the solution is given by backpropagation of the visual feature correspondences' own covariance \cite{hartley2004multiple}:
\begin{subequations}
\begin{equation}
    \vect{R}_s = \vect{\Sigma}_s = \left(\vect{J}_s^\top \vect{\Sigma}_{\vect{z,s}} \vect{J}_s \right)^{-1},
\end{equation}
with ${\vect{\Sigma}_{\vect{z,s}} = \sigma_{\vect{z,s}}^2\vect{I}}$ and  $\sigma_{\vect{z,s}}$ is obtained via M-estimation (cf. Subsection \ref{sec:me-re}). The \gls{ekf} innovation term at $t = k$ is:
\begin{equation}
\vect{\upsilon}_{s,k} = y_{s,k} \ominus \hat{u}^-_k \in \setthree ,
\end{equation}
\end{subequations}
where $\hat{u}^-_k$ is the predicted pose at $t = k$. 

\subsubsection{Measurement Gating}

An additional step is employed prior to the correction to ensure the accurate functioning of the filter. This involves subjecting the incoming measurements to a validation gate, thus discarding potential spurious data. The validation gate is a threshold on the \gls{rmse} of the residuals $\vect{r}$ as obtained by the M-estimation:
\begin{equation}
    \mathrm{RMSE}(\vect{r}) \coloneqq \sqrt{\frac{\sum^n_{i=1}r_i^2}{n}}.
\end{equation}
The \gls{rmse} provides an objective and clear interpretation of how close, in pixels, does the feature matching agree with the estimate of the pose.

\subsection{Manifold State Prediction and Correction}

The nominal state is construed as ${\vect{x}^\top = \icolsmall{u^\top, \vect{\varpi}^\top}}$, with ${u \in \mathcal{U} \cong \sethree}$ representing the relative pose mapping $\rframe{T} \to \rframe{B}$ and $\vect{\varpi} \in \mathbb{R}^6$ is the generalized velocity satisfying the kinematics equation for $\sethree$, Eq. (\ref{eq:fil-se3kinematics}). The nominal state estimate is updated with a linearized error state estimate ${\delta\vect{x}^\top = \icolsmall{\delta\vect{\xi}^\top, \delta\vect{\varpi}^\top} \in \setthree \times \mathbb{R}^6}$ via pose composition (cf. Eq. (\ref{eq:mp-posecomposition})), ensuring that $u$ remains an element of $\mathcal{U} \cong \sethree$. The algorithm's equations are valid for any chosen representation of $u$ provided the appropriate composition $\oplus$ is used. State prediction is performed as:

\begin{equation}
    \hat{u}^-_k = \hat{u}_{k-1} \oplus \Delta t \vecthat{\varpi}_{k-1}, \quad
    \vecthat{\varpi}^-_{k} = \vecthat{\varpi}_{k-1}.
\end{equation}

\noindent The state correction is given by:

\begin{equation}
\hat{u}^+_k = \hat{u}_k^- \oplus \delta\vecthat{\xi}^+_k, \quad
\vecthat{\varpi}^+_k = \vecthat{\varpi}_k^- + \delta\vecthat{\varpi}^+_k,
\end{equation}

\noindent where $\delta\vect{x}^{-\top}_k = \icolsmall{\hat{u}^{-\top}_k,\vecthat{\varpi}^{-\top}_k }$, $\delta\vect{x}^{+\top}_k = \icolsmall{\hat{u}^{+\top}_k,\vecthat{\varpi}^{+\top}_k }$ are the a priori and a posteriori error states, respectively. The covariance is calculated using the standard \gls{ekf} equations.

In terms of the parameterization of $u$, a possible choice would be taking ${u \to \vect{T}}$ directly, in which case the composition operation is given by Eq. (\ref{eq:mp-poseexp}). Alternatively, the unit quaternion is a popular choice for attitude parameterization due to its compact and singularity-free representation, particularly in aerospace applications. The exponential map ${\sotthree \to \sutwo}$ has a simple closed form given by $\exp_q = \icolsmall{\vect{\phi}^\top \sin(\phi/2)/\phi, \cos \phi/2 }^\top$, with $\phi = \lVert \vect{\phi} \rVert$. \cite{blanco2010tutorial}. In this case, the state vector becomes, with some abuse of notation, $\vect{x}^\top = \icolsmall{u^\top,\vect{\varpi}^\top} = \icolsmall{\vect{t}^\top,\vect{q}^\top,\vect{\varpi}^\top}$, which has dimension $13\times 1$.

\section{Results}
\label{sec:res}

This section presents the the experimental results that validate the framework proposed herein. The considered target spacecraft is Envisat, a complex debris object, formed by several modules, namely a solar panel array, a \gls{sar}, and several antennae, among others, connected to a main body unit which is covered by \acrfull{mli}. The experiments are structured into three main parts: \begin {enumerate*} [label=\roman*\upshape)] \item evaluation of the coarse pose classification module, \item evaluation of the full pose estimation pipeline on a synthetic dataset, and \item evaluation of the full pose estimation pipeline on a laboratory dataset. \end{enumerate*}

All processing and simulations are carried out on a setup using an Intel\textsuperscript{\textregistered} Core\textsuperscript{TM} i7-6700 @ \SI{3.40}{\giga\hertz} $\times$ 8 core processor, \SI{16}{\giga\byte} RAM system.

\subsection{Coarse Pose Classification}

The performance of the coarse pose classification module is first evaluated independently. Data is generated from the viewsphere of a \gls{cad} model of Envisat freely available from the space simulator Celestia\footnote{\url{http://celestia.space/}.}. \gls{cad} manipulation and keyframe generation are performed using the open-source 3D computer graphics software Blender\footnote{\url{http://www.blender.org/}.}. Since shape information is used as the basis for feature description, the results are extendable to the case of a real dataset. The module was built in Matlab and the Bayesian classifier was implemented using the GMMBayes toolbox\footnote{\url{http://www.it.lut.fi/project/gmmbayes/}.}.

The analysis is done via $k$-fold cross-validation. Cross-validation a set of techniques that assess how the results of a statistical analysis can generalize to an independent dataset; particularly, in $k$-fold cross-validation, the dataset is randomly partitioned into subsamples of equal size $k$. One subsample is retained to be the validation set for testing the model, and $k-1$ subsamples are used as training data. The cross-validation process is repeated $k$ times, with each of the $k$ sets used exactly once as the validation data. In this way, all observations are used for both training and validation and each observation is used for validation only once. This provides a framework to quickly assess and tune model parameters, such as number of classes and descriptor size.

\begin{table}[t]
 \begin{center}
   \caption{Settings used for $k$-folds validation of the coarse pose classification module.}
   \label{tab:res-coarse_params}
   \begin{tabular}{lcc}
	\toprule \toprule
   \multicolumn{1}{c}{\textbf{Parameter}} & \textbf{Units} & \textbf{Value}\\ \midrule
    Viewsphere azimuth mesh step, $\Delta_\text{mesh}^\text{az}$ &[\si{\deg}] & 1\\
    Viewsphere elevation mesh step, $\Delta_\text{mesh}^\text{el}$ &[\si{\deg}] & 1\\
    Viewsphere azimuth class step, $\Delta_\text{class}^\text{az}$ &[\si{\deg}] & 10\\
    Viewsphere elevation class step, $\Delta_\text{class}^\text{el}$& [\si{\deg}] & 10\\
    Total classes & [-] & 648\\
    \gls{zm} vector dimension & [-] & 60\\
    Folds, $k$ & [-] & 10\\
    \bottomrule \bottomrule
   \end{tabular}
 \end{center}
\end{table}

Table \ref{tab:res-coarse_params} depicts the used parameters for the cross-validation test. Shapes are rendered from the \gls{cad} model viewsphere with a \SI{1}{\deg} step and grouped into bins of \SI{10}{\deg} in azimuth and elevation, thus defining the minimum achievable accuracy for the coarse pose classifier. This reduces the classification problem to 648 possible classes. 10 folds are selected, meaning that \SI{10}{\percent} of the data population is used for training per fold. 

\begin{figure}[t]
	\centering
	\begin{subfigure}[t]{0.48\textwidth} 	\centering
	\includegraphics[width=\linewidth]{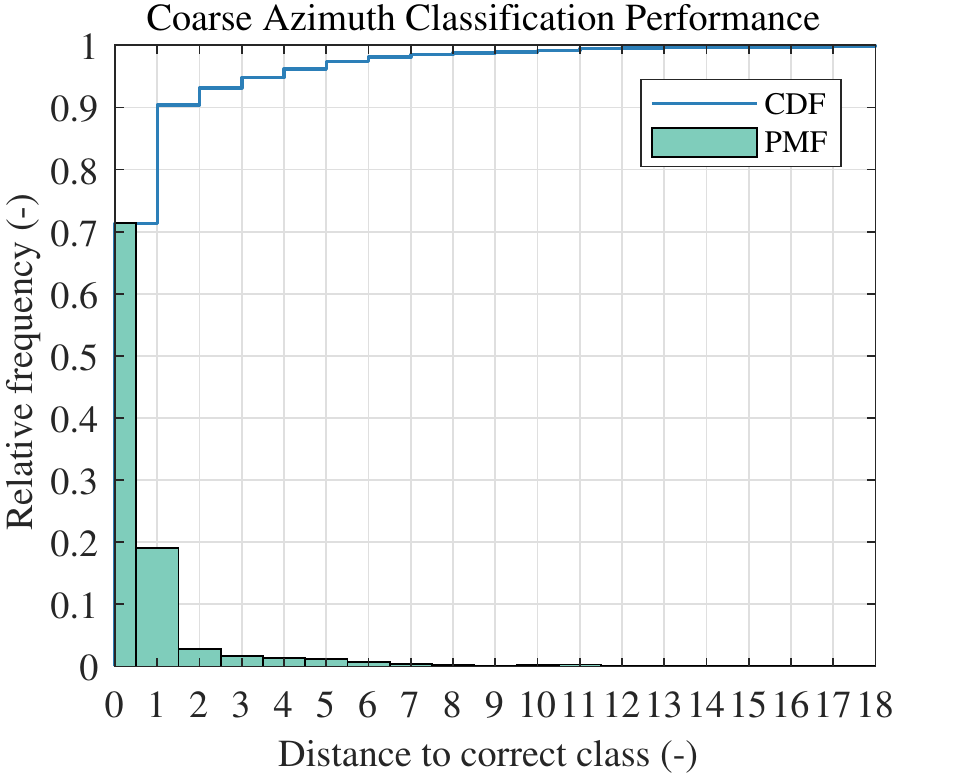}
	\caption{Azimuth} \label{fig:res-coarse_azl}
	\end{subfigure}\hfill%
	\begin{subfigure}[t]{0.48\textwidth} 	\centering
	\includegraphics[width=\linewidth]{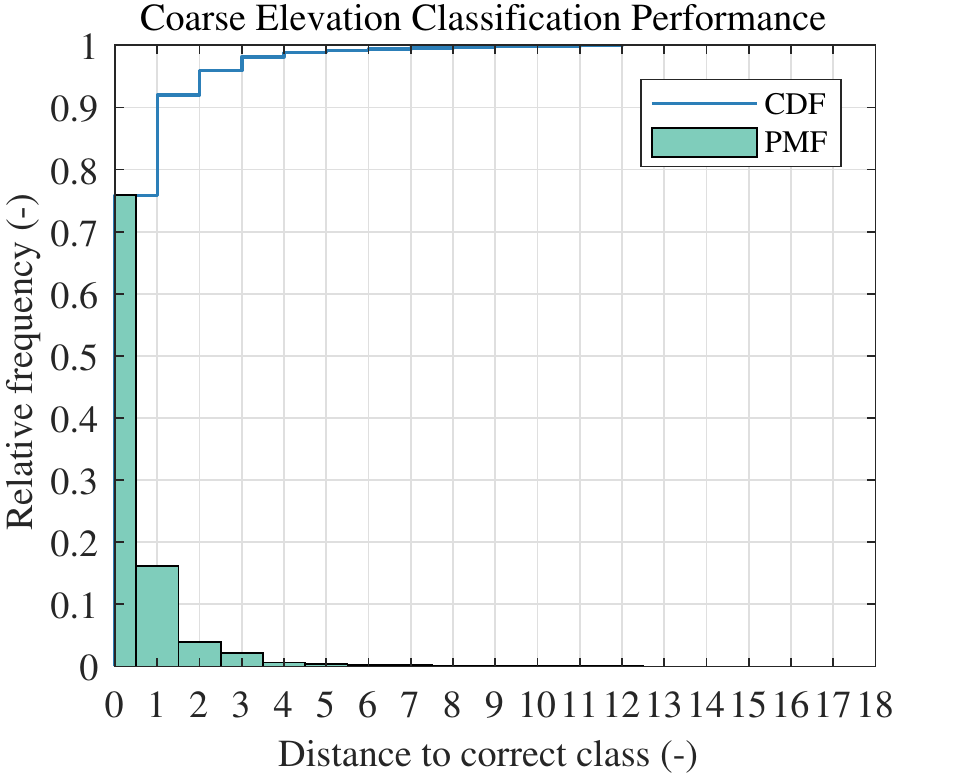}
	\caption{Elevation} \label{fig:res-coarse_el}
	\end{subfigure}%
 \caption{Histogram of results of the $k$-folds validation for the coarse pose classification..}
\label{fig:res-coarse}
\end{figure}

The results are illustrated in Fig. \ref{fig:res-coarse} for azimuth and elevation classification performance. The horizontal axis represents the error in terms of bin distance, where a value of ``0'' represents a correct classification. Two quantities are represented: the \gls{pmf} for each class in the form of a histogram, and the (discrete) \gls{cdf}. The correct classification rate for azimuth is \SI{71.35}{\percent}. \SI{90.41}{\percent} of the data is classified with a bin distance less than or equal to \si{1}, i.e. with a maximum error of \SI{20}{\deg}. Despite there being 36 possible azimuth classes with the parameters chosen in Table \ref{tab:res-coarse_params}, the shortest distance is considered, i.e. modulo \SI{180}{\deg}. The results are slightly better for the elevation, with a correct classification rate of \SI{75.86}{\percent} and \SI{92}{\percent} of the data within a bin distance of \si{1}.

\subsection{Nominal Pose Estimation}

\subsubsection{Synthetic Dataset}

In this section, the performance of the full spacecraft relative pose estimation pipeline is assessed. This includes the initialization procedure with the coarse pose classifier followed by the pose refinement using local features (cf. Section \ref{sec:so}). The pipeline is first tested with a synthetically generated rendezvous trajectory. To this end, the Envisat \gls{cad} model was modified using Blender with additional materials and textures in order to yield a realistic aspect in face of the expected low Earth orbit conditions. This includes a complete remodeling of the \gls{mli} to recreate diffuse light reflection, and the addition of reflective materials to the solar panel. Both of these aspects represent challenging imaging conditions in the visible wavelength for relative pose estimation. The framework was coded in the C++ programming language, whereas the OpenCV library\footnote{\url{https://opencv.org/}.} was used for \gls{ip}-related functions.

\begin{table}[]
 \begin{center}
   \caption{Characterization of the chaser and target's orbital motion for rendezvous simulation.}
   \label{tab:res-envisat_orbparam}
   \begin{tabular}{lcc}
	\toprule \toprule
   \multicolumn{1}{c}{\textbf{Parameter}} & \textbf{Units} & \textbf{Value}\\ \midrule
    Target orbital parameters & & \\ \midrule
    Semi-major axis & [\SI{}{\kilo\metre}] & \num{7.1427e+03}\\
    Eccentricity & [-] & \num{7.6112e-04}\\
    Inclination &[\si{\deg}] & \num{98.2164}\\
    Right ascention of the ascending node & [\si{\deg}] & \num{343.0760}\\
    Argument of perigee & [\si{\deg}] & \num{189.5264}\\
    True anomaly & [\si{\deg}] & \num{3.0109}\\
    Spin axis in target frame & [-] & Aligned with $+\vect{t}_2$ axis\\
    Spin axis in \acrshort{lvlh} frame & [-] & Aligned with H-bar axis\\
    Spin rate & [\si{\deg\per\second}] & \num{3.5}\\
   
    \bottomrule \bottomrule
   \end{tabular}
 \end{center}
\end{table}

\begin{figure}[]
	\centering
	\begin{subfigure}[t]{0.48\textwidth} 	\centering
	\includegraphics[width=\linewidth]{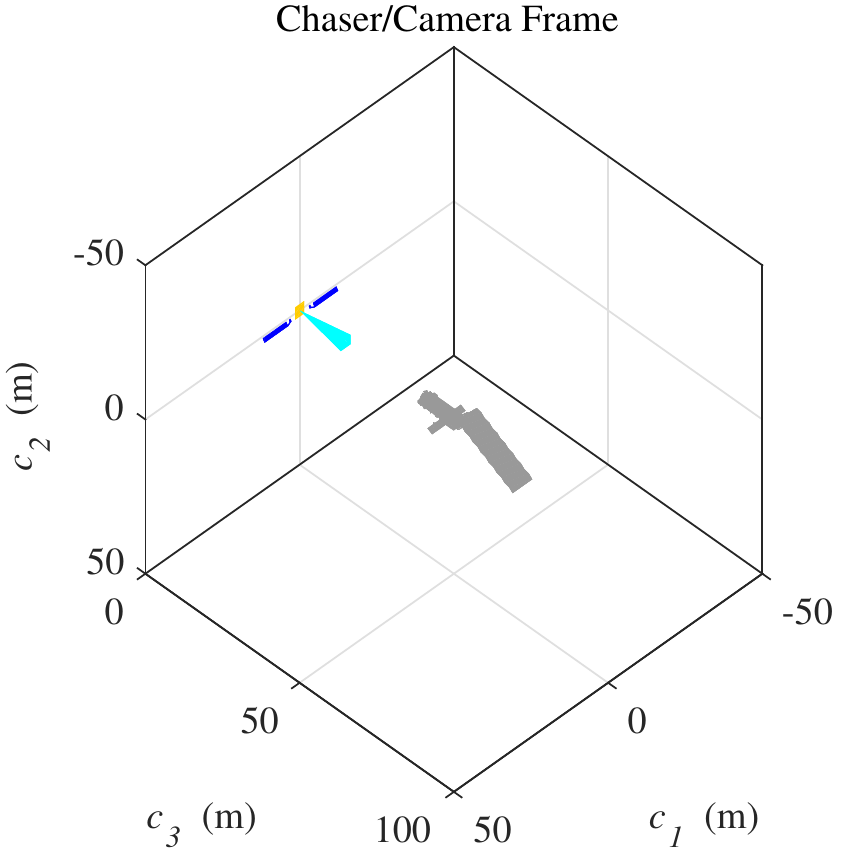}
	\caption{Chaser body $\rframe{B}$/camera $\rframe{C}$ frame} 
	\end{subfigure}\hfill%
	\begin{subfigure}[t]{0.48\textwidth} 	\centering
	\includegraphics[width=\linewidth]{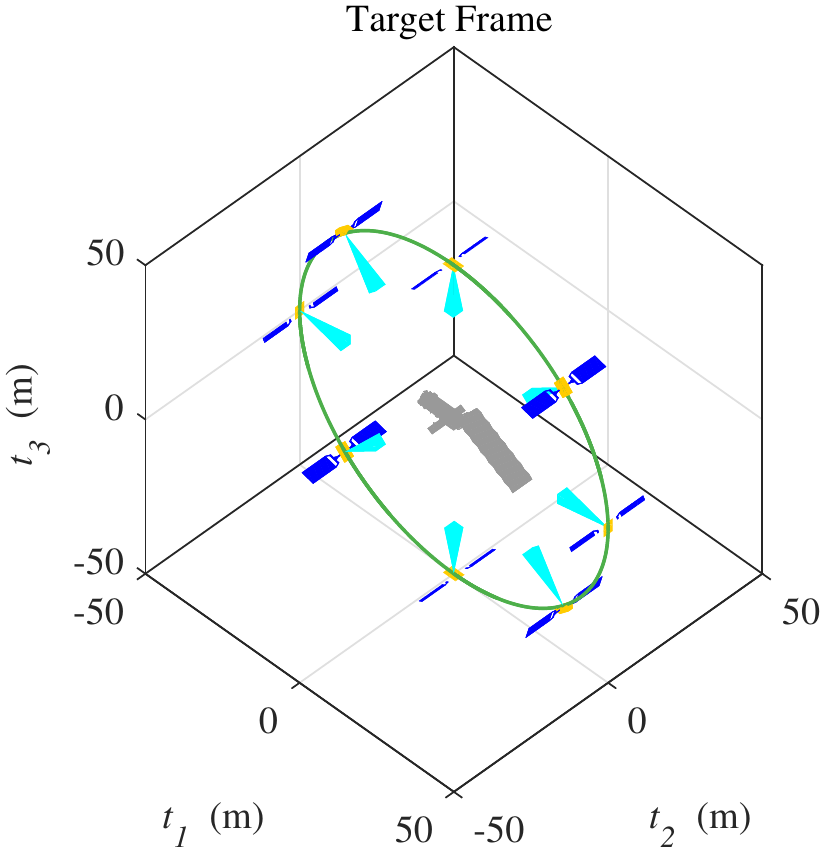}
	\caption{Target $\rframe{T}$ frame} 
	\end{subfigure}%
 \caption{Simulated trajectory around the Envisat spacecraft. The cyan pyramid represents the camera's \gls{fov}.}
\label{fig:res-synthetic_traj}
\end{figure}

The dataset was generated with the Astos Camera Simulator\footnote{\url{http://www.astos.de/}.}. The orbit of Envisat was simulated using the two-line element set corresponding to 30 October 2017\footnote{TLE data obtained from NORAD Two-Line Element Sets
Current Data \url{http://www.celestrak.com/NORAD/elements/.}}. This corresponds to the orbital parameters in Table \ref{tab:res-envisat_orbparam}. In terms of rendezvous kinematics, the chaser is assumed to observe the target from a hold point on V-bar in the target's \gls{lvlh} frame. The chaser rotates inertially along a fixed axis, thus the perceived relative motion is circular. The initial relative pose is $\vect{t}^C_{C/T,0} = \icolsmall{0,0,50}^\top$ \si{\metre}, $\vect{q}_{C/T,0} = \icolsmall{-\sqrt{2}/2,0,-\sqrt{2}/2,0}^\top$. The quaternion of attitude was chosen so as to minimize the target's area as observed by the chaser, recreating a scenario of challenging initialization. An illustration of the relative trajectory is shown in Fig. \ref{fig:res-synthetic_traj}.

\begin{table}[t]
 \begin{center}
   \caption{Pose estimation pipeline configuration and numerical settings.}
   \label{tab:res-pipeline_params}
   \begin{tabular}{lccc}
	\toprule \toprule
   \multicolumn{1}{c}{\textbf{Parameter}} & \textbf{Units} & \textbf{Value, Synthetic}  & \textbf{Value, Laboratory}\\ \midrule
    Camera & & \\ \midrule
    Focal length & [\si{\milli\metre}] & 5 & 3\\
    Resolution & [\si{\pixel}$\times$\si{\pixel}] & \num{640}$\times$\num{480} & \num{752}$\times$\num{480}\\
    \acrshort{fov} & [\si{\deg}$\times$\si{\deg}]  & \num{53.06}$\times$\num{36.82} & \num{79.52}$\times$\num{58.03}\\
    Framerate & [\si{\hertz}] & \num{10} & \num{10}\\\\
    Viewsphere & & \\ \midrule
    Azimuth step & [\si{deg}] & \num{9} & \num{9}\\
    Elevation step & [\si{deg}] & \num{9} & \num{9}\\
    Total keyframes & [-] & \num{800} & \num{800}\\
    \bottomrule \bottomrule
   \end{tabular}
 \end{center}
\end{table}

Table \ref{tab:res-pipeline_params} shows the parameters employed in the test. The compact mvBlueFOX MLC202b camera is simulated, as it is similar to the one utilized in the experimental setup (cf. Section \ref{subsubsec:res-lab}). A step of \SI{9}{\deg} both in azimuth and elevation was chosen to build the offline database, resulting in 800 keyframes. Note that, for this particular simulated trajectory, only 40 keyframes would be required, since the azimuth is fixed; however, to stress the algorithm, the full set of possible keyframes to choose from is kept. The \gls{ekf} is run with a timestep of $\SI{0.1}{\second}$ (the sampling rate of the camera). The initial filter pose state  $\vecthat{t}_0,\vecthat{q}_0$ is initialized with the result of the coarse pose estimation, while the velocity state is pessimistically assumed to be equal to zero. The initial covariance $\vecthat{P}_0$ and the process noise covariance $\sigma_\nu^2,\sigma_\omega^2$ are tuned empirically, whereas the measurement noise covariance is determined via M-estimation.

\begin{figure}[]
	\centering
	\begin{subfigure}[t]{0.48\textwidth} 	\centering
	\includegraphics[width=\linewidth]{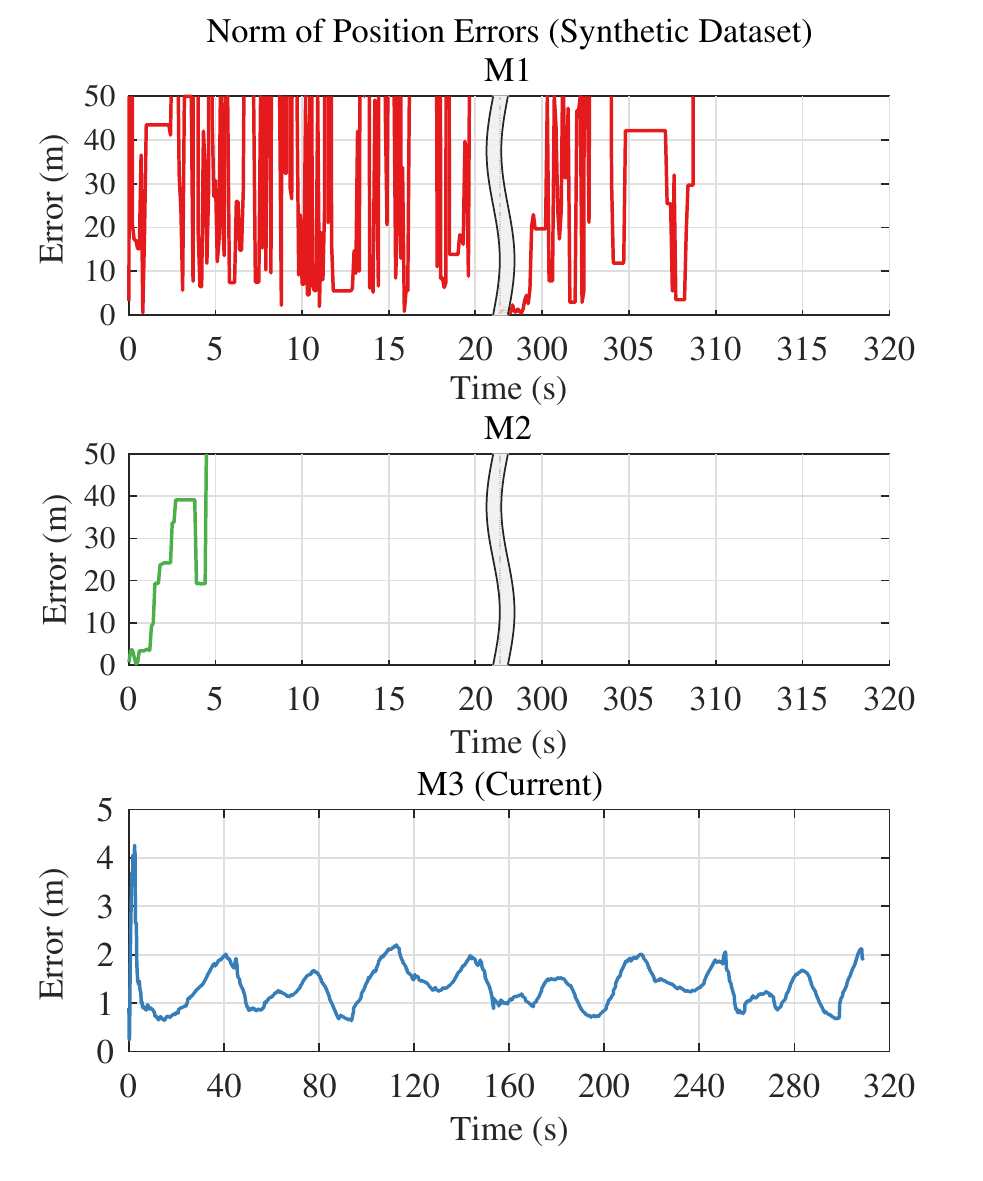}
	\caption{Position} \label{fig:res-synthetic_fine1_translation}
	\end{subfigure}\hfill%
	\begin{subfigure}[t]{0.48\textwidth} 	\centering
	\includegraphics[width=\linewidth]{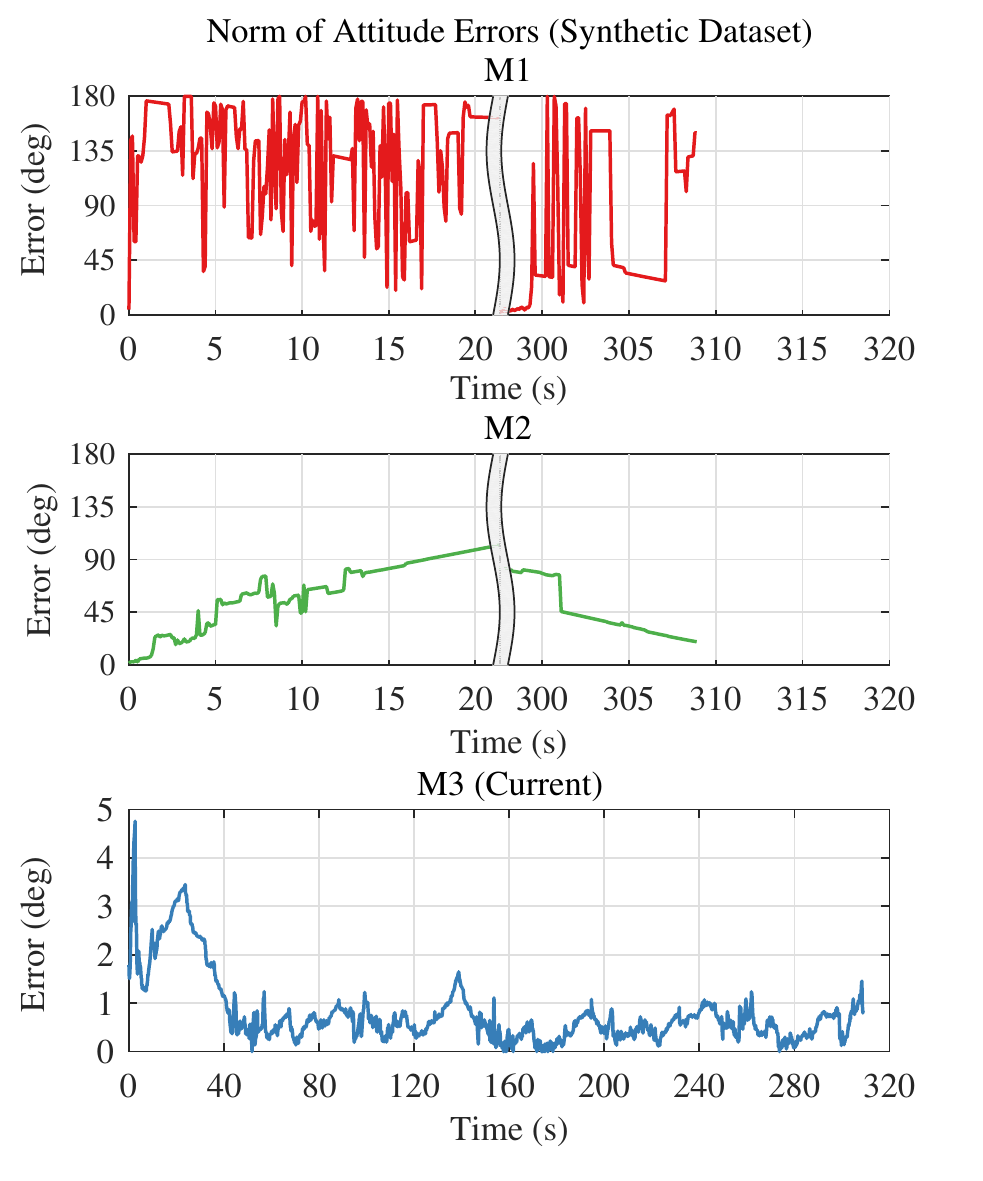}
	\caption{Attitude} \label{fig:res-synthetic_fine1_attitude}
	\end{subfigure}%
 \caption{Nominal pose estimation errors for the synthetic dataset.}
\label{fig:res-synthetic_fine1}
\end{figure}

\begin{figure}[]
	\centering
	\begin{subfigure}[t]{0.48\textwidth}
	\includegraphics[width=\linewidth]{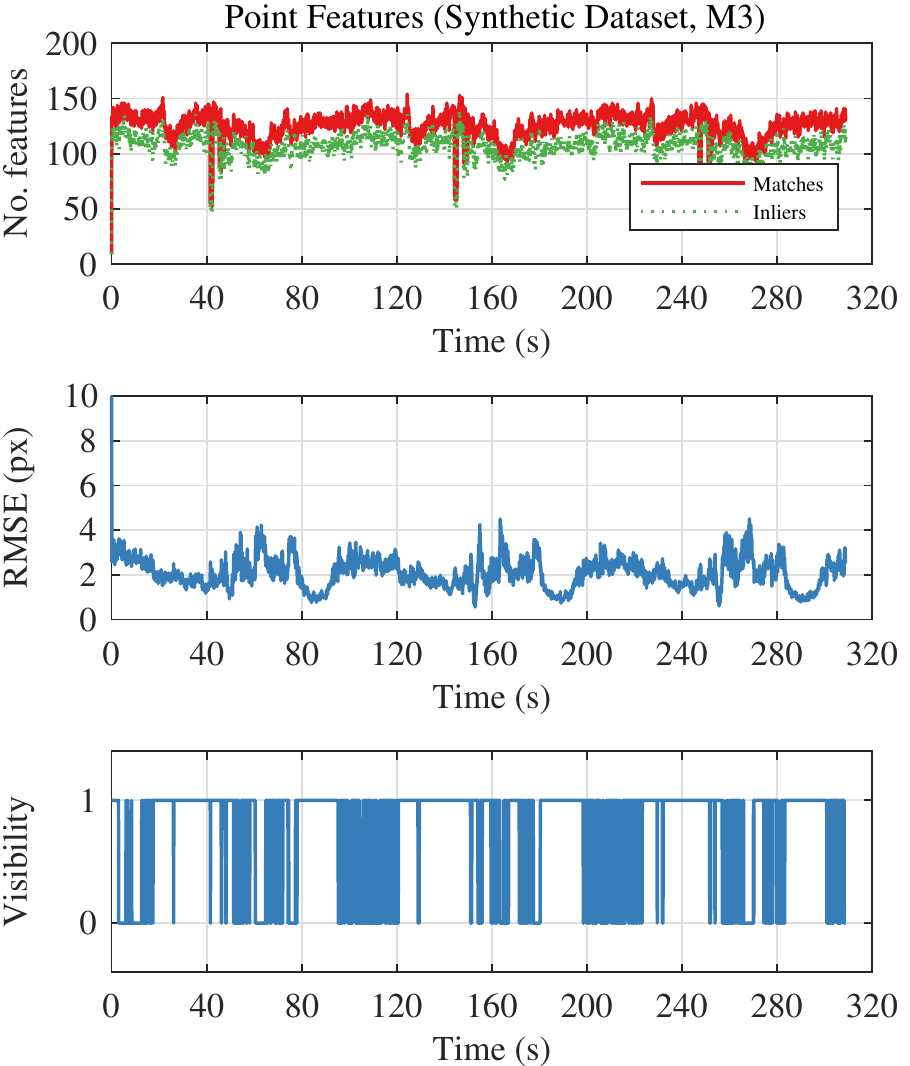} 	\centering
	\caption{Points} \label{fig:res-synthetic_features_pt}
	\end{subfigure}\hfill%
	\begin{subfigure}[t]{0.48\textwidth}
	\includegraphics[width=\linewidth]{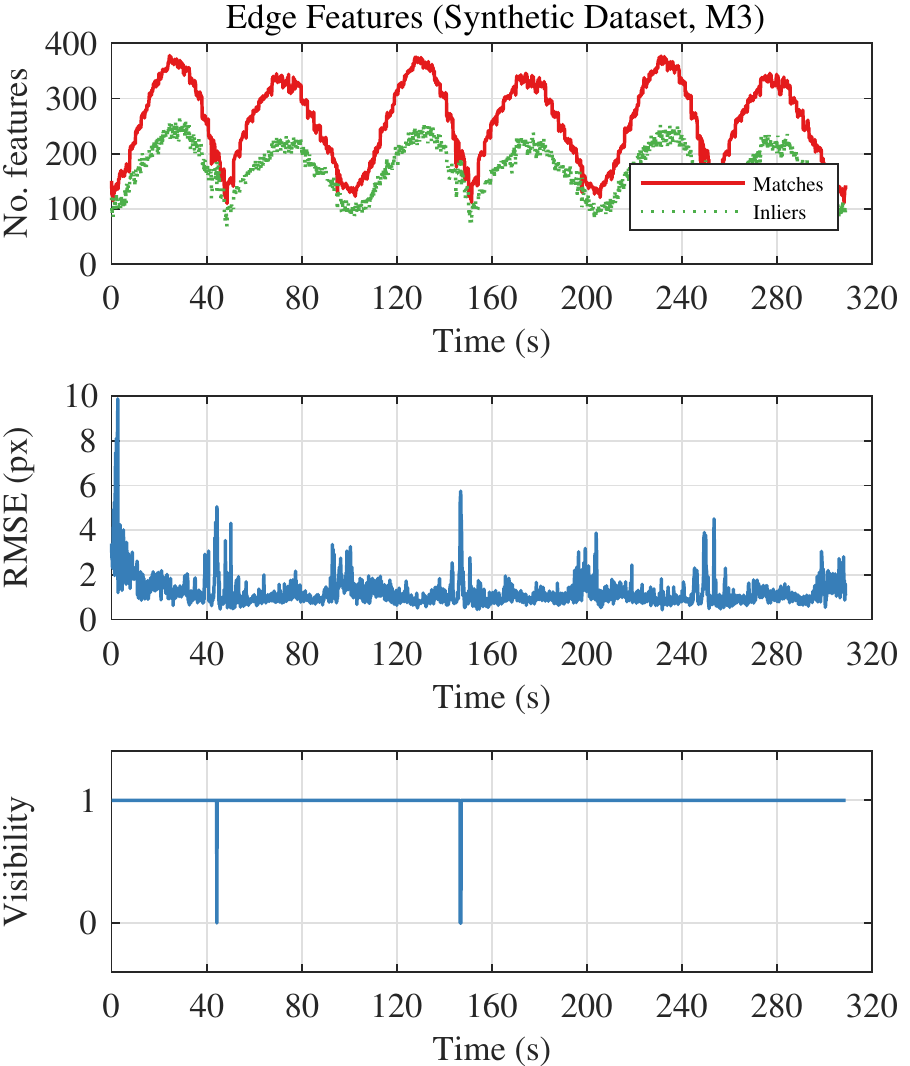} 	\centering
	\caption{Edges} \label{fig:res-synthetic_features_ed}
	\end{subfigure}%
 \caption{Feature statistics for nominal pose estimation sequence of the synthetic dataset.}
\label{fig:res-synthetic_features}
\end{figure}

\begin{figure}[]
	\centering
	\begin{subfigure}[t]{0.24\textwidth}
	\includegraphics[width=\linewidth]{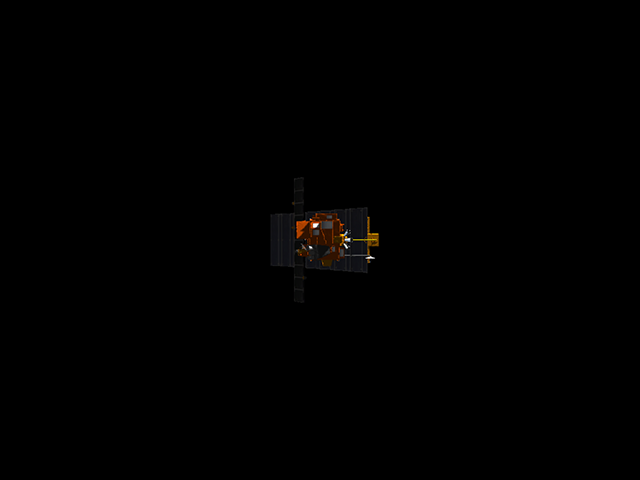}
	\caption{Initial camera frame}
	\label{fig:res-synthetic_snaps_initial}
	\end{subfigure}\hfill%
	\begin{subfigure}[t]{0.24\textwidth}
	\includegraphics[width=\linewidth]{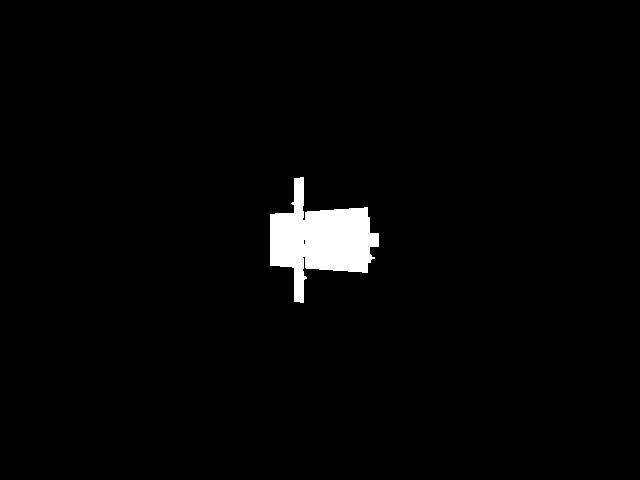}
	\caption{Segmentation}
	\end{subfigure}\hfill%
		\begin{subfigure}[t]{0.24\textwidth}
	\includegraphics[width=\linewidth]{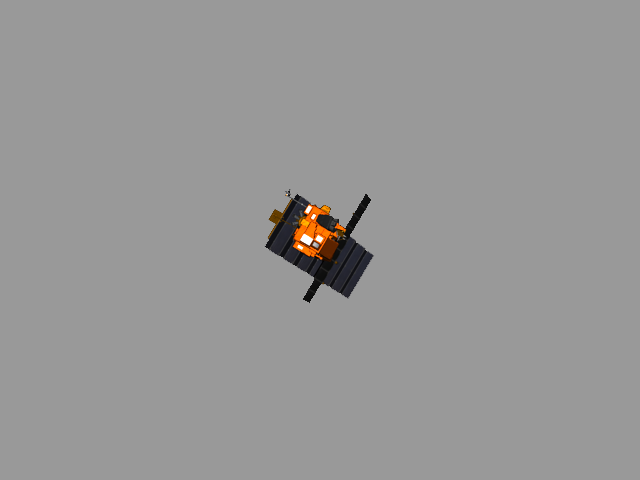}
	\caption{Retrieved keyframe}
	\end{subfigure}\hfill%
		\begin{subfigure}[t]{0.24\textwidth}
	\includegraphics[width=\linewidth]{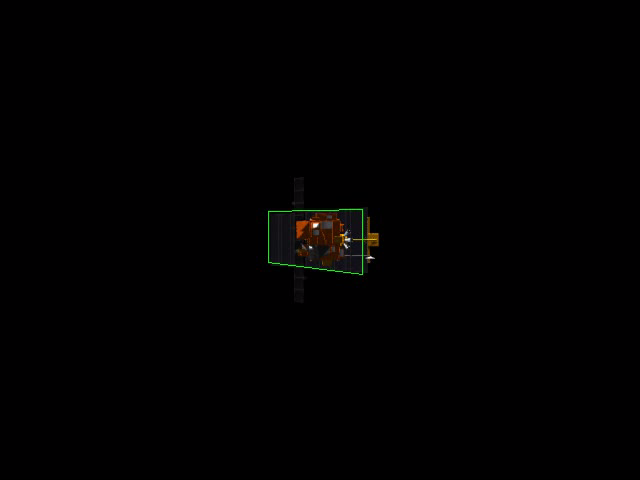}
	\caption{Initial pose at $t=$ \SI{0}{\second}}
	\end{subfigure}\\%
		\begin{subfigure}[t]{0.24\textwidth}
	\includegraphics[width=\linewidth]{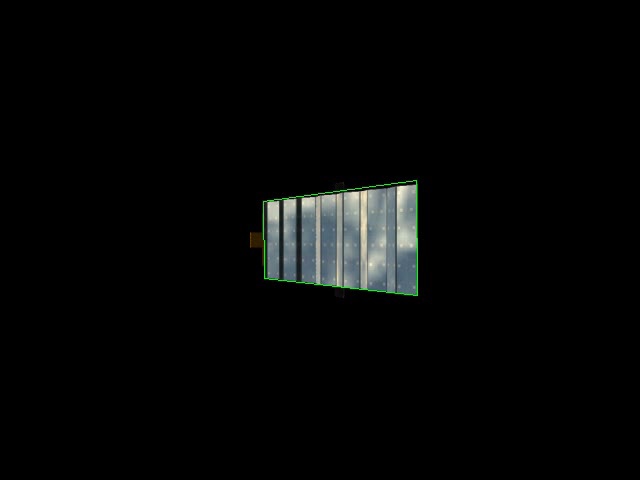}
	\caption{$t=$ \SI{51}{\second}}
	\end{subfigure}\hfill%
    \begin{subfigure}[t]{0.24\textwidth}
	\includegraphics[width=\linewidth]{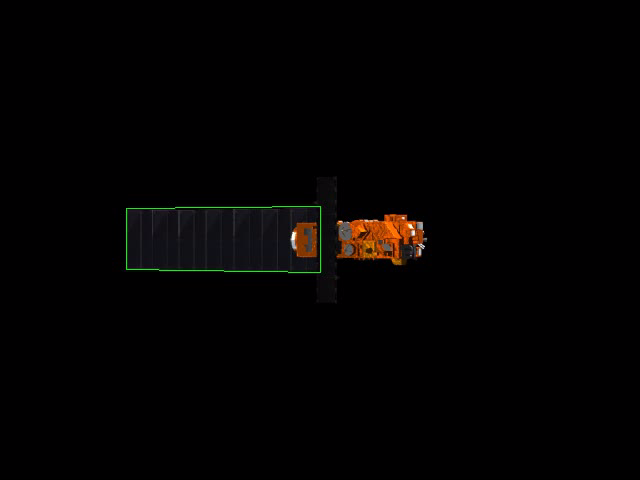}
	\caption{$t=$ \SI{116}{\second}}
	\end{subfigure}\hfill%
	\begin{subfigure}[t]{0.24\textwidth}
	\includegraphics[width=\linewidth]{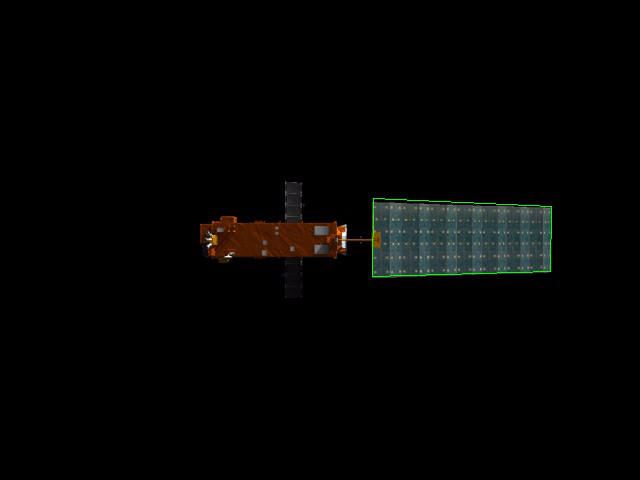}
	\caption{$t=$ \SI{180}{\second}}
	\end{subfigure}\hfill%
		\begin{subfigure}[t]{0.24\textwidth}
	\includegraphics[width=\linewidth]{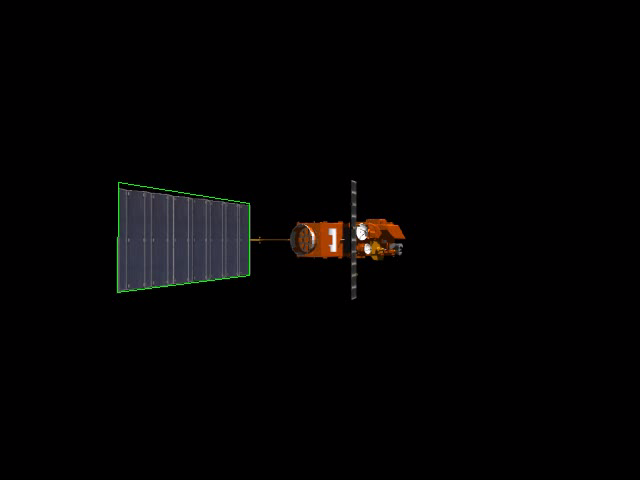}
	\caption{$t=$ \SI{244}{\second}}
	\end{subfigure}\\%
 \caption{Results of the relative pose estimation for the synthetic Envisat rendezvous dataset. The edges of the solar panel are reprojected in green using the estimated pose.}
\label{fig:res-synthetic_snaps}
\end{figure}

The results of the relative pose estimation are shown in Fig. \ref{fig:res-synthetic_fine1}. Three different methods are compared:  \begin{enumerate*} [label=\roman*\upshape)] \item EP$n$P with feature point matches \cite{lepetit2009epnp} and RANSAC for outlier rejection (denoted by ``M1''); \item the method developed in the authors' previous paper \cite{rondao2018multiview}, using M-estimation fusing point and edge features (denoted by ``M2''); and \item the framework proposed in this paper (denoted by ``M3'') \end{enumerate*}. In the case of M1 and M2, the next keyframe is determined by the pose estimated in the previous time-step. It can be seen that M1 is not able to converge at all. M2 yields a decent estimate for the first few frames, but the error quickly begins to grow until the algorithm diverges completely at $t=$ \SI{5}{\second}. This occurs due to the challenging initial configuration of the target, where its perceived cross-section from the chaser's point of view is minimal (cf. Fig. \ref{fig:res-synthetic_snaps_initial}). On the other hand, the proposed framework (M3) converges at around  $t=$ \SI{10}{\second}. The steady state error is bounded at approximately \SI{2}{\metre} for position, which corresponds to \SI{4}{\percent} of the range distance, whereas the attitude error is bounded at \SI{1.5}{\deg}. Figure \ref{fig:res-synthetic_features} exhibits some figures of merit pertaining to the point and edge features in the simulation run, namely the number of matches and inliers, the \gls{rmse} of the M-estimation, and the feature visibility with respect to the validation gating applied prior to the filtering. A threshold of \SI{2.5}{\pixel} was applied for the points and \SI{5}{\pixel} for the edges. The number of matches fluctuates more in the case of edges; this is due to the relative circular trajectory in which the imaging area of the target changes. The peaks correspond to the sections where the $\vect{t}_1-\vect{t}_2$ plane $\in \rframe{T}$ is imaged by the chaser, whereas the valleys correspond to an imaging of the $\vect{t}_2-\vect{t}_3$ plane. However, the \gls{rmse} of the point features is on average greater than that of the edges, which results in fewer periods of visibility for the former. The periods of higher \gls{rmse} correspond to images of the $\vect{t}_1-\vect{t}_2$ plane, where the image of the target is dominated by the \gls{mli} coverage and the solar panel. Despite this, the guided feature matching algorithm prevents the point features' visibility from being constantly null during these periods.

\begin{figure}[]
	\centering
	\begin{subfigure}[t]{0.48\textwidth}
	\includegraphics[width=\linewidth]{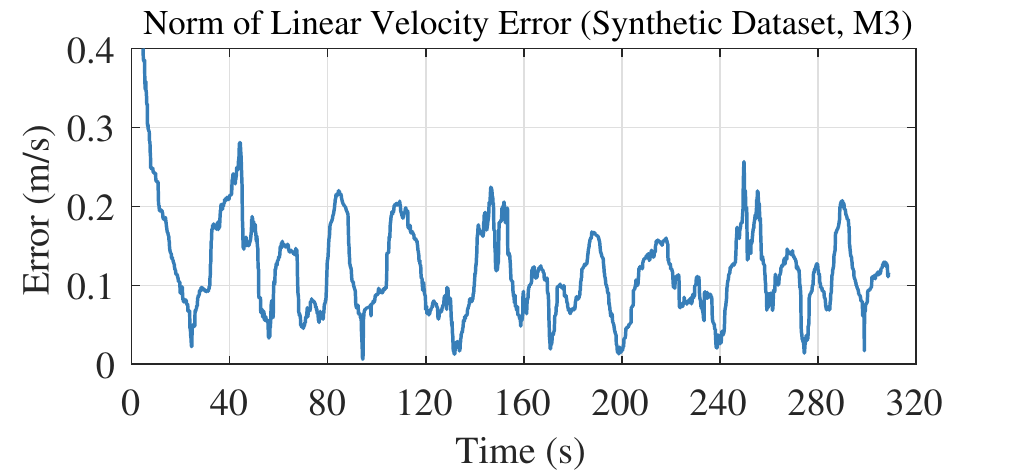} 	\centering
	\caption{Linear velocity} \label{fig:res-synthetic_fine2_linear}
	\end{subfigure}\hfill%
	\begin{subfigure}[t]{0.48\textwidth}
	\includegraphics[width=\linewidth]{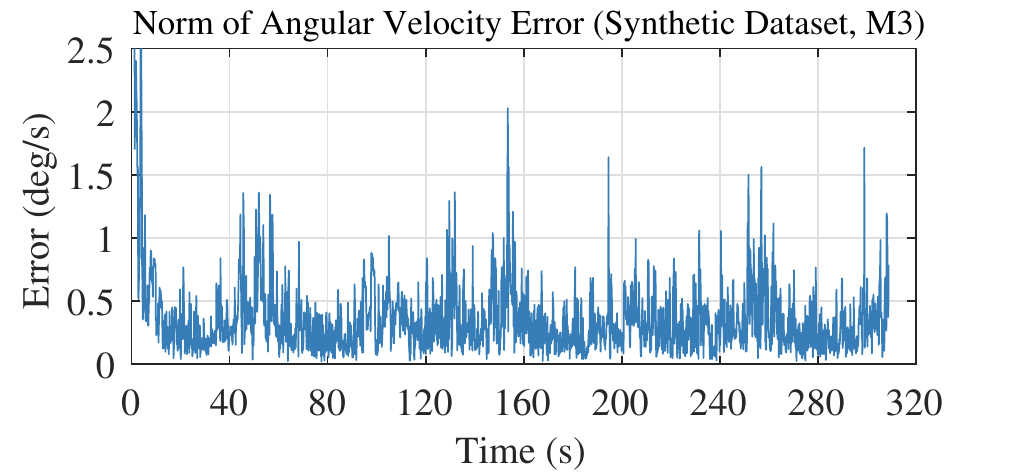} 	\centering
	\caption{Angular velocity} \label{fig:res-synthetic_fine1_angular}
	\end{subfigure}%
 \caption{Nominal velocity estimation errors for the synthetic dataset for the proposed framework.}
\label{fig:res-synthetic_fine2}
\end{figure}

The relative velocity estimation errors as output by the filter are also shown (Fig. \ref{fig:res-synthetic_fine2}). The linear velocity steady-state error does not exceed \SI{0.3}{\metre\per\second}, whereas the angular velocity is bounded at \SI{2}{\deg\per\second}. The latter quantity is much noisier than the former, since there are two out-of-plane dimensions, compared to one for the linear velocity, highlighting the challenge of depth estimation with a monocular setup. Lastly, Fig. \ref{fig:res-synthetic_snaps} illustrates the initialization procedure and some frames of the synthetic dataset with the estimated pose superimposed.

\subsubsection{Laboratory Dataset}
\label{subsubsec:res-lab}

\begin{figure}[]
\centering
\includegraphics[width=0.5\linewidth]{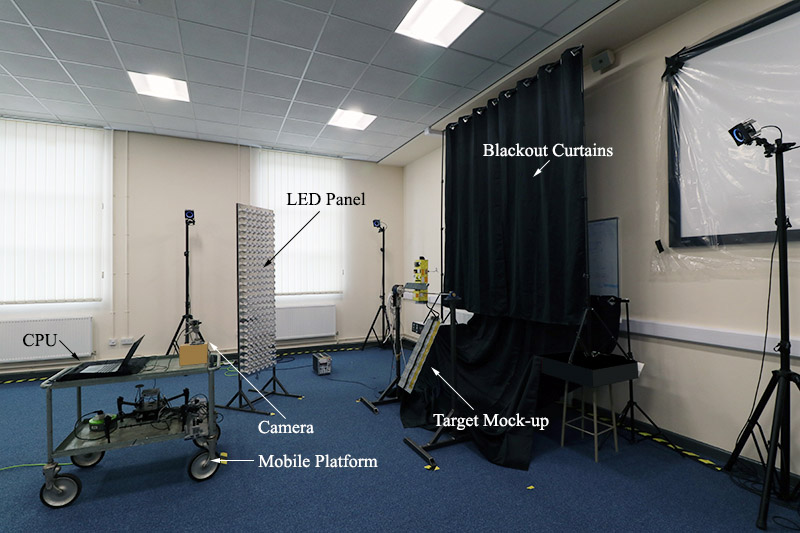}
\caption{Setup for laboratory validation.}
\label{fig:res-lab_setup}
\end{figure}

In this section, the framework is validated in laboratory. The experimental setup is portrayed in Fig. \ref{fig:res-lab_setup}. A scaled model of Envisat was built with 1-\gls{dof} in rotational motion (along the $\vect{t}_1$ axis). Due to the dimensions of the mock-up, a custom-built LED illumination panel running at \SI{900}{\watt} is used to simulate direct sunlight. The mvBlueFOX MLC200wC camera with a \SI{3}{\milli\metre} lens is used to acquire the dataset. The initial position and attitude are, respectively, $\vect{t}^C_{C/T, 0} = \icolsmall{-0.1418,-0.0449,1.9312}^\top$ \si{\metre} and  $\vect{q}_{C/T,0} = \icolsmall{-0.0356,-0.0336,-0.01751,0.9986}^\top$. The motion is analogous to the simulated one (cf. Fig. \ref{fig:res-synthetic_traj}) and the rotation rate is constant and equal to \SI{5.73}{\deg\per\second}. The Envisat mockup is modeled in Blender and textured with real images to generate the offline keyframe database. The ground truth is obtained by manually registering the first frame and by propagating the state using the rigid body's constant rotation rate from the static camera's viewpoint. It is assumed that the background can be subtracted correctly.

\begin{figure}[]
	\centering
	\begin{subfigure}[t]{0.48\textwidth} 	\centering
	\includegraphics[width=\linewidth]{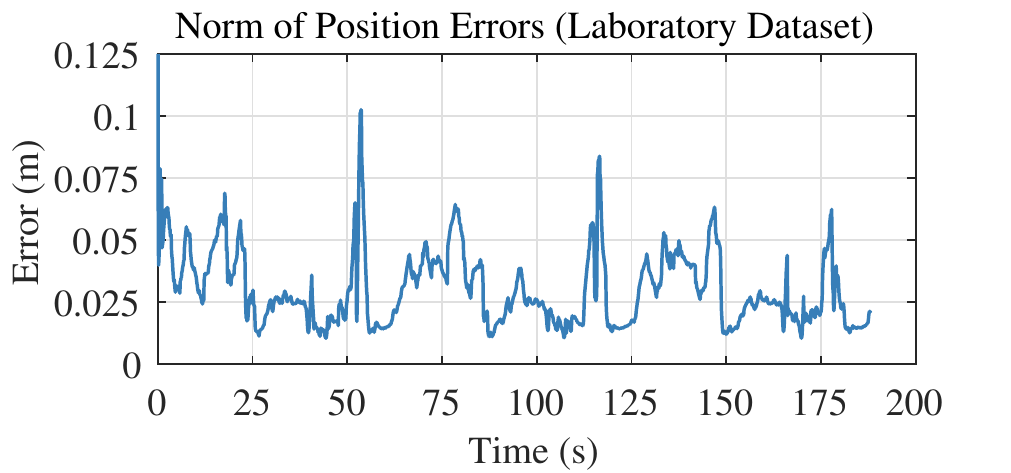}
	\caption{Position} \label{fig:res-lab_pose_position}
	\end{subfigure}\hfill%
	\begin{subfigure}[t]{0.48\textwidth} 	\centering
	\includegraphics[width=\linewidth]{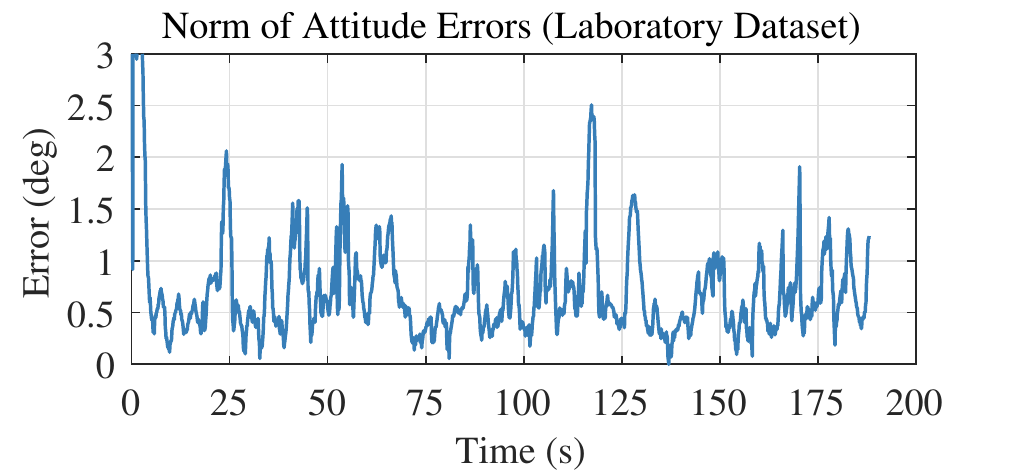}
	\caption{Attitude} \label{fig:res-lab_pose_attitude}
	\end{subfigure}%
 \caption{Pose estimation errors for the laboratory dataset.}
\label{fig:res-lab_pose}
\end{figure}

\begin{figure}[]
	\centering
	\begin{subfigure}[t]{0.48\textwidth} 	\centering
	\includegraphics[width=\linewidth]{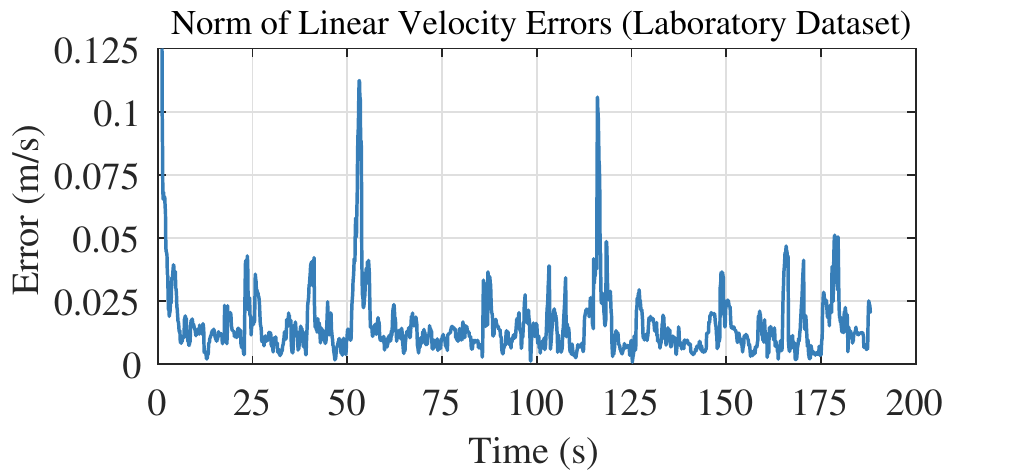}
	\caption{Linear velocity} \label{fig:res-lab_velocity_lin}
	\end{subfigure}\hfill%
	\begin{subfigure}[t]{0.48\textwidth} 	\centering
	\includegraphics[width=\linewidth]{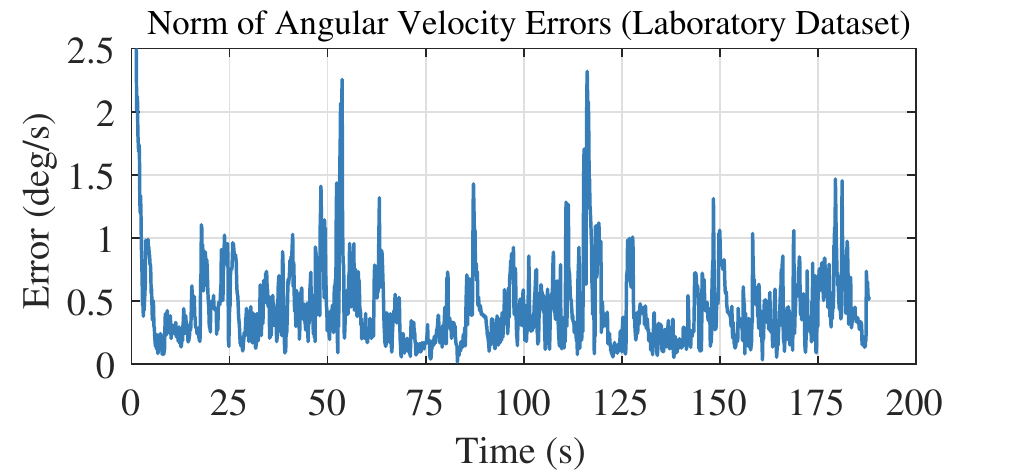}
	\caption{Angular velocity} \label{fig:res-lab_velocity_ang}
	\end{subfigure}%
 \caption{Velocity estimation errors for the laboratory dataset.}
\label{fig:res-lab_velocity}
\end{figure}

\begin{figure}[]
	\centering
		\begin{subfigure}[t]{0.24\textwidth}
	\includegraphics[width=\linewidth]{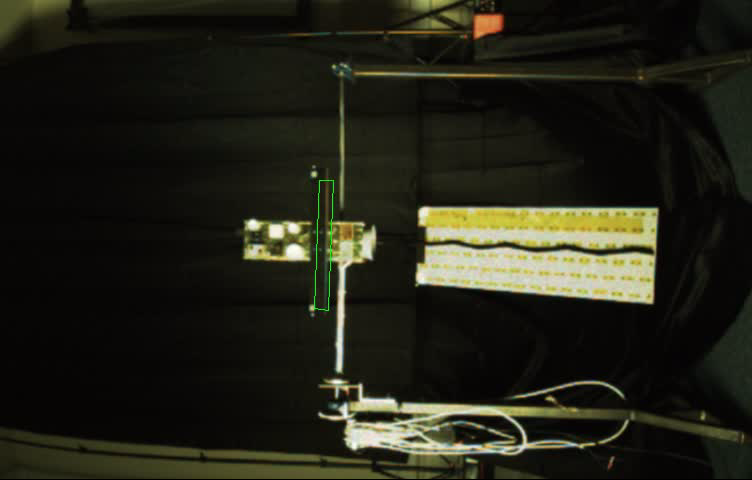}
	\caption{Initial pose at $t=$ \SI{0}{\second}}
	\end{subfigure}\hfill%
		\begin{subfigure}[t]{0.24\textwidth}
	\includegraphics[width=\linewidth]{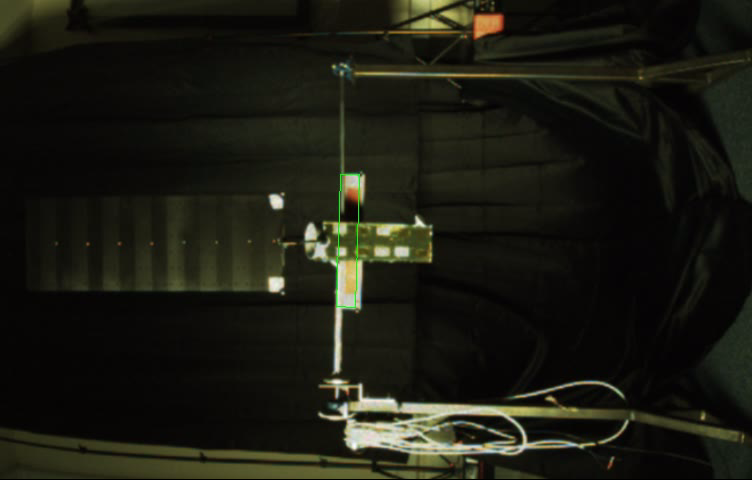}
	\caption{$t=$ \SI{37}{\second}}
	\end{subfigure}\hfill%
    \begin{subfigure}[t]{0.24\textwidth}
	\includegraphics[width=\linewidth]{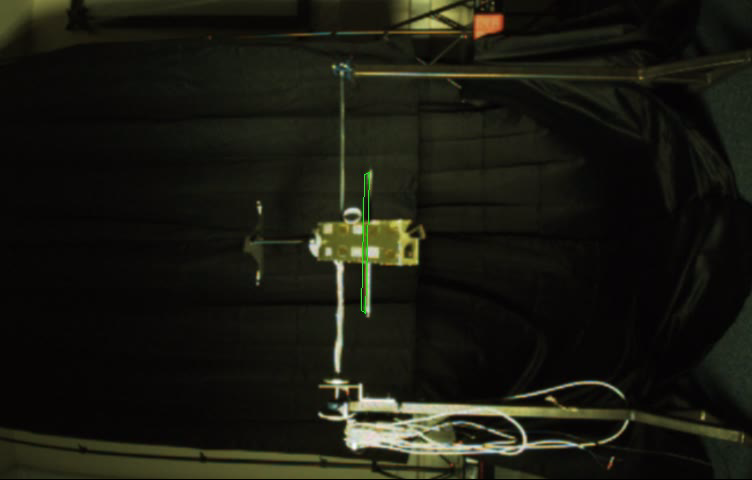}
	\caption{$t=$ \SI{87}{\second}}
	\end{subfigure}\hfill%
	\begin{subfigure}[t]{0.24\textwidth}
	\includegraphics[width=\linewidth]{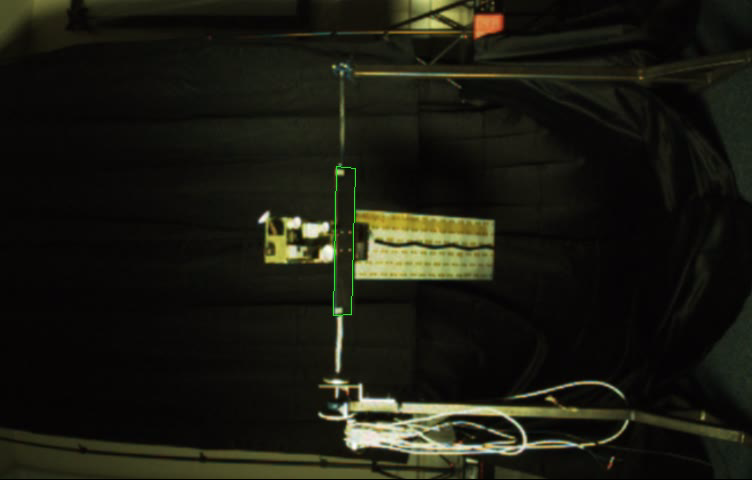}
	\caption{$t=$ \SI{137}{\second}}
	\end{subfigure}
 \caption{Results of the relative pose estimation for the laboratory Envisat rendezvous dataset. The edges of the \gls{sar} are reprojected in green using the estimated pose.}
\label{fig:res-lab_snaps}
\end{figure}

Figure \ref{fig:res-lab_pose} displays the pose estimation errors as achieved by the framework; Fig. \ref{fig:res-lab_velocity} shows the velocity estimation errors; Fig. \ref{fig:res-lab_snaps} depicts a set of frames from the lab sequence including initialization and reprojection of the pose. The magnitude of the attained errors is analogous to that obtained for the synthetic dataset: a maximum of \SI{5}{\percent} position estimation error with respect to the range, whereas the attitude error in steady-state does not exceed \SI{2.5}{\deg} (while remaining generally under \SI{1.5}{\deg}). As expected, the angular velocity error estimation is more noisy than the linear velocity one.

\section{Conclusion}
\label{sec:co}

In this work, a robust, innovative model-based solution for spacecraft relative navigation using a single visible wavelength camera has been developed. The proposed contribution stands on the fact that the relative navigation solution is achieved using a set of discrete keyframes rendered in an offline stage from a three-dimensional model of the target spacecraft, where a 3D-2D problem is converted into a 2D-2D approach relying only on computer vision methods capable of running exclusively on the CPU. The proposed method was tested both using a synthetic dataset closely simulating the imaging conditions experienced in-orbit and a real dataset generated in laboratory, featuring a (relative) circular observation trajectory about the complex spacecraft Envisat. The aspect of the target in each keyframe is learned using shape features and \glspl{gmm} that are used to train a Bayesian classifier; the coarse viewpoint as seen by the camera is identified approximately $\SI{90}{\percent}$ of the time with an error under \SI{20}{\deg}. The pose estimate is refined by matching hybrid features between the current image and train keyframe. Automatic outlier rejection is assured via M-estimation. An \gls{ekf} is employed to fuse the pose hypotheses generated by each feature type; it is designed to operate on the tangent space of $\sethree$, allowing it to seamlessly integrate the previous stage by assimilating the covariance matrices generated by the M-estimation directly as the measurement noise, independently of the pose parameterization. The attained solution showcases rapid convergence and yields a maximum error of \SI{5}{\percent} of the range distance in position and \SI{2.5}{\deg} in attitude; on average, steady-state errors are observed in the order of $\SI{1}{\percent}$ of the range in position and \SI{0.5}{\deg} in attitude. A novel matching strategy by predicting feature locations using the \gls{ekf}, alongside a \gls{rmse}-based validation gate, assures the stability and accuracy of the solution is maintained, even in the face of highly discrepant frames with respect to the database keyframes caused by light-scattering \gls{mli} and solar panel reflections.

\section*{Acknowledgements}

The authors would like to thank Olivier Dubois-Matra from \gls{esa} for providing them access to the Astos Camera Simulator (contract no. 4000117583/16/NL/HK/as).

\bibliography{references}

\begin{thebibliography}{54}
\newcommand{\enquote}[1]{``#1''}
\providecommand{\natexlab}[1]{#1}
\providecommand{\url}[1]{\texttt{#1}}
\providecommand{\urlprefix}{URL }
\expandafter\ifx\csname urlstyle\endcsname\relax
  \providecommand{\doi}[1]{doi:\discretionary{}{}{}#1}\else
  \providecommand{\doi}{doi:\discretionary{}{}{}\begingroup
  \urlstyle{rm}\Url}\fi

\bibitem[{Bhaskaran et~al.(1998)Bhaskaran, Desai, Dumont, Kennedy, Null,
  Owen~Jr, Riedel, Synnott, and Werner}]{bhaskaran1998orbit}
Bhaskaran, S., Desai, S., Dumont, P., Kennedy, B., Null, G., Owen~Jr, W.,
  Riedel, J., Synnott, S., and Werner, R., \enquote{{Orbit Determination
  Performance Evaluation of the Deep Space 1 Autonomous Navigation System},}
  \emph{AAS/AIAA Space Flight Mechanics Meeting}, Monterey, CA, 1998.

\bibitem[{Castellini et~al.(2015)Castellini, Antal-Wokes, de~Santayana, and
  Vantournhout}]{castellini2015far}
Castellini, F., Antal-Wokes, D., de~Santayana, R.~P., and Vantournhout, K.,
  \enquote{{Far Approach Optical Navigation and Comet Photometry for the
  Rosetta Mission},} \emph{Proceedings of the 25\textsuperscript{th}
  International Symposium on Space Flight Dynamics}, Munich, Germany, 2015.

\bibitem[{Bercovici and McMahon(2017)}]{bercovici2017point}
Bercovici, B., and McMahon, J.~W., \enquote{Point-Cloud Processing Using
  Modified Rodrigues Parameters for Relative Navigation,} \emph{Journal of
  Guidance, Control, and Dynamics}, Vol.~40, No.~12, 2017, pp. 3167--3179.
\newblock \doi{10.2514/1.g002787}.

\bibitem[{Bercovici and McMahon(2019)}]{bercovici2019robust}
Bercovici, B., and McMahon, J.~W., \enquote{Robust Autonomous Small-Body Shape
  Reconstruction and Relative Navigation Using Range Images,} \emph{Journal of
  Guidance, Control, and Dynamics}, Vol.~42, No.~7, 2019, pp. 1473--1488.
\newblock \doi{10.2514/1.g003898}.

\bibitem[{Kechagias-Stamatis et~al.(2019)Kechagias-Stamatis, Aouf, and
  Richardson}]{kechagias2019high}
Kechagias-Stamatis, O., Aouf, N., and Richardson, M., \enquote{High-speed
  multi-dimensional relative navigation for uncooperative space objects,}
  \emph{Acta Astronautica}, Vol. 160, 2019, pp. 388--400.
\newblock \doi{10.1016/j.actaastro.2019.04.050}.

\bibitem[{Biesbroek et~al.(2017)Biesbroek, Innocenti, Wolahan, and
  Serrano}]{biesbroek2017edeorbit}
Biesbroek, R., Innocenti, L., Wolahan, A., and Serrano, S.,
  \enquote{{e.Deorbit–ESA’s Active Debris Removal Mission},}
  \emph{7\textsuperscript{th} European Conference on Space Debris}, ESA Space
  Debris Office, 2017.

\bibitem[{Tweddle et~al.(2014)Tweddle, Saenz-Otero, Leonard, and
  Miller}]{tweddle2015factor}
Tweddle, B.~E., Saenz-Otero, A., Leonard, J.~J., and Miller, D.~W.,
  \enquote{Factor Graph Modeling of Rigid-body Dynamics for Localization,
  Mapping, and Parameter Estimation of a Spinning Object in Space,}
  \emph{Journal of Field Robotics}, Vol.~32, No.~6, 2014, pp. 897--933.
\newblock \doi{10.1002/rob.21548}.

\bibitem[{Olson et~al.(2016)Olson, Russell, and Carpenter}]{olson2016small}
Olson, C.~G., Russell, R.~P., and Carpenter, J.~R., \enquote{Small-Body Optical
  Navigation Using the Additive Divided Difference Sigma Point Filter,}
  \emph{Journal of Guidance, Control, and Dynamics}, Vol.~39, No.~4, 2016, pp.
  922--928.
\newblock \doi{10.2514/1.g001160}.

\bibitem[{Razgus et~al.(2017)Razgus, Mooij, and Choukroun}]{razgus2017relative}
Razgus, B., Mooij, E., and Choukroun, D., \enquote{Relative Navigation in
  Asteroid Missions Using Dual Quaternion Filtering,} \emph{Journal of
  Guidance, Control, and Dynamics}, Vol.~40, No.~9, 2017, pp. 2151--2166.
\newblock \doi{10.2514/1.g002805}.

\bibitem[{Y{\i}lmaz et~al.(2017)Y{\i}lmaz, Aouf, Majewski, and
  Sanchez-Gestido}]{yilmaz2017using}
Y{\i}lmaz, O., Aouf, N., Majewski, L., and Sanchez-Gestido, M., \enquote{{Using
  Infrared Based Relative Navigation for Active Debris Removal},}
  \emph{10\textsuperscript{th} International ESA Conference on Guidance,
  Navigation and Control Systems}, Salzburg, Austria, 2017, pp. 1--16.
\newblock \urlprefix\url{http://dspace.lib.cranfield.ac.uk/handle/1826/12079}.

\bibitem[{Augenstein and Rock(2009)}]{augenstein2009simultaneous}
Augenstein, S., and Rock, S., \enquote{Simultaneous Estimaton of Target Pose
  and 3-D Shape Using the {FastSLAM} Algorithm,} \emph{{AIAA} Guidance,
  Navigation, and Control Conference}, American Institute of Aeronautics and
  Astronautics, 2009.
\newblock \doi{10.2514/6.2009-5782}.

\bibitem[{Cropp(2001)}]{cropp2001pose}
Cropp, A., \enquote{{Pose Estimation and Relative Orbit Determination of a
  Nearby Target Microsatellite using Passive Imagery},} Ph.D. thesis,
  University of Surrey, United Kingdom, 2001.
\newblock \urlprefix\url{http://epubs.surrey.ac.uk/843875/}.

\bibitem[{Kelsey et~al.(2006)Kelsey, Byrne, Cosgrove, Seereeram, and
  Mehra}]{kelsey2006vision}
Kelsey, J., Byrne, J., Cosgrove, M., Seereeram, S., and Mehra, R.,
  \enquote{Vision-Based Relative Pose Estimation for Autonomous Rendezvous And
  Docking,} \emph{2006 {IEEE} Aerospace Conference}, {IEEE}, 2006.
\newblock \doi{10.1109/aero.2006.1655916}.

\bibitem[{Petit et~al.(2013)Petit, Marchand, and Kanani}]{petit2013robust}
Petit, A., Marchand, E., and Kanani, K., \enquote{A Robust Model-Based Tracker
  Combining Geometrical and Color Edge Information,} \emph{2013 {IEEE}/{RSJ}
  International Conference on Intelligent Robots and Systems}, {IEEE}, 2013.
\newblock \doi{10.1109/iros.2013.6696887}.

\bibitem[{Oumer(2014)}]{oumer2014monocular}
Oumer, N.~W., \enquote{Monocular 3D Pose Tracking of a Specular Object,}
  \emph{Proceedings of the 9th International Conference on Computer Vision
  Theory and Applications}, {SCITEPRESS} - Science and and Technology
  Publications, 2014.
\newblock \doi{10.5220/0004667304580465}.

\bibitem[{Cai et~al.(2015)Cai, Huang, Zhang, and Wang}]{cai2015tsr}
Cai, J., Huang, P., Zhang, B., and Wang, D., \enquote{A {TSR} Visual Servoing
  System Based on a Novel Dynamic Template Matching Method,} \emph{Sensors},
  Vol.~15, No.~12, 2015, pp. 32152--32167.
\newblock \doi{10.3390/s151229884}.

\bibitem[{Zou et~al.(2016)Zou, Wang, Zhang, and Song}]{zou2016combining}
Zou, Y., Wang, X., Zhang, T., and Song, J., \enquote{Combining Point and Edge
  for Satellite Pose Tracking Under Illumination Varying,} \emph{2016 12th
  World Congress on Intelligent Control and Automation ({WCICA})}, {IEEE},
  2016.
\newblock \doi{10.1109/wcica.2016.7578814}.

\bibitem[{Gansmann et~al.(2017)Gansmann, Mongrard, and
  Ankersen}]{gansmann20173d}
Gansmann, M., Mongrard, O., and Ankersen, F., \enquote{{3D Model-Based Relative
  Pose Estimation for Rendezvous and Docking Using Edge Features},}
  \emph{10\textsuperscript{th} International ESA Conference on Guidance,
  Navigation and Control Systems}, 2017.

\bibitem[{Szeliski(2011)}]{szeliski2010computer}
Szeliski, R., \emph{{Computer Vision: Algorithms and Applications}},
  1\textsuperscript{st} ed., Springer-Verlag, London, UK, 2011, pp. 45--47,
  284--286.
\newblock \doi{10.1007/978-1-84882-935-0}.

\bibitem[{Murray et~al.(1994)Murray, Sastry, and
  Zexiang}]{murray1994mathematical}
Murray, R.~M., Sastry, S.~S., and Zexiang, L., \emph{{A Mathematical
  Introduction to Robotic Manipulation}}, 1\textsuperscript{st} ed., CRC Press,
  Inc., Boca Raton, FL, USA, 1994, pp. 34--39, 41--42, 53--54.
\newblock \doi{10.1201/9781315136370}.

\bibitem[{Stillwell(2008)}]{stillwell2008naive}
Stillwell, J., \emph{{Naive Lie Theory}}, Springer, New York, NY, 2008, pp. 82,
  98.
\newblock \doi{10.1007/978-0-387-78214-0}.

\bibitem[{Gallier(2011)}]{gallier2013geometric}
Gallier, J., \emph{{Geometric Methods and Applications: For Computer Science
  and Engineering}}, 2\textsuperscript{nd} ed., Springer, New York, NY, 2011,
  pp. 468, 504--505.
\newblock \doi{10.1007/978-1-4419-9961-0}.

\bibitem[{Selig(2004)}]{selig2004lie}
Selig, J.~M., \enquote{{Lie Groups and Lie Algebras in Robotics},}
  \emph{Computational Noncommutative Algebra and Applications}, edited by
  J.~Byrnes, Springer, Dordrecht, Netherlands, 2004, pp. 101--125.
\newblock \doi{10.1007/1-4020-2307-3_5}.

\bibitem[{Absil et~al.(2008)Absil, Mahony, and
  Sepulchre}]{absil2007optimization}
Absil, P.-A., Mahony, R., and Sepulchre, R., \emph{{Optimization Algorithms on
  Matrix Manifolds}}, Princeton University Press, Princeton, NJ, 2008, pp.
  54--56.
\newblock \doi{10.1515/9781400830244}.

\bibitem[{Gallier and Quaintance(2019)}]{gallier2011notes}
Gallier, J., and Quaintance, J., \enquote{{Differential Geometry and Lie Groups
  I: A Computational Perspective},} University of Pennsylvania (book in
  progress), August 2019.
\newblock \urlprefix\url{http://www.cis.upenn.edu/~jean/gbooks/manif.html}, p.
  591.

\bibitem[{Markley and Crassidis(2014)}]{markley2014fundamentals}
Markley, F.~L., and Crassidis, J.~L., \emph{{Fundamentals of Spacecraft
  Attitude Determination and Control}}, Space Technology Library, Springer, New
  York, NY, 2014, p.~45.
\newblock \doi{10.1007/978-1-4939-0802-8}.

\bibitem[{{Shuster}(1993)}]{shuster1993survey}
{Shuster}, M.~D., \enquote{{{A Survey of Attitude Representations}},}
  \emph{Journal of the Astronautical Sciences}, Vol.~41, No.~4, 1993, pp.
  439--517.

\bibitem[{Flusser et~al.(2016)Flusser, Suk, and Zitov{\'a}}]{flusser20162d}
Flusser, J., Suk, T., and Zitov{\'a}, B., \emph{2D and 3D Image Analysis by
  Moments}, 1\textsuperscript{st} ed., John Wiley {\&} Sons, Ltd, 2016, pp.
  47--48, 352--356.
\newblock \doi{10.1002/9781119039402}.

\bibitem[{Kintner(1976)}]{kintner1976mathematical}
Kintner, E.~C., \enquote{{On the Mathematical Properties of the Zernike
  Polynomials},} \emph{Optica Acta: International Journal of Optics}, Vol.~23,
  No.~8, 1976, pp. 679--680.

\bibitem[{Duda et~al.(2012)Duda, Hart, and Stork}]{duda2012pattern}
Duda, R.~O., Hart, P.~E., and Stork, D.~G., \emph{{Pattern Classification}},
  2\textsuperscript{nd} ed., John Wiley \& Sons, New York, NY, 2012, pp. 21,
  124--125.

\bibitem[{Figueiredo and Jain(2002)}]{figueiredo2002unsupervised}
Figueiredo, M., and Jain, A., \enquote{Unsupervised Learning of Finite Mixture
  Models,} \emph{{IEEE} Transactions on Pattern Analysis and Machine
  Intelligence}, Vol.~24, No.~3, 2002, pp. 381--396.
\newblock \doi{10.1109/34.990138}.

\bibitem[{Li et~al.(2009)Li, Lee, and Pun}]{li2008complex}
Li, S., Lee, M.-C., and Pun, C.-M., \enquote{Complex Zernike Moments Features
  for Shape-Based Image Retrieval,} \emph{{IEEE} Transactions on Systems, Man,
  and Cybernetics - Part A: Systems and Humans}, Vol.~39, No.~1, 2009, pp.
  227--237.
\newblock \doi{10.1109/tsmca.2008.2007988}.

\bibitem[{Lepetit et~al.(2008)Lepetit, Moreno-Noguer, and
  Fua}]{lepetit2009epnp}
Lepetit, V., Moreno-Noguer, F., and Fua, P., \enquote{{EPnP}: An Accurate O(n)
  Solution to the {PnP} Problem,} \emph{International Journal of Computer
  Vision}, Vol.~81, No.~2, 2008, pp. 155--166.
\newblock \doi{10.1007/s11263-008-0152-6}.

\bibitem[{Hartley and Zisserman(2004)}]{hartley2004multiple}
Hartley, R., and Zisserman, A., \emph{{Multiple View Geometry in Computer
  Vision}}, 2\textsuperscript{nd} ed., Cambridge University Press, Cambridge,
  UK, 2004, pp. 181, 135, 141--142.
\newblock \doi{10.1017/cbo9780511811685}.

\bibitem[{Kanatani(1996)}]{kanatani1996statistical}
Kanatani, K., \emph{{Statistical Optimization for Geometric Computation: Theory
  and Practice}}, Elsevier Science Inc., New York, NY, 1996, p.~67.
\newblock \doi{10.1016/s0923-0459(96)x8019-4}.

\bibitem[{Rondao and Aouf(2018)}]{rondao2018multiview}
Rondao, D., and Aouf, N., \enquote{Multi-View Monocular Pose Estimation for
  Spacecraft Relative Navigation,} \emph{2018 {AIAA} Guidance, Navigation, and
  Control Conference}, American Institute of Aeronautics and Astronautics,
  2018.
\newblock \doi{10.2514/6.2018-2100}.

\bibitem[{Rublee et~al.(2011)Rublee, Rabaud, Konolige, and
  Bradski}]{rublee2011orb}
Rublee, E., Rabaud, V., Konolige, K., and Bradski, G., \enquote{{ORB}: An
  Efficient Alternative to {SIFT} or {SURF},} \emph{2011 International
  Conference on Computer Vision}, {IEEE}, 2011, pp. 2564--2571.
\newblock \doi{10.1109/iccv.2011.6126544}.

\bibitem[{Rosten and Drummond(2006)}]{rosten2006machine}
Rosten, E., and Drummond, T., \enquote{Machine Learning for High-Speed Corner
  Detection,} \emph{Computer Vision {\textendash} {ECCV} 2006}, Springer Berlin
  Heidelberg, 2006, pp. 430--443.
\newblock \doi{10.1007/11744023_34}.

\bibitem[{Rondao et~al.(2018)Rondao, Aouf, and
  Dubois-Matra}]{rondao2018multispectral}
Rondao, D., Aouf, N., and Dubois-Matra, O., \enquote{{Multispectral Image
  Processing for Navigation Using Low Performance Computing},}
  \emph{69\textsuperscript{th} International Astronautical Congress (IAC)
  2018}, IAF, Bremen, Germany, 2018.
\newblock \urlprefix\url{https://dspace.lib.cranfield.ac.uk/handle/1826/13558}.

\bibitem[{Canny(1987)}]{canny1987computational}
Canny, J., \enquote{A Computational Approach to Edge Detection,} \emph{Readings
  in Computer Vision}, Elsevier, 1987, pp. 184--203.
\newblock \doi{10.1016/b978-0-08-051581-6.50024-6}.

\bibitem[{Lee et~al.(2014)Lee, Lee, Zhang, Lim, Chung, and
  Suh}]{lee2014outdoor}
Lee, J.~H., Lee, S., Zhang, G., Lim, J., Chung, W.~K., and Suh, I.~H.,
  \enquote{Outdoor Place Recognition in Urban Environments using Straight
  Lines,} \emph{2014 {IEEE} International Conference on Robotics and Automation
  ({ICRA})}, {IEEE}, 2014.
\newblock \doi{10.1109/icra.2014.6907675}.

\bibitem[{Alahi et~al.(2012)Alahi, Ortiz, and Vandergheynst}]{alahi2012freak}
Alahi, A., Ortiz, R., and Vandergheynst, P., \enquote{{FREAK}: Fast Retina
  Keypoint,} \emph{2012 {IEEE} Conference on Computer Vision and Pattern
  Recognition}, {IEEE}, 2012, pp. 510--517.
\newblock \doi{10.1109/cvpr.2012.6247715}.

\bibitem[{Markovsky and Mahmoodi(2009)}]{markovsky2008least}
Markovsky, I., and Mahmoodi, S., \enquote{Least-Squares Contour Alignment,}
  \emph{{IEEE} Signal Processing Letters}, Vol.~16, No.~1, 2009, pp. 41--44.
\newblock \doi{10.1109/lsp.2008.2008588}.

\bibitem[{Rousseeuw and Leroy(1987)}]{rousseeuw2005robust}
Rousseeuw, P.~J., and Leroy, A.~M., \emph{Robust Regression and Outlier
  Detection}, John Wiley {\&} Sons, Inc., New York, NY, 1987, pp. 1--4, 12.
\newblock \doi{10.1002/0471725382}.

\bibitem[{Zhang(1997)}]{zhang1997parameter}
Zhang, Z., \enquote{Parameter Estimation Techniques: A Tutorial with
  Application to Conic Fitting,} \emph{Image and Vision Computing}, Vol.~15,
  No.~1, 1997, pp. 59--76.
\newblock \doi{10.1016/s0262-8856(96)01112-2}.

\bibitem[{Holland and Welsch(1977)}]{holland1977robust}
Holland, P.~W., and Welsch, R.~E., \enquote{Robust Regression Using Iteratively
  Reweighted Least-Squares,} \emph{Communications in Statistics - Theory and
  Methods}, Vol.~6, No.~9, 1977, pp. 813--827.
\newblock \doi{10.1080/03610927708827533}.

\bibitem[{Huber(1977)}]{huber1977robust}
Huber, P., \enquote{Robust Methods of Estimation of Regression Coefficients,}
  \emph{Series Statistics}, Vol.~8, No.~1, 1977, pp. 41--53.
\newblock \doi{10.1080/02331887708801356}.

\bibitem[{Beaton and Tukey(1974)}]{beaton1974fitting}
Beaton, A.~E., and Tukey, J.~W., \enquote{The Fitting of Power Series, Meaning
  Polynomials, Illustrated on Band-Spectroscopic Data,} \emph{Technometrics},
  Vol.~16, No.~2, 1974, pp. 147--185.
\newblock \doi{10.1080/00401706.1974.10489171}.

\bibitem[{Stewart(1999)}]{stewart1999robust}
Stewart, C.~V., \enquote{Robust Parameter Estimation in Computer Vision,}
  \emph{{SIAM} Review}, Vol.~41, No.~3, 1999, pp. 513--537.
\newblock \doi{10.1137/s0036144598345802}.

\bibitem[{Huber(2009)}]{huber2009robust}
Huber, P.~J., \emph{Robust Statistics}, 2\textsuperscript{nd} ed., John Wiley
  {\&} Sons, Inc., 2009, pp. 175--186.
\newblock \doi{10.1002/0471725250}.

\bibitem[{Barfoot(2017)}]{barfoot2017state}
Barfoot, T.~D., \emph{{State Estimation for Robotics}}, 1\textsuperscript{st}
  ed., Cambridge University Press, New York, NY, 2017, pp. 265, 359.
\newblock \doi{10.1017/9781316671528}.

\bibitem[{Davison et~al.(2007)Davison, Reid, Molton, and
  Stasse}]{davison2007monoslam}
Davison, A.~J., Reid, I.~D., Molton, N.~D., and Stasse, O.,
  \enquote{{MonoSLAM}: Real-Time Single Camera {SLAM},} \emph{{IEEE}
  Transactions on Pattern Analysis and Machine Intelligence}, Vol.~29, No.~6,
  2007, pp. 1052--1067.
\newblock \doi{10.1109/tpami.2007.1049}.

\bibitem[{Grewal and Andrews(2015)}]{grewal2015kalman}
Grewal, M., and Andrews, A., \emph{{Kalman Filtering: Theory and Practice Using
  MATLAB}}, 4\textsuperscript{th} ed., Wiley, Hoboken, NJ, 2015, p. 136.

\bibitem[{{Blanco, Jos\'e-Luis}(2018)}]{blanco2010tutorial}
{Blanco, Jos\'e-Luis}, \enquote{{A Tutorial on SE(3) Transformation
  Parameterizations and On-Manifold Optimization},} Tech. rep., University of
  M\'alaga, Oct. 2018.
\newblock
  \urlprefix\url{https://w3.ual.es/~jlblanco/publications/#publications}.

\end{thebibliography}

\end{document}